\documentclass[runningheads]{llncs}
\usepackage[utf8]{inputenc}
\usepackage[letterpaper,top=2cm,bottom=2cm,left=3cm,right=3cm,marginparwidth=1.75cm]{geometry}

\usepackage{amsmath}
\usepackage{graphicx}
\usepackage{indentfirst}
\usepackage {subfigure}
\usepackage[colorlinks=true, allcolors=blue]{hyperref}
\usepackage{comment}
\usepackage {float}
\usepackage{xcolor}
\usepackage{enumerate}

\title{Evaluation of Explanation Methods of AI - CNNs in Image Classification Tasks with Reference-based and No-reference Metrics}
\author{Alexey Zhukov \email alexey.zhukov@etu.u-bordeaux.fr\\
Jenny Benois-Pineau \email jenny.benois-pineau@u-bordeaux.fr\\
 Romain Giot  \email romain.giot@u-bordeaux.fr}
\institute{ Univ. Bordeaux, CNRS, Bordeaux INP, LaBRI, UMR 5800, F-33400 Talence, France}
\authorrunning{Zhukov et al.}
\titlerunning{Metrics to Evaluate Explanation Methods of Classification CNNs}

\begin{document}
\maketitle
\raggedbottom

\begin{abstract}
The most popular methods in AI-machine learning paradigm are mainly black boxes. This is why explanation of AI decisions is of emergency. Although dedicated explanation tools have been massively developed, the evaluation of their quality remains an open research question. 
In this paper, we generalize the methodologies of evaluation of post-hoc explainers  of CNNs' decisions in visual classification tasks with reference and no-reference based metrics. We apply them on our previously developed explainers  (FEM\footnote{Available in Open Source at https://github.com/labribkb/fem/blob/main/FEM.ipynb}, MLFEM), and popular Grad-CAM.  
The reference-based metrics are \emph{Pearson correlation coefficient} and \emph{Similarity} computed between the explanation map and its ground truth represented by a Gaze Fixation Density Map obtained with a psycho-visual experiment. As a no-reference metric, we use \emph{stability} metric, proposed by Alvarez-Melis and Jaakkola. We study its behaviour, consensus with reference-based metrics and show that in case of several kinds of degradation on input images, this metric is in agreement with reference-based ones. Therefore, it can be used for evaluation of the quality of explainers when the ground truth is not available.
\end{abstract}

\begin{keywords}
Convolutional Neural Network, Explainable Machine Learning, Evaluation
\end{keywords}

\section{Introduction}
 \label{section:Introduction}
    Artificial intelligence-based systems are daily used by the general public. The most popular methods and algorithms for AI are, for the vast majority, black boxes. They  can be an acceptable solution to unimportant problems (in the sense of the degree of impact) but have a fatal flaw for the rest. More than the final solution of a problem, we also need the reasoning and the factors taken into consideration to produce the result.
    We can easily imagine a scenario where, the process to come up with the result, is more important than the result itself. For instance, for tools that are meant to assist humans in taking a decision, it is useful to point out the relevant factors yielding each decision. This is the subject of explainability methods. In their work \cite{InterpretabilityDenisV22} Denis and Varenne show the difference between interpretability and explainability. An \textit{interpretability} of a method for a human subject means the capacity of a representation to see itself composed of elements (signs, figures, concepts, data, etc.) which have meaning for the subject in question.
    This is the subject of naturally interpretable AI methods, such as \cite{ParticulQuenot} building representations from interpretable parts. The \textit{explainability} denotes the capacity of deployment and explainability of the algorithm or its outputs in a series of steps linked together by what a human being can sensibly interpret as causes or reasons. The explainability methods we consider link input and output, showing what elements in the input influence the output of AI tool the most.\\
    
    We also can use explainability methods to assist us in creating better black box methods by an iterative process, removing from CNNs input information or intermediate feature layers irrelevant for the decision. This process can be particularly useful to detect when a network is over parameterized to a problem. For instance, when it is clear, through visualization, that some layers of the network are scrambling previously ordered information, there is an opportunity to simplify the network \cite{DBLP:conf/grapp/HalnautGBA20,Prunning'2021}. Nowadays, the explainability of AI tools is a strongly researched subject \cite{DBLP:journals/corr/ZhouKLOT15} in various classification tasks. In particular, a bunch of methods has been developed for explanation of image classifiers on the basis of Deep Neural Networks (DNNs). We can cite here Grad-CAM \cite{DBLP:journals/corr/SelvarajuDVCPB16}, LRP \cite{inbook} which belong to the most popular methods of the state-of-the-art. These methods called \emph{features attribution} tools allow for identifying the input data which have contributed the most into prediction by classifiers, with an \emph{explanation map}, also named \emph{saliency map} in the context of image classification (here the pixels with a high contribution to the final decision correspond to the hot points of the map).\\
    
    These explanation methods have to be evaluated and compared between them. While the quality of classifiers is evaluated with usual metrics such as accuracy, recall, precision, f-Score on annotated validation and test datasets, the evaluation of the quality of explainers remains an open problem. In \cite{DBLP:conf/icprai/BourrouxBBG22}, the evaluation of explanation maps (ExMs) obtained by a DNN explainers is proposed. It consists in comparison of ExMs with Gaze Fixation Density Maps (GFDMs) obtained from humans who observed images in a given classification task. The comparison is realized with two largely used metrics for comparison of saliency maps: i) Pearson Correlation Coefficient (PCC) and ii) Similarity (SIM) \cite{LeMeur'2013}. The intuition behind this comparison is that the explainer is good in case of true positives if it shows the area in the image which attracted human attention, as Deep Neural networks are biologically inspired models and are supposed to imitate human brain in decision-making. Other metrics have been recently proposed in \cite{DBLP:journals/corr/abs-2102-13076}. \\

    This paper evaluates explanation methods FEM~\cite{ahmedasiffuad:hal-02963298},  Multilayered FEM (MLFEM)~\cite{DBLP:conf/icprai/BourrouxBBG22} and Grad-CAM~\cite{DBLP:journals/corr/SelvarajuDVCPB16} using two approaches: the GFDM-based metrics we previously proposed and the so-called \emph{stability metric} from \cite{DBLP:journals/corr/abs-2102-13076}. We also study how this metric correlates to open the way to evaluation of the quality of explainers without ground truth.\\
    
    Section \ref{section:SOA} summarizes the SOTA in evaluation of explainers for image classification. Section \ref{section:Evaluation of explanation}  explains the metric and the evaluation methodology. Section \ref{section:protocol} describes the evaluation protocol and Section \ref{section:results} describes its results. Section \ref{section:conclusion} concludes this work and outlines its perspectives. \\

\section{State-of-the-art in Evaluation of Explainers for CNN classification}
    \label{section:SOA}

    The first methods to evaluate feature attribution methods were purely qualitative and performed by humans \cite{DBLP:journals/corr/SelvarajuDVCPB16}. A step ahead to the automatic evaluation of explainers has been recently made in \cite{DBLP:conf/icprai/BourrouxBBG22} when comparing the ExMs with Gaze Fixation Density Maps (GFDMs) computed upon gaze fixations of human observers. The latter were instructed to observe images with the same classification task in mind, which was required from a CNN classifier. If the classification result by a CNN is correct, then the concordance of a human GFDM and an explanation map will be observed, it means that the explainer is of a good quality. The latter could be measured by comparison metrics of saliency maps, such as in \cite{LeMeur'2013}. Nevertheless, such an evaluation requires the availability of ground truth, as GFDMs.\\
    
    Both, interpretation by humans or comparison with human interest expressed by GFDMs can be qualified as \textit{reference-based} evaluation. While it is quite tedious to conduct a human judgement experiment for evaluation of an explainer, the comparison of explanation maps with human GFDMs is quite realistic nowadays. Indeed, a large amount of publicly available databases with Gaze Fixations or their approximations in action-prehension paradigm do exist, such as Hollywood-2 \cite{10.1109/TPAMI.2014.2366154}, DHF1K \cite{wang2018revisiting}, IRCCyN \cite{5414458} (for video), or MexCulture \cite{DBLP:journals/pr/ObesoBGR22}, SALICON \cite{jiang2015salicon} and others.\\
    
    It is a common knowledge that Deep Neural Network(DNN) models are data dependent and cannot be applied without transfer learning on the data which do not follow the distribution on which the DNN was trained. In case of their explainers, we suppose that \textit{a good explainer is universal}. It has to explain a DNN model trained on any data. Thus, if its quality can be assessed on an available ground truth, it can be considered for other explanation tasks. \\

    Another research trend is to design an evaluation strategy which quantifies the sensitivity of an explainer as of any other method to the perturbations in the data. In this case, to evaluate an explainer, human judgment or GFDMs, that is "a reference" is not needed: we can call the metrics proposed in this case as \textit{no-reference}. One group of these methods constitute the fidelity metrics \cite{DBLP:conf/icprai/GomezFM22}. These evaluation methods are based on the principle that if the perturbations are induced in the parts of input data highlighted as important by explainers, then the classification score will change. In which case the explainer can be considered as good. The metrics Deletion Area Under Curve (DAUC) and Insertion Area Under Curve (IAUC) proposed in \cite{DBLP:conf/bmvc/PetsiukDS18} are based on this principle. DAUC tracks the changes in the classification score of the image where the areas are masked progressively according to their importance in the explanation map from the highest score pixels to the full masking. The metric is computed as the area under curve of the class score of an image as a function of masked proportion of its pixels. The lower is this metrics, the better is the explainer, as masking of important (accordingly to the explainer) parts in the input should yield the decrease in initial class score. The computation of IAUC follows the opposite strategy. Here a strongly blurred image is considered first, then are added pixels accordingly to their importance in the ExM. Higher is this metric, better is the explainer. \\

    Other metrics such as “Increase in Confidence” or “Average Drop” or “Average Drop in Deletion” have been proposed in \cite{DBLP:conf/wacv/ChattopadhyaySH18}. All of them are based on the changes of the class score for a given image after it has been modified accordingly to the importance of pixels in explanation maps. In \cite{DBLP:conf/icprai/GomezFM22} the authors criticize DAUC and IAUC for the fact that they use only rank of the score and not its value, and propose Deletion Correlation (DC) and Insertion Correlation (IC) metrics. They are computed as correlation coefficient between the difference of scores due to masking or adding details and the score of saliency of pixels masked/added progressively. All these metrics are based on the influence of perturbations in images accordingly to explanation maps, on the classification score. \\

    The metric Sparsity \cite{DBLP:conf/icprai/GomezFM22} qualifies the distribution of importance scores in the explanation maps per se. Its authors claim that higher sparsity makes the map more interpretable by humans, as it is concentrated on a small amount of elements. \\

    All these metrics are objective as they do not put the human in the loop (except sparsity) and have their reason to exist. Nevertheless, it is interesting to come back to the general approach in design of image processing and analysis algorithms, such as their stability with regard to the level of noise and/or degradation in the input images. Hence, we consider the stability of explanation maps with regard to the  noise and degradations usual in digital images. \\

    In his work, Bodria \cite{DBLP:journals/corr/abs-2102-13076} cites various methods to test the stability of the explainers for black-box models. The method based on the Lipschitz constant, proposed by \cite{DBLP:journals/corr/abs-1806-07538}  seemed to us the most appropriate to measure stability of the ExMs in case when the classifier is stable to the noise. The main potential advantages of the method are that it is quite general, and therefore does not depend on the original design of the system, and is easy to implement. The method does not require the references, i.e. Gaze Fixation Density Maps. For its operation, it is enough to have a wide layer of control (original) data and their corrupted versions. The creation of corrupted data also does not pose great financial and technical problems for the current capabilities of the industry and the scientific community. Thus, thanks to a more detailed study and a wider range of experiments, we have the opportunity to obtain a potentially high-precision and low-cost tool to test the stability of explainers. Furthermore, despite Bodria speaks about evaluation of ``black-box" explainers, the methodology is generic.

\section{Evaluation of CNN Explainers with Reference-based and No-reference-based Metrics}
\label{section:Evaluation of explanation}
This section introduces the evaluation metrics (with and without reference) of explainers and presents our evaluation methodology. 
 
    \subsection{Reference-based Metrics with Gaze Fixation Density Maps}
    \label{section: MetricsWithGFDM}

    Reference based metrics need a ground truth that can be created with a Gaze Fixation Density Map that is then compared to the explanation map using a dedicated comparison metric.
    
\begin{figure}[H]
        \centering
        \subfigure[\label{fig:SALICON_dataset}]{
        \includegraphics[width=1\textwidth]{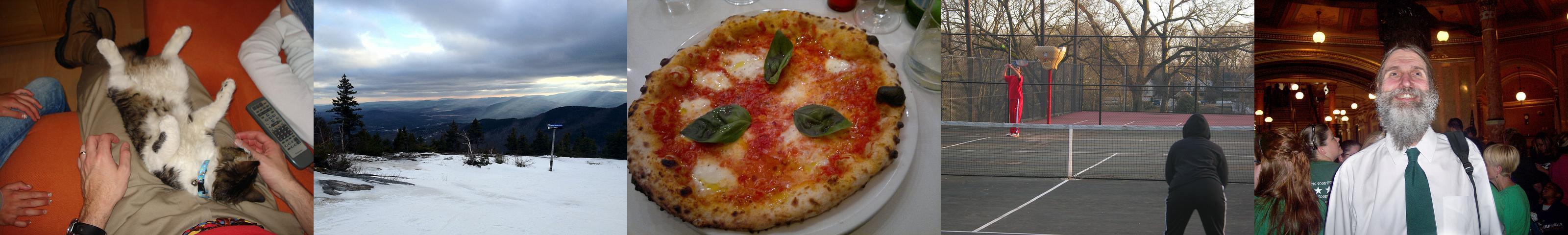}      
        }
        \subfigure[ \label{fig:GFDM}]{
        \includegraphics[width=1\textwidth]{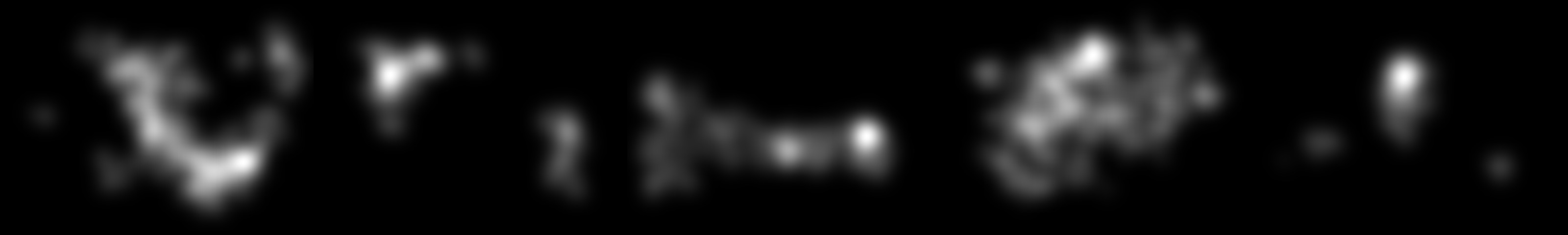}
        }
        \caption{(a) Example of 5 images from SALICON dataset. (b) GFDMs of these images.\label{fig_salicon}}
        \end{figure}

    \paragraph{\textbf{Gaze Fixation Density Map as a Reference}}
    \label{subsubsection:GFDM}
        A Gaze Fixation Density Map (GFDM) is a means to identify parts of an image relevant to a human. The general principle of GFDM computation consists in conducting a psycho-visual experiment, when humans observe visual content and their gaze fixations are recorded by an eye tracking device. Then on each fixation a Gaussian surface is centred, which scale parameter $\sigma$ is computed from the geometry of the experiment to represent the projection of the most sensitive retina area, the fovea, into the image plane. Summing up and normalizing by maximum multi-Gaussian surface from different observers on the same image, the GFDM is obtained. This is the case of GFDMs used e.g. in \cite{DBLP:journals/pr/ObesoBGR22} and available in referenced MexCulture Dataset. In SALICON \cite{jiang2015salicon} dataset, "gaze fixations" were obtained by mouse clicks using visual action anticipation paradigm. Some examples of GFDMs from SALICON  dataset  \cite{jiang2015salicon}  are given in figure \ref{fig_salicon}.  
        Bourroux \emph{et al.}~\cite{DBLP:conf/icprai/BourrouxBBG22} used GFDM as the ground truth for comparing explanation maps with it with PCC, equation \ref{eq:Corr} and SIM, equation \ref{eq:Sim} metrics. 
        
    \paragraph{\textbf{PCC and SIM comparison metrics}}
    \label{subsubsec:PCC_SIM}
    In \cite{DBLP:conf/icprai/BourrouxBBG22} evaluation of ExMs was proposed by their comparison with GFDMs. Two saliency maps comparison metrics were used: Pearson Correlation Coefficient (PCC) and similarity(SIM). PCC measures statistical correlation between two maps as signals, SIM measures concordance between them as 2D distributions.
\emph{Pearson Correlation Coefficient (PCC)} is computed between an ExM and our available ground through, i.e. GFDM:
        \begin{equation}
        \label{eq:Corr}
        corr(x,y) = \frac{\sum_{(u,v)\in W\times H} \left(x(u,v) - \overline{x}\right) \left(y(u,v) - \overline{y}\right)}{\sqrt{\sum_{(u,v)\in W\times H} \left(x(u,v) - \overline{x}\right)^2} \sqrt{\sum_{(u,v)\in W\times H} \left(y(u,v) - \overline{y}\right)^2}}
        \end{equation}

\noindent $(u,v)$ correspond to the pixel coordinates in the map, $W$ and $H$ being the width and the height of the image respectively,  $\overline{a}$ corresponds to the mean over the values of $a$.\\

\emph{Similarity metric (SIM)} between the same maps is defined as:
        \begin{equation}
        \label{eq:Sim}
        sim(x,y) = \Sigma_{(u,v)\in W\times H} min(x(u,v),y(u,v))
        \end{equation}
    with $x$ the explanation map, $y$ the ground truth and $(u,v)$ being spatial coordinates.

    \subsection{No-reference-based metric: Stability}
    \label{subsection:Stability Metric}
    The stability metric for evaluation of black-box explanation methods was proposed in \cite{DBLP:journals/corr/abs-2102-13076} and is based on the \emph{Lipschitz constant}. Given an image space $X$, trained classifiers $C$ and explanation methods $D$, we consider the explainer function $f(x,c,d)$ as a mapping from $X \times C \times D$ to the metric space $E$ representing explanation maps. For the sake of simplicity, as we do not change the classifier and the explanation method, we will further denote an explanation map for the given image $x \in X$ by $e=f(x)$. \\
    
    The \emph{Lipschitz mapping}  increases the distance between arguments by no more than $L$ times, where $L$ is called the \emph{Lipschitz constant} of the mapping. 
    More formally, let us consider a metric space $X$ with the metric $p_X$ and a metric space $E$ with the metric $p_E$. A mapping $f$ of a metric space (X, $p_X$) to a metric space (E, $p_E$) is called Lipschitzian if there is such a constant $L$ (the Lipschitz constant of this mapping) that $p_E \left(f(x), f(x') \right) \leq L \cdot p_X(x,x')$ for any $x, x' \in X$. This condition is called the \emph{Lipschitz condition}. In the present work, we consider Euclidean metric for both spaces $X$ and $E$.\\
    
    According to the work of Bodria \cite{DBLP:journals/corr/abs-2102-13076}, the stability metric aims at validating how consistent the explanations are for similar inputs. The higher the value, the better is the explanation model to present similar explanations for similar inputs. Stability can be evaluated by exploiting the Lipschitz constant. In their work, Bodria et al. propose to compute the maximal Lipschitz constant in a certain neighbourhood $N_x$ of the given data point $x$:  
    
    \begin{equation}
     L_x = max \frac{\|{e_x-e_{x'}}\|}{\|{x-x'}\|} , \forall x' \in N_x  
    \label{eq:Lipschitz}
    \end{equation}\\
     where $x$ is the explained instance that is our image, $e_x$ the explanation map and  $x'$ are the data similar to our data $x$, $\|.\|$ is the Euclidean norm.
     The theoretical radius $r(N_x)$ of the neighbourhood $N_x$ in case of images is defined by the maximal value $Q$ for the given quantization of all colour components and of image size $W\times H$ with $W$ width and $H$ height of images. In the case of Euclidean metric, it is $r=\sqrt{n_c}\times Q\times \sqrt{W\times H}$. Here $n_c$, is the number of colour channels, $n_c=3$ for RGB images, and $Q=255$ for 8-bit quantization of colour channels.\\
     
     In practice, the Euclidean distance between original images and degraded ones $\|{x-x'}\|$ is smaller than the theoretical radius $r(N_x)$. Indeed, Euclidean norm of difference is computed as:
    
     \begin{equation}
     \|{x-x'}\| = \sqrt{\Sigma_{(u,v) \in W\times H}\left({\left(x_R(u,v)-x'_R(u,v)\right)}^2+{\left(x_G(u,v)-x'_G(u,v)\right)}^2+{\left(x_B(u,v)-x'_B(u,v)\right)}^2\right)} 
    \label{eq:EuclideanNorm}
    \end{equation}\\
    In each colour channel, the degraded pixel value is computed as described in Appendix \ref{section:Noisy_images}, with controlled parameters of each degradation. Therefore, with a high probability in a pixel $(u,v)$, the channel value difference $\Delta_{Ch}(u,v)=x_{Ch}(u,v)-x'_{Ch}(u,v)$ will be lower than $Q$ in its absolute value. Therefore, the original $x$ and degraded $x'$ images will be closer in image space $X$.\\

    The idea of using the Lipschitz constant is that if the image is corrupted by noise, then the classifier will be probably forced to look for other elements or objects in the image to classify it. To illustrate this let us consider the case of e.g. blur degradation of an image to be classified with a trained network, and one of its convolutional filters which was trained in an end-to-end training process to detect contours. Then a strong blur in the image to be classified at a generalization step can totally smooth the contour locally. Therefore, the feature maps of the blurred image will change, and the final classifier will take decision without using the area of blurred contour. Hence, if we apply a "good explainer" in such a case, then the explanation maps on original image and distorted one will differ. Therefore, when the norm of difference between the original image and the corrupted image grows, the norm of difference of explanation maps will grow also. Thus, Lipschitz constant should not increase accordingly to  equation~\ref{eq:Lipschitz}. Therefore, if the explanation method is stable, then with the growing noise, Lipschitz constant will show a stable behaviour. We will explore it on well classified images and badly classified (because of the noise) images. 
    This gives us several options:
        (i) the Lipschitz constant will show a drop (or no change), as well as stabilization with strong image noise, which is the correct result;
        (ii) the increase of the Lipschitz constant indicates an error in the method of explanation.

       \subsection{Methodology of Explainers Comparison}
    \label{subsection:Comparison Methodology}
    The comparison methodology comprises several preliminary stages before computing the evaluation metrics:
    \begin{enumerate}[\hspace{1cm}1.]
       \item 
     \emph{Generation} of $n$ noisy images with different noise levels from  each source image with controlled noise/distortions parameters. We used additive Gaussian noise, Gaussian blur, perspective deformations, uniform brightness distortions (appendix~\ref{section:Noisy_images} describe them);\\
        \item \emph{Classification} of the original and distorted images with the model to be explained;\\
        \item \emph{Application} of
       an explanation method to be evaluated to create an explanation map for each classified image;\\
        \item \emph{Clustering} of
      the classification data obtained into two groups: well-classified images (the labels of the original and noisy images are the same) and poorly classified (the labels did not match).
   \end{enumerate}
    
    We then compute three metrics: i) Lipschitz constant for comparison of original explanation map and the explanation map of noisy images accordingly to the equation \ref{eq:Lipschitz} (Stability metric), ii) PCC, iii) SIM.The mean value together with standard deviation are computed for each of three metrics, over each set of images: well classified and badly classified, as a function of the level of the noise. We track their behaviour as a function of distortions. Finally, we will analyse the agreement between these three metrics by computing the Pearson correlation coefficient between them.

    \section{Experimental Protocol}
\label{section:protocol}

    The evaluation metrics has been tested within a controlled protocol that implies the choice of a model, explainers, and database of images.

    \paragraph{\textbf{Evaluated Model}}
        \label{section:NN}
        The selected explainers are applied on ResNet50~\cite{rousseau:hal-01796729} for image classification, as it is the most popular CNN classifier for image classification tasks. 
        ResNet50 is trained, with the Adam optimizer \cite{article} and a  cross entropy loss on the SALICON dataset \cite{7298710}  with 20000 images split as 10000 for training, 5000 for validation and 5000 for test sets.



            \paragraph{\textbf{Evaluated Explainers}}
     The reference Grad-CAM \cite{DBLP:journals/corr/SelvarajuDVCPB16} as well as its new competitors FEM \cite{FEM_FuadMGBBZ20} and MLFEM \cite{DBLP:conf/icprai/BourrouxBBG22} have been applied to this network.
     These methods have been chosen because Grad-CAM is clearly a SOTA method widely used in the literature and FEM and MLFEM are interesting competitors that have been proven to outperform it. LRP, another SOTA method, has not been chosen because it is less ubiquitous due to the use of layer specific rules.
             ResNet50 contains 16 residual blocks: for MLFEM, which is a multi-layered explainer on the basis of FEM (see Appendix \ref{section:ExplanationMethods} for more detailed description of it) we apply FEM to the output of each residual block, after their activation function. This generates 16 different applications of FEM fused together using a had-hock fusion encoder trained to generate an explanation map from the 16 FEM maps.

    \paragraph{\textbf{Image dataset used for the evaluation}}
        \label{subsection:original_dataset}
        The image dataset contains 50 images of the SALICON dataset \cite{7298710}. We have retained this volume because of hardware constraints. 
        In the SALICON dataset, 10000 images are supplied  with GFDMs. SALICON is a subset of another dataset, MS COCO \cite{DBLP:journals/corr/LinMBHPRDZ14}. The latter contains images of 80 different categories of objects in context. It is common to have multiple categories present for each image.  The categories are not uniformly represented.
        To compute the GFDMs, the subjects have participated in psycho-visual experiment with free viewing conditions, that is they were invited to "look around" in the image and not searching for a particular object. In such kind of psycho-visual experiment, the "bottom up", stimuli driven, component of human visual attention is activated first, as the subject has no goal of a specific visual search. Nevertheless, the bottom-up attention is activated only during the first moments of observation of a visual scene. As far as the subject interprets the scene, top-down, task-driven attention will be predominant, which means that the subject foveates on semantic objects. This well-studied psycho-visual phenomenon was observed by us in \cite{ChaabouniBTAZ17}. Thus, the gaze fixations obtained in free-viewing conditions can be used in our opinion to build gaze fixation density maps as the ground truth of attention in visual recognition problems.\\ 
        
        Furthermore, the authors of \cite{7298710} did not record gaze fixations to define where the subject looks in the image. They have designed a gaze-contingent multi-resolution mechanism where subjects could move the mouse to direct the high-resolution fovea to where they find interesting in the image stimuli. The mouse trajectories from multiple subjects were aggregated to indicate where people look most in the images.
         The paradigm was first tested on a database of images created by imitating the eccentricity-based sensitivity of the Human Visual System to the contrast in the images. For these images, the gaze fixations recorded with eye-trackers were also available. All participants had normal or corrected-to-normal vision, and normal colour vision as assessed by Ishihara test. All subjects had not participated in any eye-tracking experiment or seen the presented images before. The images were presented to the subjects in 700 trials at random order. Each trial consists  of a 5-second image presentation followed by a 2-second waiting interval. The mouse cursor was displayed as a red circle with a radius of 2 degrees of visual field that is sufficiently large not to block the high-resolution region of focus, and automatically moved to the image centre when the image onset. The subjects were instructed to explore the image freely by moving the mouse cursor to anywhere they wanted to look. Once validated, the protocol was repeated in a crowdsourcing scheme of Amazon Mechanical Turk (AMT) and 10,000 MS COCO images viewed by 60 observers each were supplied with GFDMs. 
        In the 50 images we retained from this dataset the presence of objects of more than one category in a source image is possible. Furthermore, all these 50 source images were correctly classified by our ResNet50 classifier. 

\paragraph{\textbf{Distorted image dataset}}
        \label{subsection:distorted_dataset}
        For each image from original dataset and for each of considered degradation (Additive Gaussian Noise, Gaussian Blur, Uniform Brightness shift, Perspective distortion, all explained in appendix \ref{section:Noisy_images}) we have generated 40 distorted images. Thus, the whole image set for one degradation was 2000. 
        Results of experiments  are individually analysed for each distortion.\\

         \emph{For the additive Gaussian noise}:
    (i) for each of 50 original images, we set eight maximal shift values ($k$): [25..200] with a step of 25. These values give 95\% of noise values accordingly to the two sigma rule (appendix \ref{subsection:GaussianNoise}). For each $k$ we generate five shift values which are different for each pixel and applied for each colour channel. Each pixel is considered as i.i.d process and the noise values are generated individually. 
         (ii) the number of generated noisy images for each $k$ value is a parameter of our method $M$ ($M = 5$). 
        The number of corrupted images is thus 2000. \\

\emph{For the Gaussian blur}: 
    (i) for each of 50 original images we generate a corrupted image for each mask of size: (5x5), (7x7), (9x9), (11x11) (appendix \ref{subsection:GaussianBlur}) with  scale parameter $\sigma_{gb}$ values from [$1.25$, $1.5$, $1.75$, $2$, $2.5$, $3$, $3.5$, $4$, $5$, $6$].
        (ii) The number of generated noisy images for each $\sigma_{gb}$ value was taken as ($M = 4$) thus totalling 2000 corrupted images. \\

\emph{For the uniform brightness distortion}: (i) for each of 50 original images, we set eight maximal shift values which give 95\% of noise values accordingly to the two sigma rule (appendix \ref{subsection:uniforBrightness}) ($\beta$): [25..200] with a step of 25. Then, the random shift is applied to all pixel values in the image. 
       (ii) The number of generated noisy images for each $\beta$ value is ($M =50$) totalling 2000 corrupted images.\\

\emph{For the perspective distortion}: 
     (i) for each of 50 original images we generate a 
                corrupted image for each direction of the narrow part of 
                the distortion trapezoid: (top), (bottom), (left), (right) (appendix \ref{subsection:PerspectiveDistorsion}) putting each image at the scale 
                $l$: $[1, 2, 3, 4, 5, 6, 7, 8, 9, 10]$;
      (ii) the number of generated noisy images for each $l$ value is a parameter of our method ($M =4$). 
           The total number of distorted images is thus 2000.

\section{Results}
\label{section:results}

    This section presents the results of the experiments.  The three considered explanation methods, Grad-CAM, FEM and MLFEM, have been already evaluated in \cite{DBLP:conf/icprai/BourrouxBBG22} with reference-based Similarity and Pearson correlation coefficient metrics. Here, we compute these metrics also to assess the behaviour - agreement or not of a non-reference stability metric (Lipschitz constant) with them on a selected dataset. 
    The results are presented for each degradation.

         \begin{table}[H]
        \centering
        \caption{Pearson Correlation Coefficient value between different metrics: L-stability, PCC - Pearson Correlation Coefficient with GFDMs, SIM - similarity with GFDMs}
        \subtable[Gaussian noise\label{tab:correlation_of_metrics_Gaussian_noise}]{
            \scalebox{0.60}{ 
            \setlength{\tabcolsep}{3.0mm}
            \begin{tabular}{c|c|c|c}
                \hline
                {\textbf{}} & {\textbf{L -$>$ PCC}} & {\textbf{L -$>$ SIM}} & \textbf{PCC -$>$ SIM}\\
                \hline 
                \hline
                \textit{Well FEM}  & \textbf{0.73064026} & 0.76240769  & \textbf{0.99000345} \\[5pt]
                \textit{Well MLFEM}  & 0.7297274 & \textbf{0.78237337}  & 0.90253736 \\[5pt]
                \textit{Well GRAD-CAM} & 0.55420942 & 0.74570872 & 0.70806073\\[5pt]
                \textit{Badly FEM} & 0.52068572 & 0.60428322  & 0.9497896 \\ [5pt]
                \textit{Badly MLFEM} & 0.64827606  & 0.68642392  & 0.76710794 \\ [5pt]
                \textit{Badly GRAD-CAM} & 0.21625147 & 0.5458757  & 0.43491874 \\ [5pt]
                \hline
            \end{tabular}}
        }%
    \subtable[Gaussian blur\label{tab:correlation_of_metrics_Gaussian_blur}]{
            \scalebox{0.60}{
            \setlength{\tabcolsep}{3.0mm}
            \begin{tabular}{c|c|c|c}
                \hline
                {\textbf{}} & {\textbf{L -$>$ PCC}} & {\textbf{L -$>$ SIM}} & \textbf{PCC -$>$ SIM}\\
                \hline 
                \hline
                \textit{Well FEM}  & 0.66402961 & 0.75673739  & \textbf{0.98377656} \\[5pt]
                \textit{Well MLFEM}  & \textbf{0.74303627} & \textbf{0.86852772}  & 0.90567752 \\[5pt]
                \textit{Well GRAD-CAM} & 0.48042003 & 0.66826275 & 0.73623874\\[5pt]
                \textit{Badly FEM} & 0.65029179 & 0.73264899  & 0.98267173 \\ [5pt]
                \textit{Badly MLFEM} & 0.73262894  & 0.84587245  & 0.87830155 \\ [5pt]
                \textit{Badly GRAD-CAM} & 0.33506892 & 0.79015328  & 0.67772217 \\ [5pt]
                \hline
            \end{tabular}}
        }

        \subtable[Uniform Brightness Distortion \label{tab:correlation_of_metrics_Uniform_Brightness_Distortion}]{

            \scalebox{0.60}{
            \setlength{\tabcolsep}{3.0mm}
            \begin{tabular}{c|c|c|c}
                \hline
                {\textbf{}} & {\textbf{L -$>$ PCC}} & {\textbf{L -$>$ SIM}} & \textbf{PCC -$>$ SIM}\\
                \hline 
                \hline
                \textit{Well FEM}  & 0.19982932 & 0.24093856  & 0.98639818 \\[5pt]
                \textit{Well MLFEM}  & 0.20139727 & 0.29660534  & 0.88265143 \\[5pt]
                \textit{Well GRAD-CAM} & -0.0096461 & 0.1950881 & 0.6507635\\[5pt]
                \textit{Badly FEM}  & 0.65728164 & 0.66616354  & \textbf{0.99851441} \\ [5pt]
                \textit{Badly MLFEM} & \textbf{0.77474594}  & \textbf{0.77519272}  & 0.97498555 \\ [5pt]
                \textit{Badly GRAD-CAM} & 0.36310712 & 0.67503664  & 0.79365511 \\ [5pt]
                \hline
            \end{tabular}}
      
  }%
    \subtable[Perspective Distortion\label{tab:correlation_of_metrics_Perspective_Distortion}]{
            \scalebox{0.60}{
            \setlength{\tabcolsep}{3.0mm}
            \begin{tabular}{c|c|c|c}
                \hline
                {\textbf{}} & {\textbf{L -$>$ PCC}} & {\textbf{L -$>$ SIM}} & \textbf{PCC -$>$ SIM}\\
                \hline 
                \hline
                \textit{Well FEM}  & 0.48519926 & 0.63057759  & 0.96087489 \\[5pt]
                \textit{Well MLFEM}   & 0.52108631 & 0.78461745  & 0.69494603 \\[5pt]
                \textit{Well GRAD-CAM} & 0.22590666 & 0.53638187 & 0.46277711\\[5pt]
                \textit{Badly FEM}  & \textbf{0.82879462} & 0.86541922  & \textbf{0.9896904} \\ [5pt]
                \textit{Badly MLFEM} & 0.79858929  & \textbf{0.93149344}  &  0.90147702 \\ [5pt]
                \textit{Badly GRAD-CAM} & 0.51340872 & 0.91090457  & 0.66553841 \\ [5pt]
                \hline
            \end{tabular}}
        }

    \end{table}   
        

        \paragraph{\textbf{Additive Gaussian noise}}
        \label{subsection:Experiment_Additive Gaussian noise}
        
\begin{figure}[H]
      \centering
        \subfigure []{
            \label{fig:subfig:figs_Gaussian_noise:L_fem_fig} 
            \includegraphics[scale=0.040]{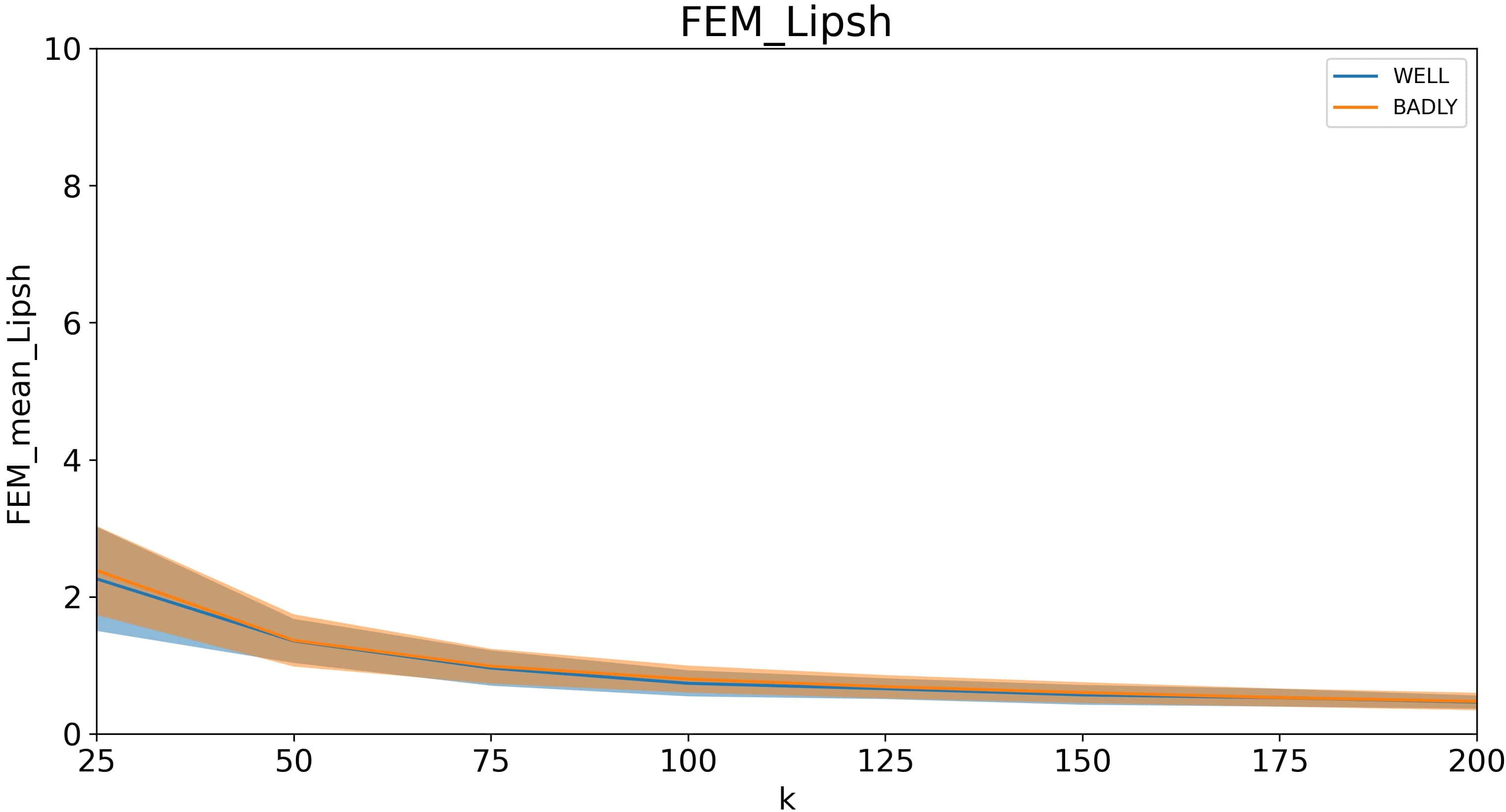}}
        \subfigure []{
            \label{fig:subfig:figs_Gaussian_noise:L_mlfem_fig} 
            \includegraphics[scale=0.040]{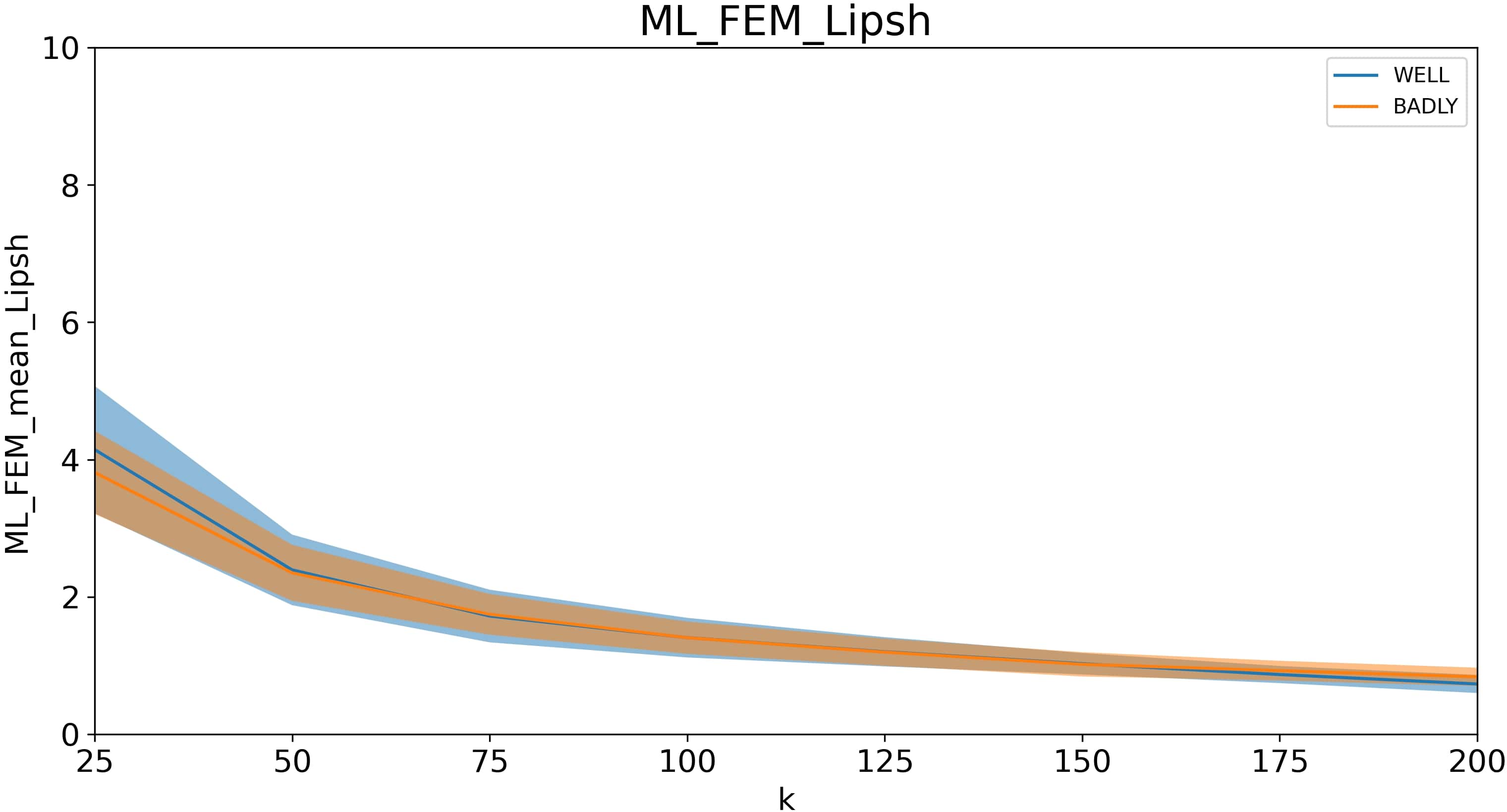}}
        \subfigure []{
            \label{fig:subfig:figs_Gaussian_noise:L_gradcam_fig} 
            \includegraphics[scale=0.040]{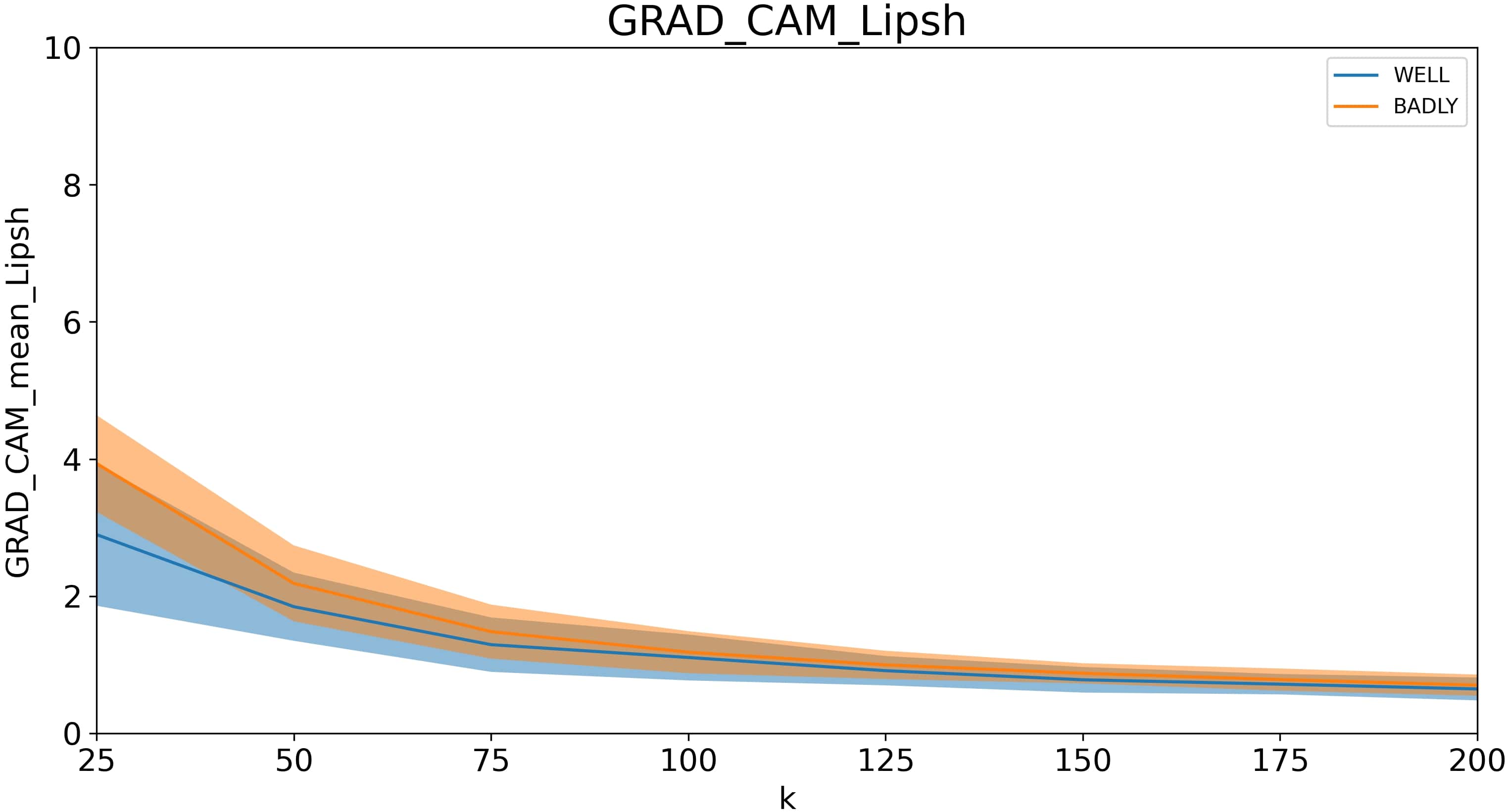}}
        \subfigure []{
            \label{fig:subfig:figs_Gaussian_noise:PCC_fem_fig} 
            \includegraphics[scale=0.040]{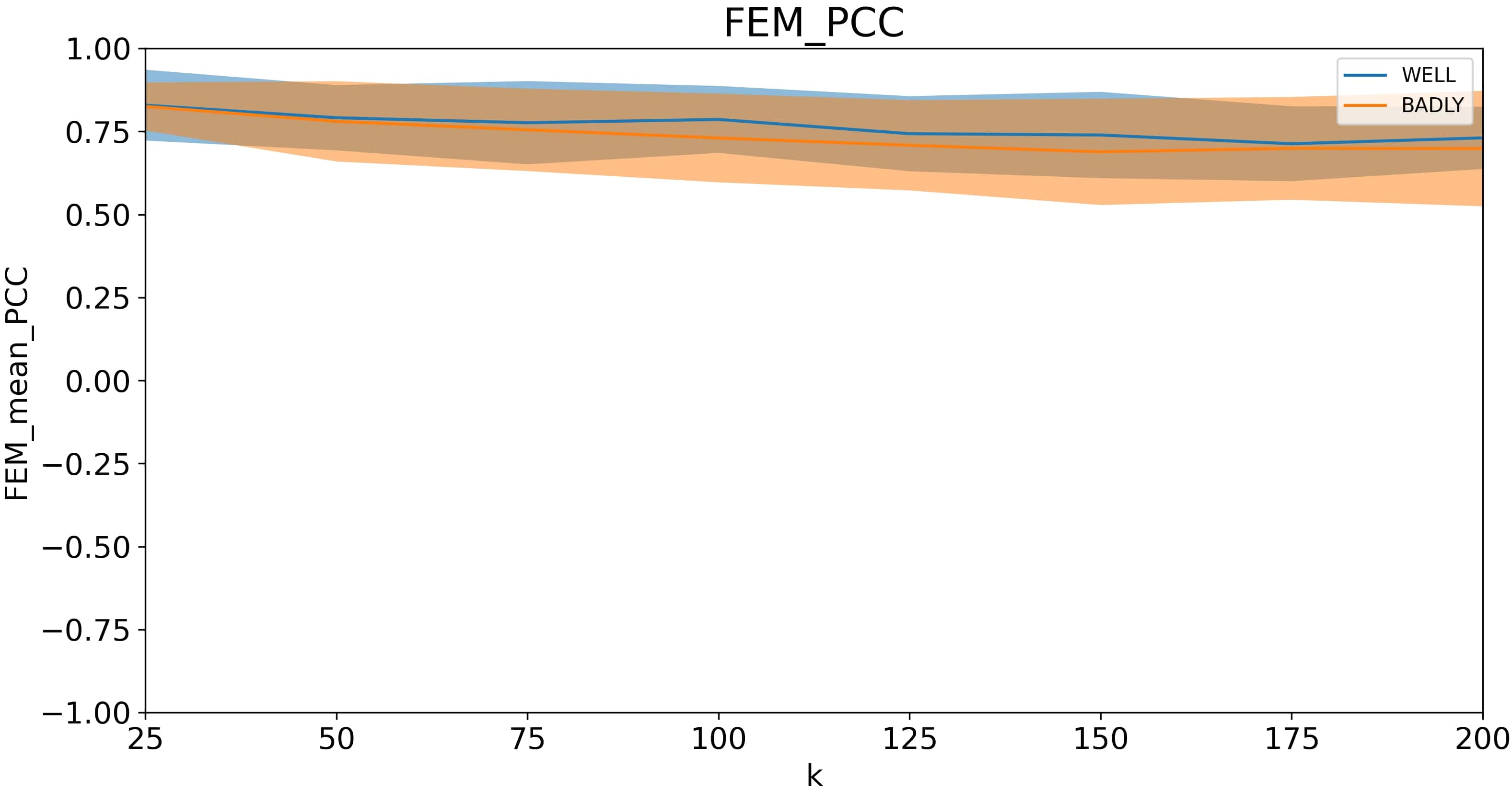}}
        \subfigure []{
            \label{fig:subfig:figs_Gaussian_noise:PCC_mlfem_fig} 
            \includegraphics[scale=0.040]{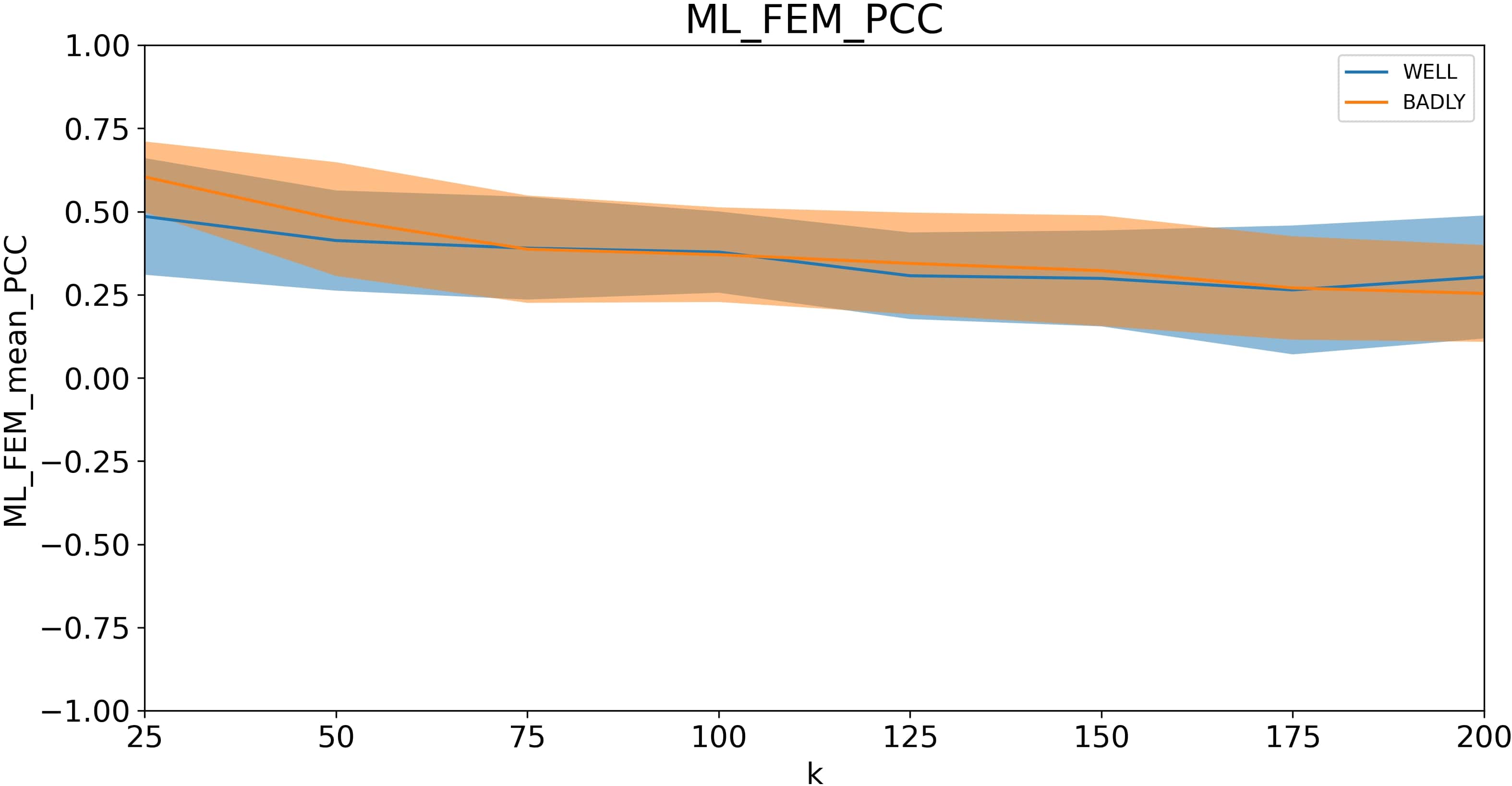}}
        \subfigure []{
            \label{fig:subfig:figs_Gaussian_noise:PCC_gradcam_fig} 
            \includegraphics[scale=0.040]{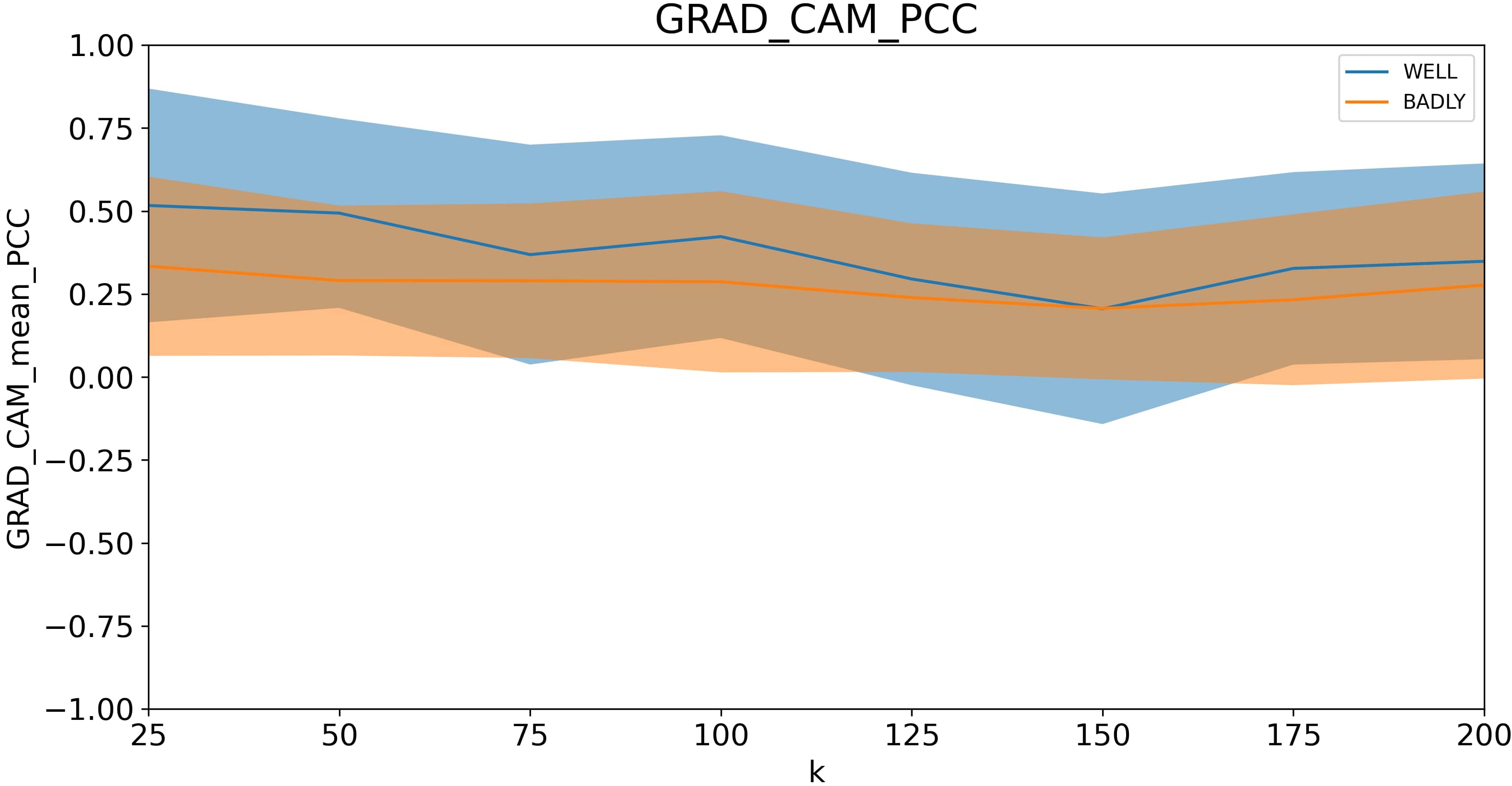}}
        \subfigure []{
            \label{fig:subfig:figs_Gaussian_noise:SIM_fem_fig} 
            \includegraphics[scale=0.040]{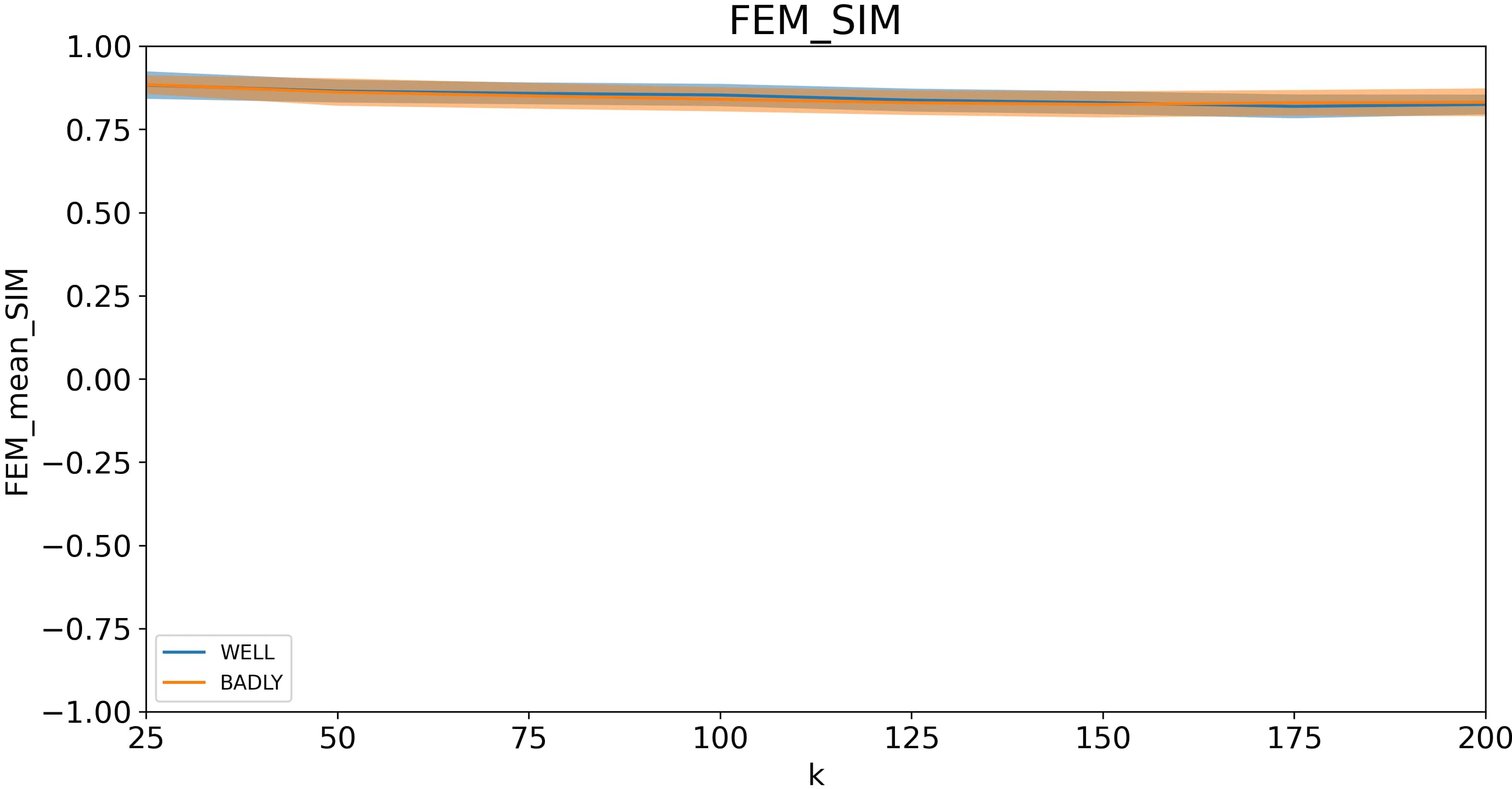}}
        \subfigure []{
            \label{fig:subfig:figs_Gaussian_noise:SIM_mlfem_fig} 
            \includegraphics[scale=0.040]{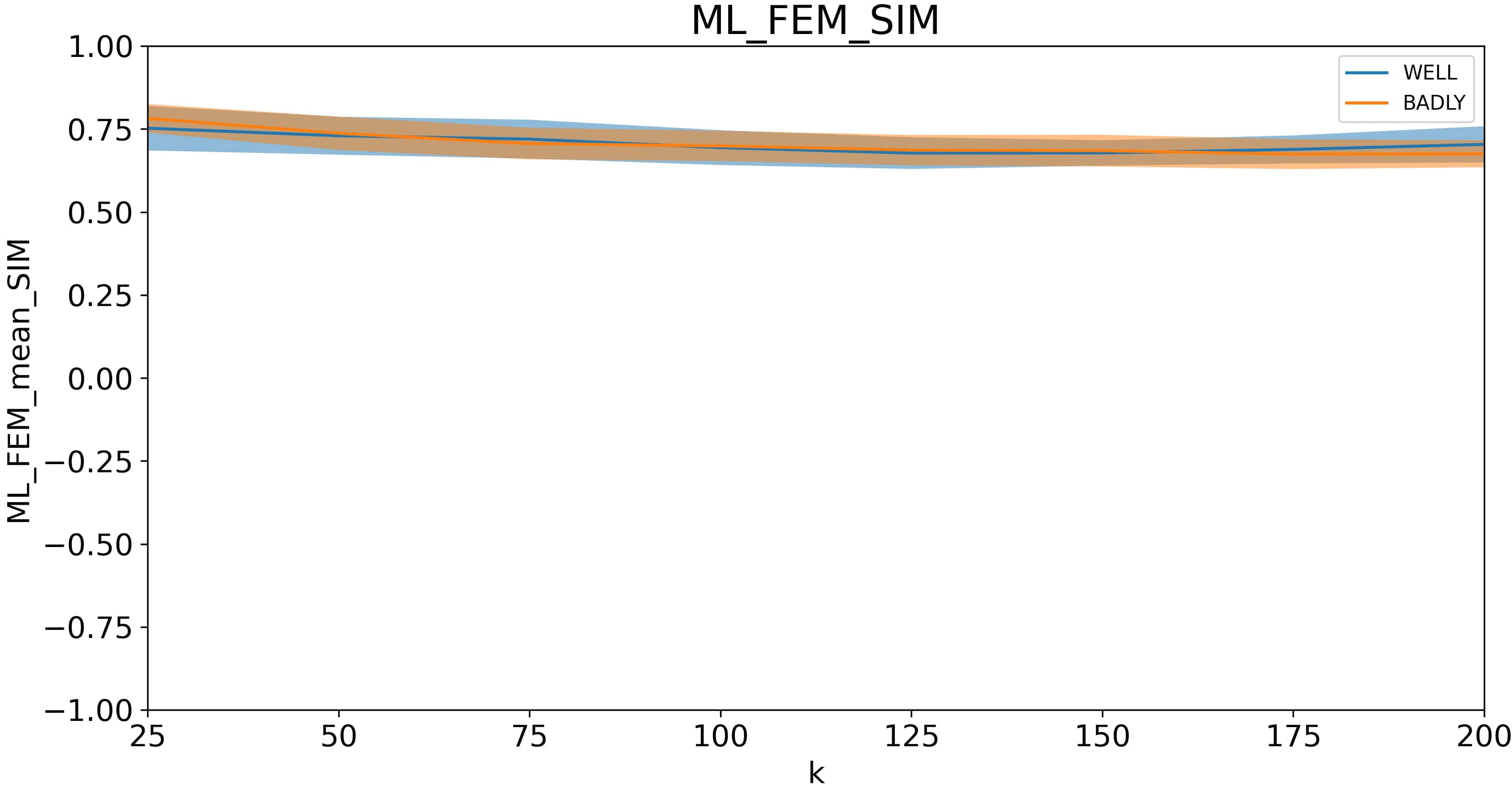}}
        \subfigure []{
            \label{fig:subfig:figs_Gaussian_noise:SIM_gradcam_fig} 
            \includegraphics[scale=0.040]{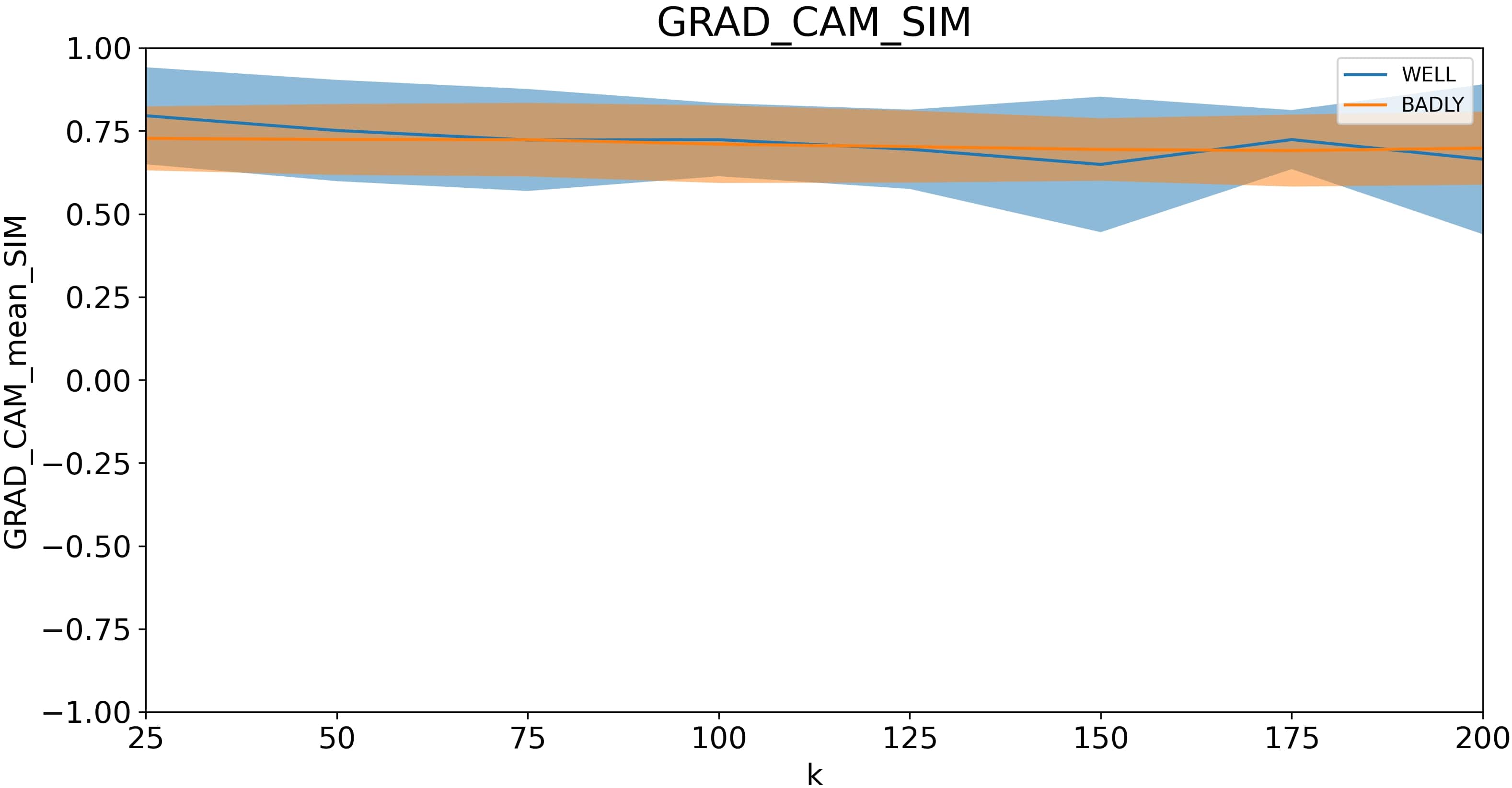}}
         \caption {Gaussian Noise: Behaviour of Lipschitz constant, PCC and SIM measures as a function of noise level: (a)FEM-Lipschitz, (b)MLFEM-Lipschitz, (c)GRAD-CAM-Lipschitz, (d)FEM-PCC, (e)MLFEM-PCC, (f)GRAD-CAM-PCC, (g)FEM-SIM, (h)MLFEM-SIM, (i)GRAD-CAM-SIM}
      \label{fig:figs_Gaussian_noise} 
    \end{figure}

              The behaviour of the stability metric is expressed as the mean value and standard deviation of Lipschitz constant measured on 250 images for each value of parameterized maximal shift $k$.
               It is illustrated, for three explanation methods, in figures  \ref{fig:subfig:figs_Gaussian_noise:L_fem_fig}, \ref{fig:subfig:figs_Gaussian_noise:L_mlfem_fig}, \ref{fig:subfig:figs_Gaussian_noise:L_gradcam_fig}. We plot the mean value of $L$ over image set corrupted with Gaussian noise as a function of the level of the noise $k$.
              We can state that, generally, with the increase of noise level Lipschitz constant for all the methods stabilizes, this confirms our hypothesis that higher is the noise, greater is the distance between explanation maps. 
              Furthermore, FEM method is the best in this sense as it is very much stable across generated dataset. It's $\pm\sigma$ interval for any $k$ is tighter than for two other methods. The behaviour is similar on well classified and badly classified images. \\

              We also studied the stability of the Lipschitz constant,  as a function of the noise level. To do this, we compute $s = \frac{|L_{k(j)} - L_{k(j+1)}|}{L_k} \cdot 100\%, j = 1,...,J.$ $J$ is the index of the final distortion level. We note that the best stabilization for Well classified images ($s = 7.242\%$) is observed for FEM and for Badly classified images ($s = 9.046\%$) is observed for MLFEM method ($k:150\rightarrow175$), see table \ref{tab:L_Mean_Sigma_tab_Gaussian_noise}. We nevertheless stress that for high levels of noise $k=175,...,200$, the saturation effects are stronger in the corrupted images, but still for the intermediate levels of noise MLFEM and FEM stability remains better than that one of GRAD-CAM. 
              A similar analysis is done for PCC and SIM. 
              Their behaviour, see figures  \ref{fig:subfig:figs_Gaussian_noise:PCC_fem_fig}, \ref{fig:subfig:figs_Gaussian_noise:PCC_mlfem_fig}, \ref{fig:subfig:figs_Gaussian_noise:PCC_gradcam_fig} and  \ref{fig:subfig:figs_Gaussian_noise:SIM_fem_fig}, \ref{fig:subfig:figs_Gaussian_noise:SIM_mlfem_fig}, \ref{fig:subfig:figs_Gaussian_noise:SIM_gradcam_fig}, is stable for different levels of noise. 
              Their stability studies are presented in \ref{tab:PCC_Mean_Sigma_tab_Gaussian_noise} for PCC, and in table  for SIM, \ref{tab:SIM_Mean_Sigma_tab_Gaussian_noise} as a function of the noise level. The behaviour is similar for well-classified and badly classified images. Hence, we can conclude that these metrics are less sensitive to the noise level, and thus to the slight differences in explanation maps and GFDMs. \\
              
              Consensus of metrics measured by Pearson correlation coefficient between them is presented  in the table \ref{tab:correlation_of_metrics_Gaussian_noise}. It can be seen that, the no-reference stability metric demonstrates consensus with referenced-based metrics. Thus, \textit{the Lipschitz constant can be used to determine the quality of explainers even in the absence of ground truth (GFDMs)}

        \paragraph{\textbf{Gaussian Blur}}
        \label{subsection:Experiment_Gaussian Blur}
        
\begin{figure}[H]
      \centering
        \subfigure []{
            \label{fig:subfig:figs_Gaussian_blur:L_fem_fig} 
            \includegraphics[scale=0.040]{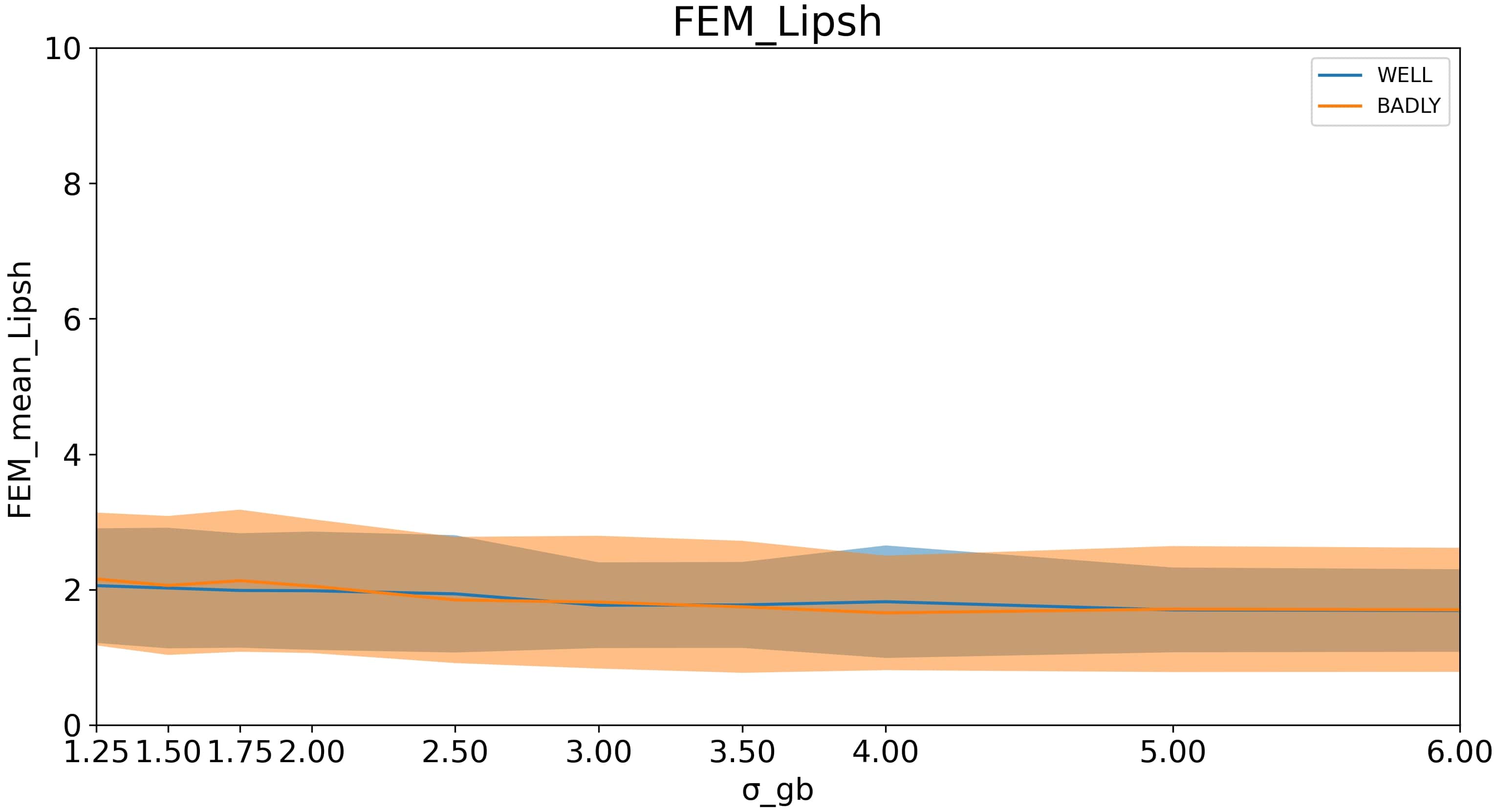}}
        \subfigure []{
            \label{fig:subfig:figs_Gaussian_blur:L_mlfem_fig} 
            \includegraphics[scale=0.040]{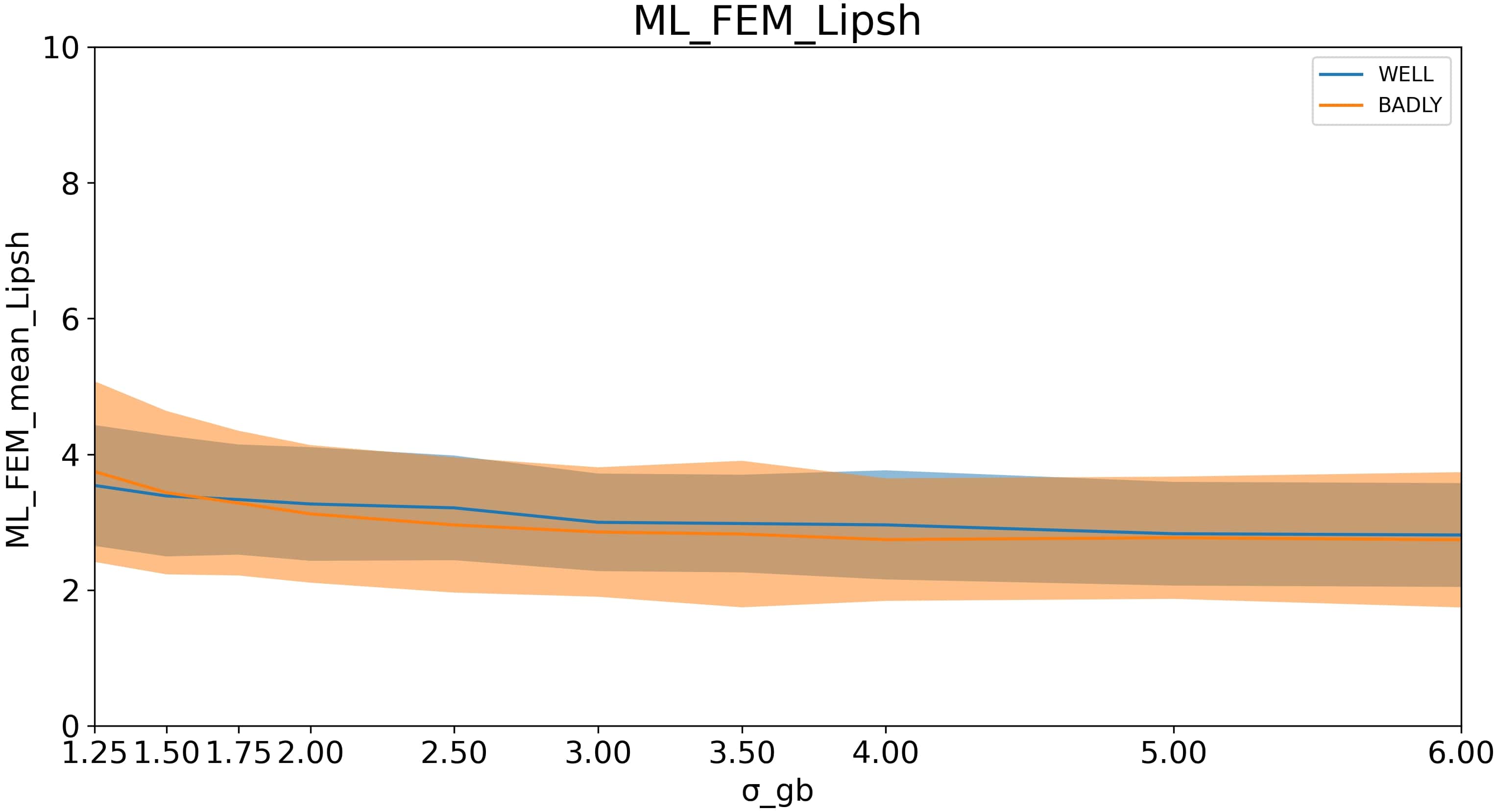}}
        \subfigure []{
            \label{fig:subfig:figs_Gaussian_blur:L_gradcam_fig} 
            \includegraphics[scale=0.040]{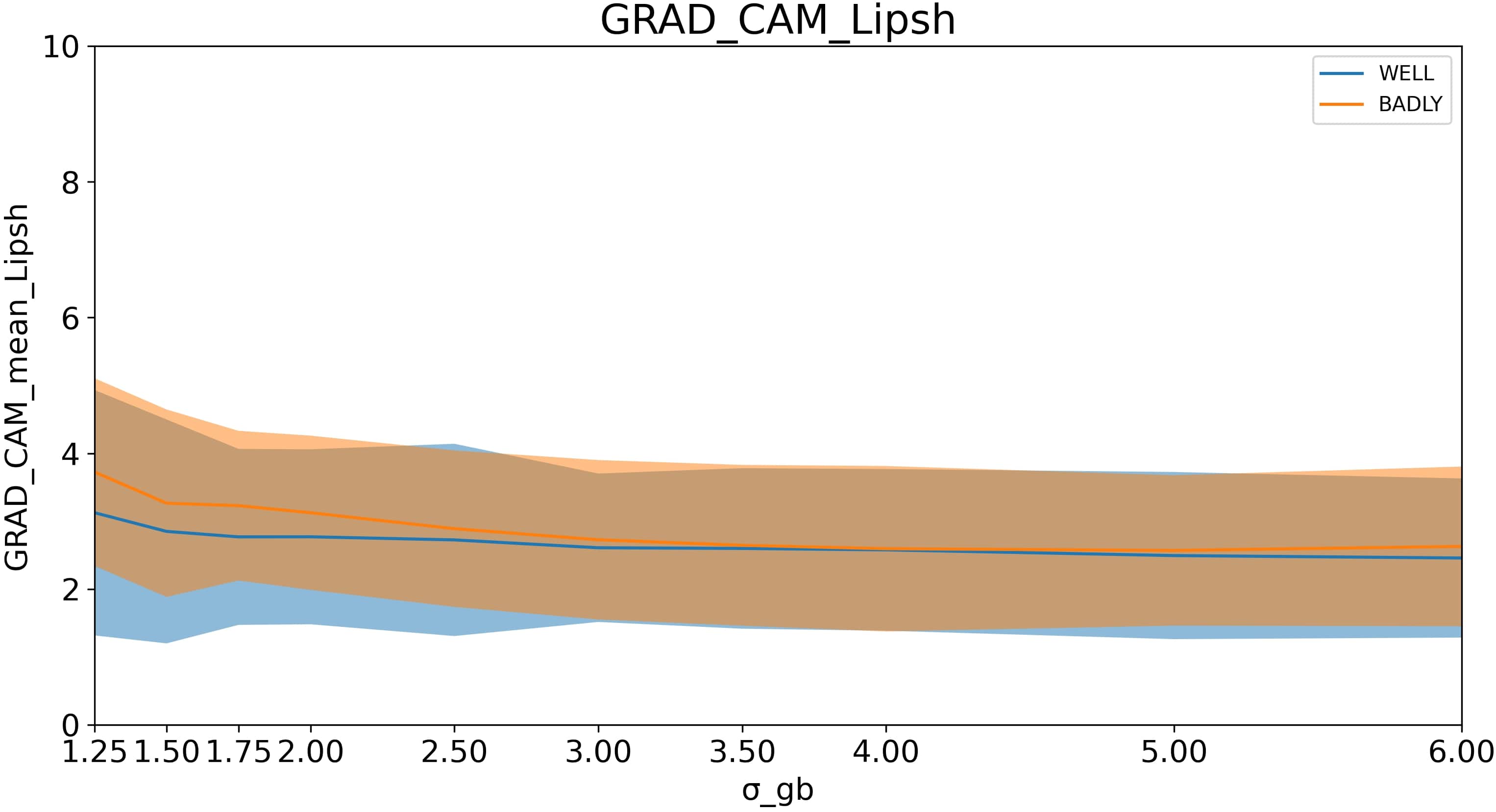}}
        \subfigure []{
            \label{fig:subfig:figs_Gaussian_blur:PCC_fem_fig} 
            \includegraphics[scale=0.040]{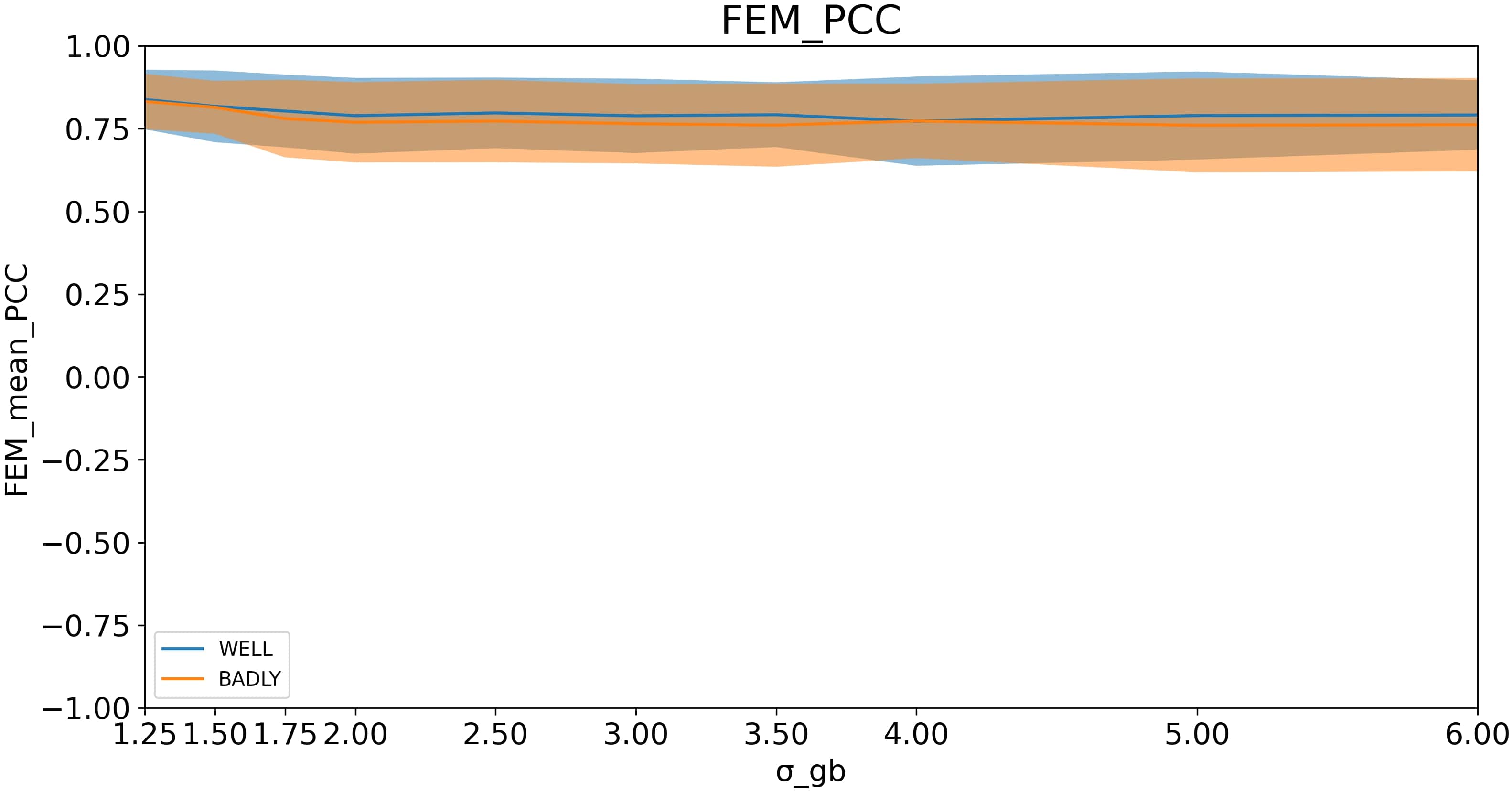}}
        \subfigure []{
            \label{fig:subfig:figs_Gaussian_blur:PCC_mlfem_fig} 
            \includegraphics[scale=0.040]{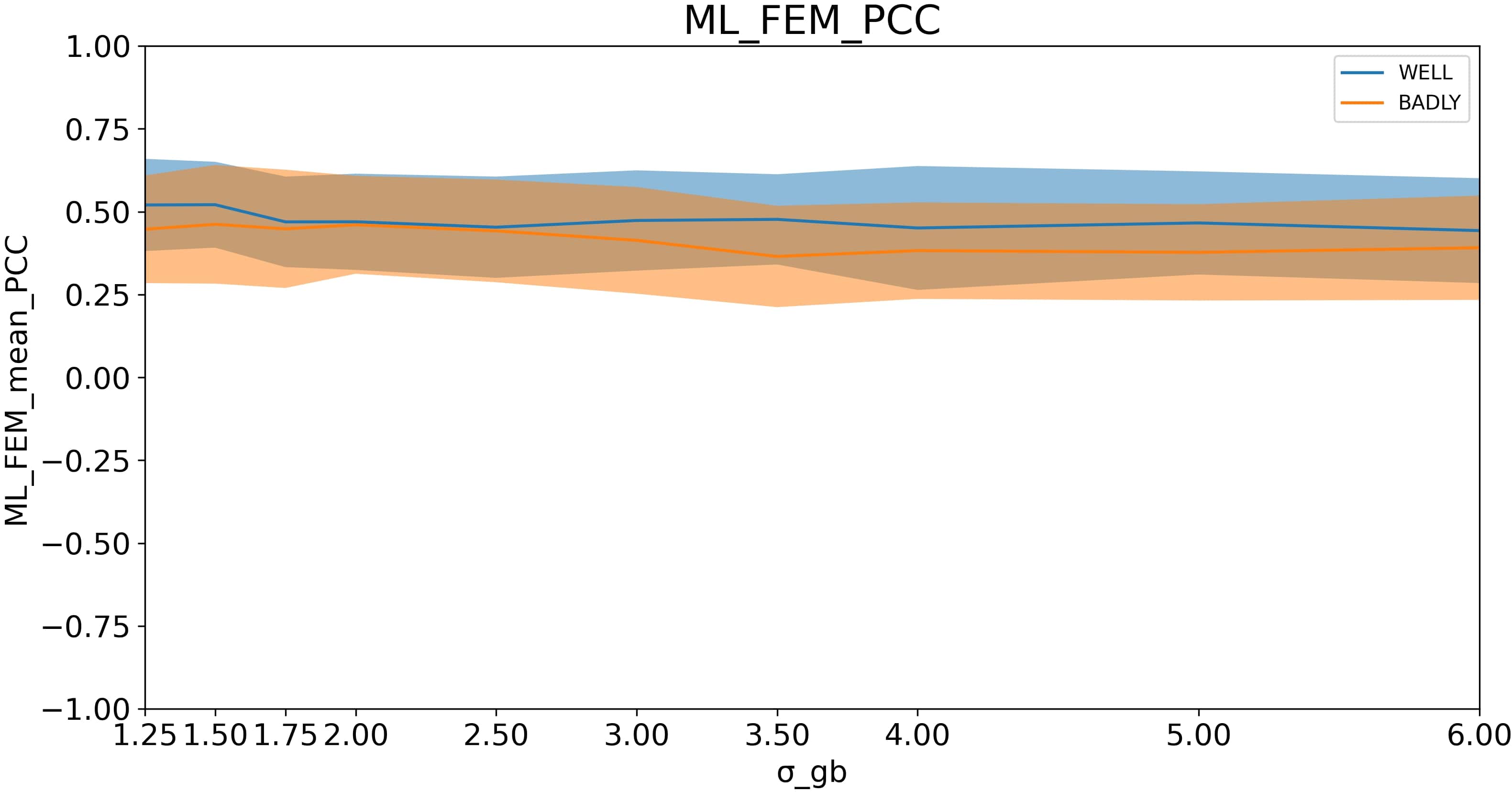}}
        \subfigure []{
            \label{fig:subfig:figs_Gaussian_blur:PCC_gradcam_fig} 
            \includegraphics[scale=0.040]{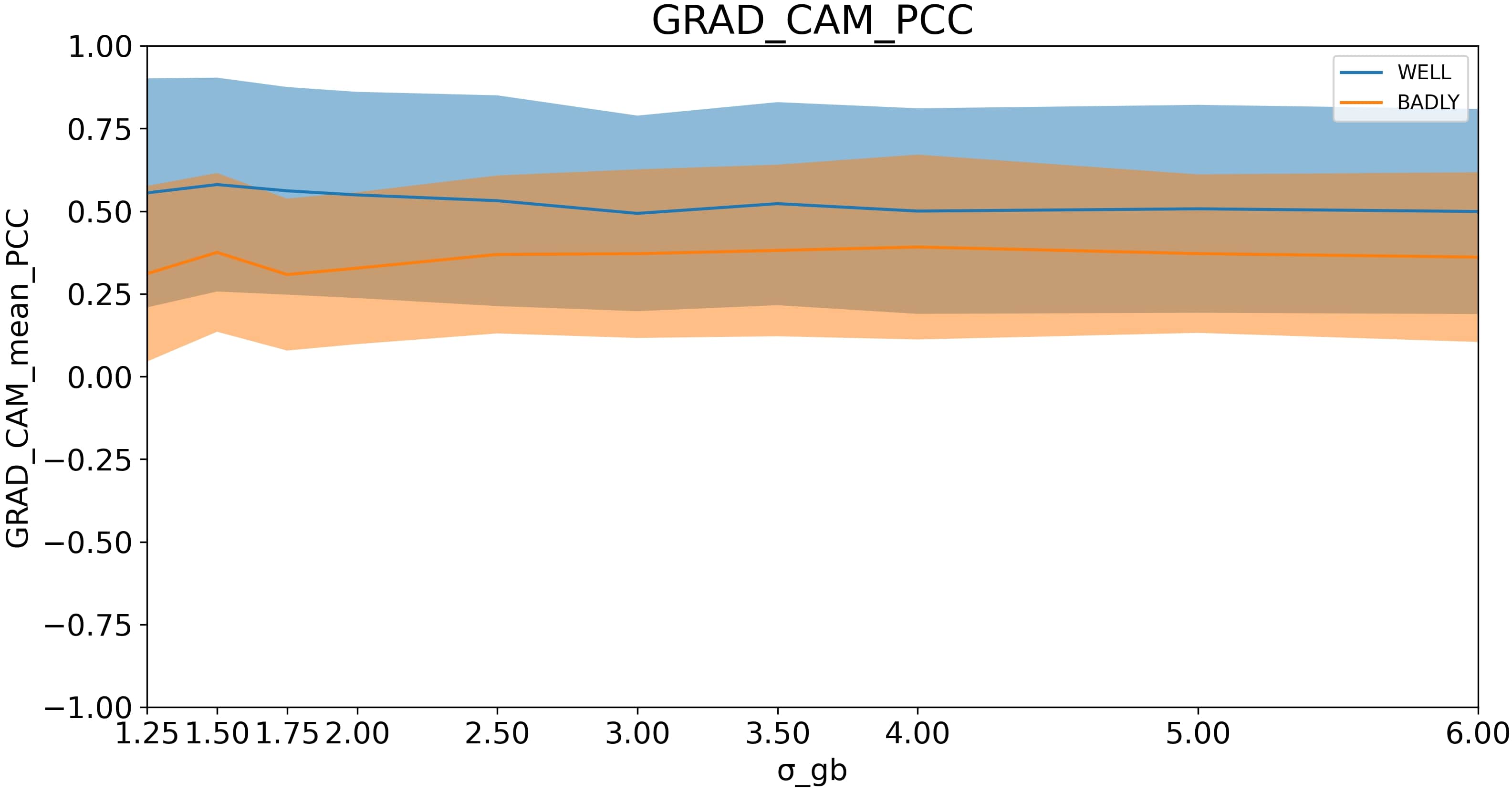}}
        \subfigure []{
            \label{fig:subfig:figs_Gaussian_blur:SIM_fem_fig} 
            \includegraphics[scale=0.040]{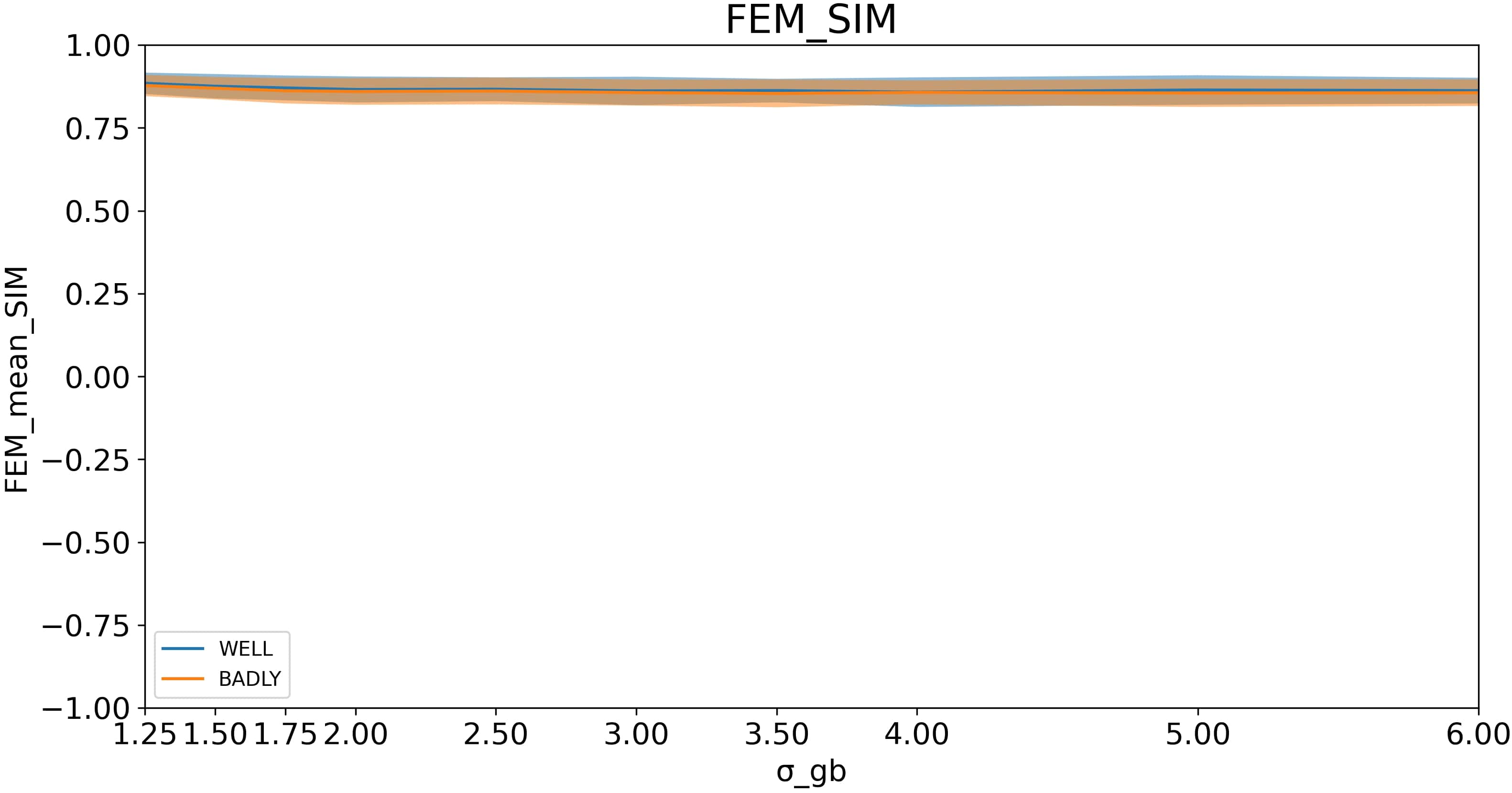}}
        \subfigure []{
            \label{fig:subfig:figs_Gaussian_blur:SIM_mlfem_fig} 
            \includegraphics[scale=0.040]{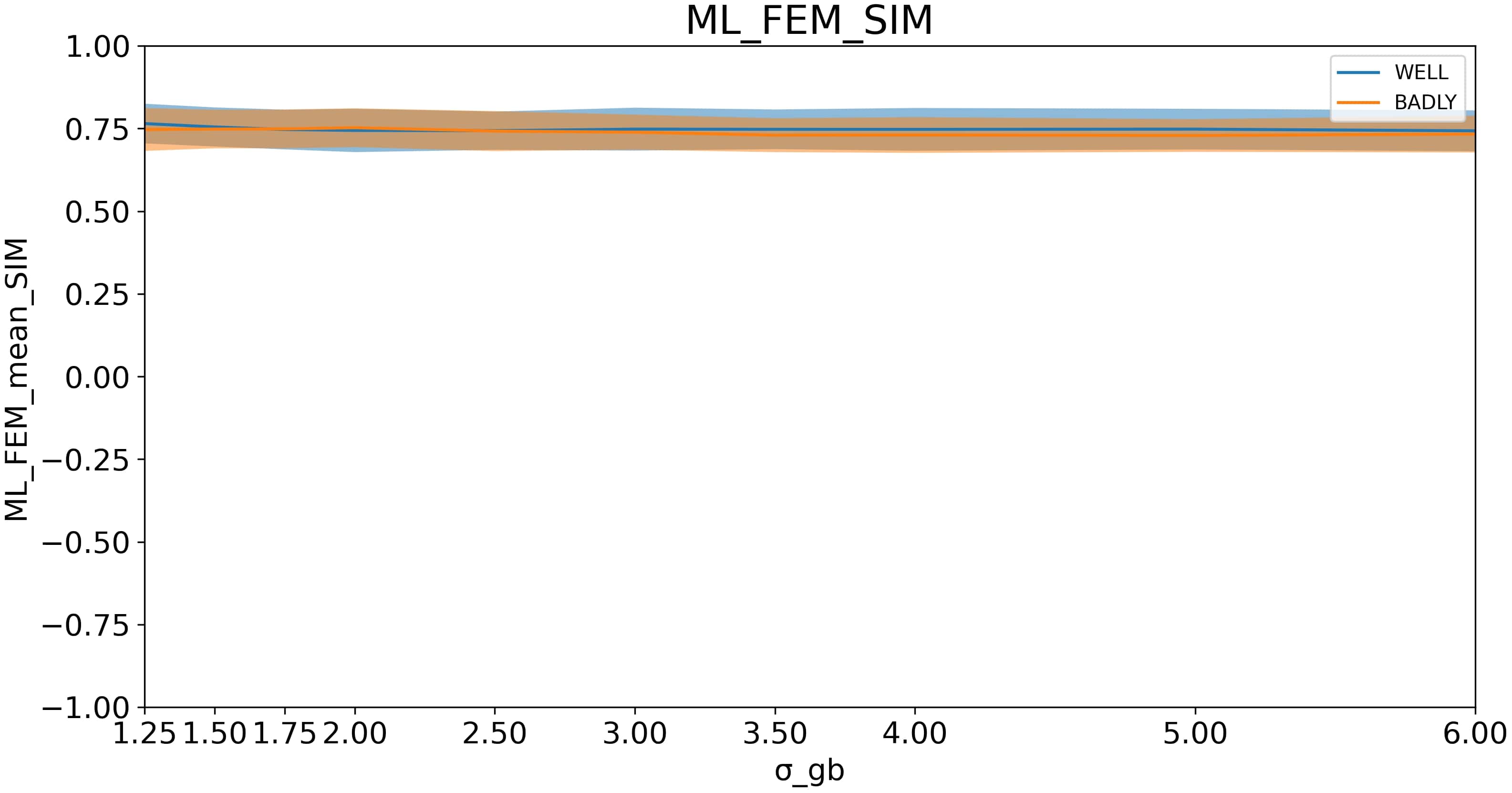}}
        \subfigure []{
            \label{fig:subfig:figs_Gaussian_blur:SIM_gradcam_fig} 
            \includegraphics[scale=0.040]{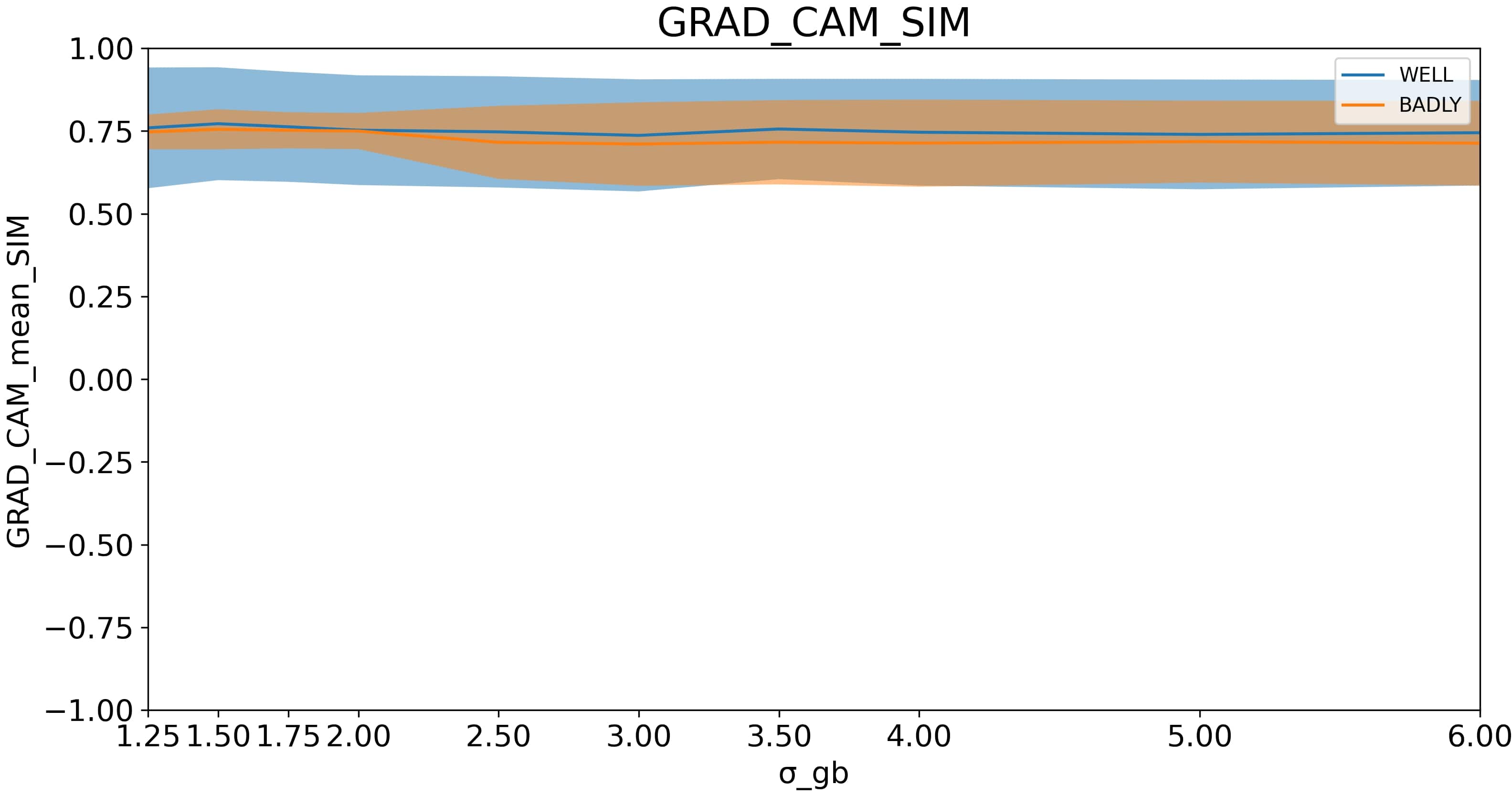}}
         \caption {Gaussian Blur: Behaviour of Lipschitz constant, of PCC and of SIM measures as a function of noise level: (a)FEM-Lipschitz, (b)MLFEM-Lipschitz, (c)GRAD-CAM-Lipschitz, (d)FEM-PCC, (e)MLFEM-PCC, (f)GRAD-CAM-PCC, (g)FEM-SIM, (h)MLFEM-SIM, (i)GRAD-CAM-SIM}
      \label{fig:figs_Gaussian_blur} 
    \end{figure}

    The behaviour of explanation methods is illustrated in figures \ref{fig:subfig:figs_Gaussian_blur:L_fem_fig}, \ref{fig:subfig:figs_Gaussian_blur:L_mlfem_fig}, \ref{fig:subfig:figs_Gaussian_blur:L_gradcam_fig} in terms of stability metric $L$, in figures
            \ref{fig:subfig:figs_Gaussian_blur:PCC_fem_fig}, \ref{fig:subfig:figs_Gaussian_blur:PCC_mlfem_fig}, \ref{fig:subfig:figs_Gaussian_blur:PCC_gradcam_fig} in terms of PCC and in figures
                        \ref{fig:subfig:figs_Gaussian_blur:SIM_fem_fig}, \ref{fig:subfig:figs_Gaussian_blur:SIM_mlfem_fig}, \ref{fig:subfig:figs_Gaussian_blur:SIM_gradcam_fig} in terms of SIM. Gaussian blur shows similar behaviour compared to Gaussian noise, but a wider range of standard deviation can be noticed. 
            FEM algorithm demonstrates the best stability results for the Lipschitz constant over the distorted database, see table \ref{tab:L_Mean_Sigma_tab_Gaussian_blur}. For comparison, in tables \ref{tab:PCC_Mean_Sigma_tab_Gaussian_blur} and \ref{tab:SIM_Mean_Sigma_tab_Gaussian_blur} figures for PCC and SIM metrics are given respectively. \\

             Consensus of metrics is presented in the table \ref{tab:correlation_of_metrics_Gaussian_blur}. Based on the data obtained, it can be concluded that in the case of Gaussian blur, the no-reference stability metric demonstrates consensus with SIM. Therefore, the Lipschitz constant can be used to determine the quality of explainers.

        \paragraph{\textbf{Uniform Brightness Distortion}}
        \label{subsection:Experiment_Uniform Brightness Distortion}
        
\begin{figure}[H]
      \centering
        \subfigure []{
            \label{fig:subfig:figs_Uniform_Brightness_Distortion:L_fem_fig} 
            \includegraphics[scale=0.040]{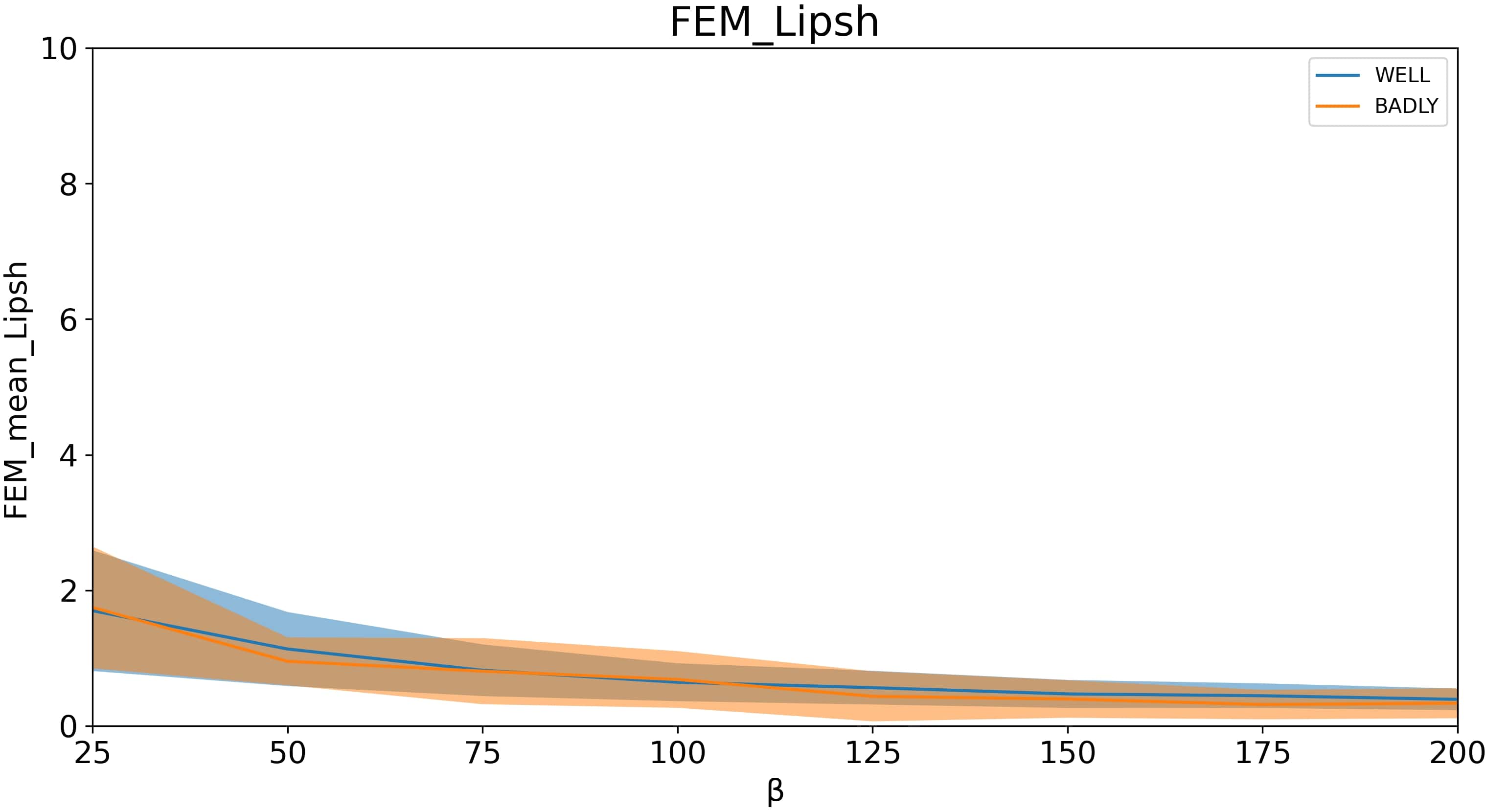}}
        \subfigure []{
            \label{fig:subfig:figs_Uniform_Brightness_Distortion:L_mlfem_fig} 
            \includegraphics[scale=0.040]{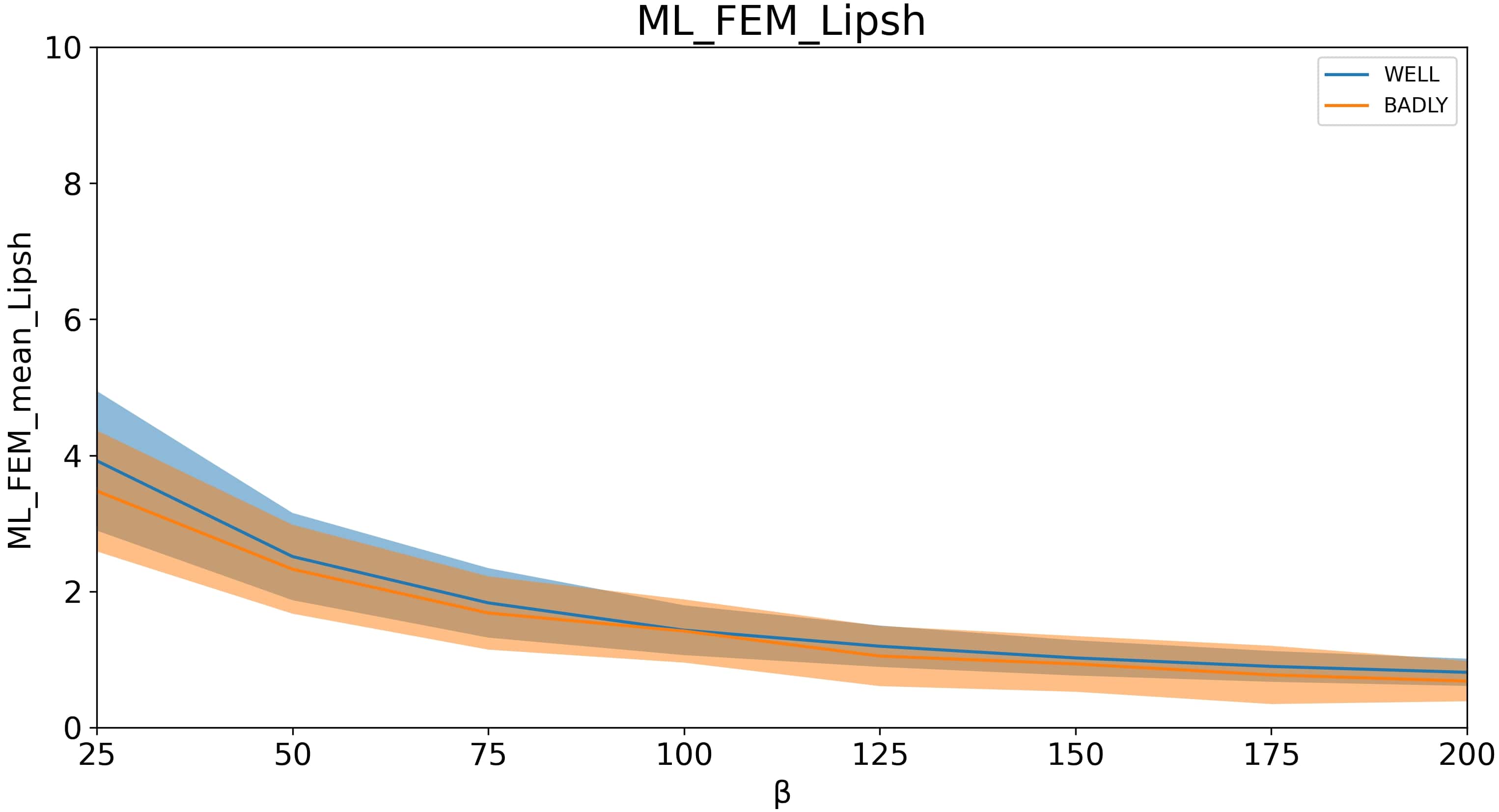}}
        \subfigure []{
            \label{fig:subfig:figs_Uniform_Brightness_Distortion:L_gradcam_fig} 
            \includegraphics[scale=0.040]{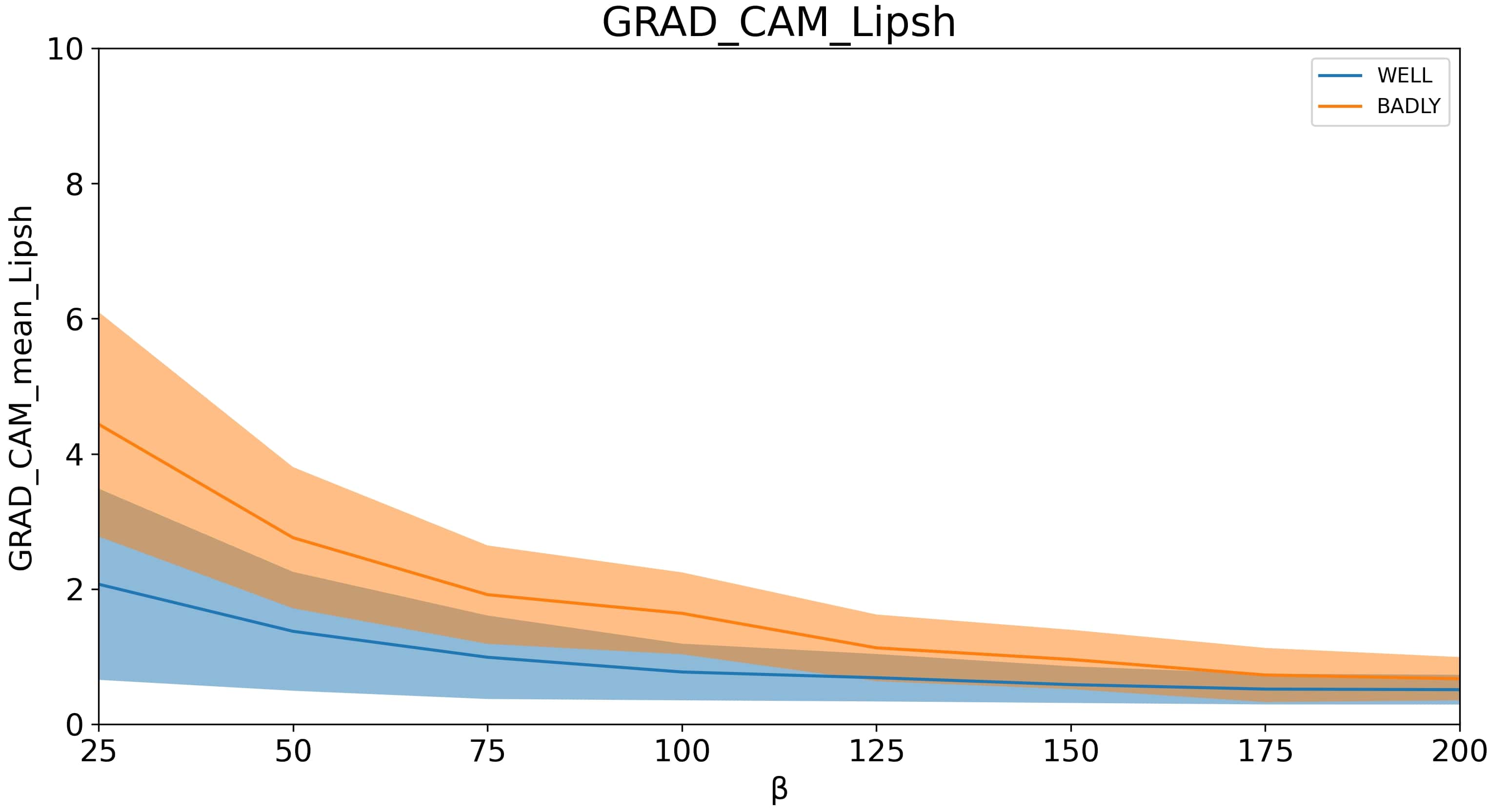}}
        \subfigure []{
            \label{fig:subfig:figs_Uniform_Brightness_Distortion:PCC_fem_fig} 
            \includegraphics[scale=0.040]{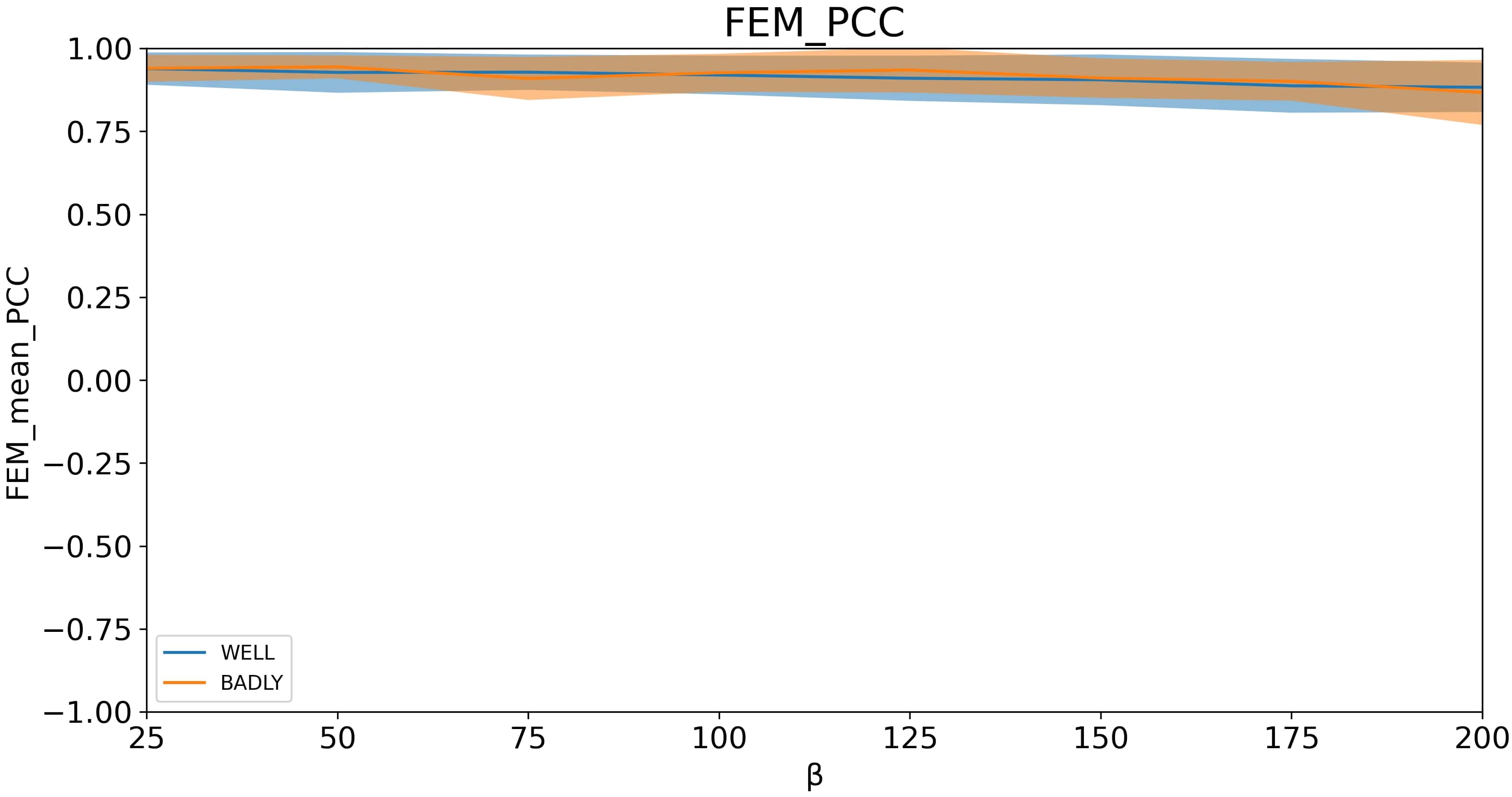}}
        \subfigure []{
            \label{fig:subfig:figs_Uniform_Brightness_Distortion:PCC_mlfem_fig} 
            \includegraphics[scale=0.040]{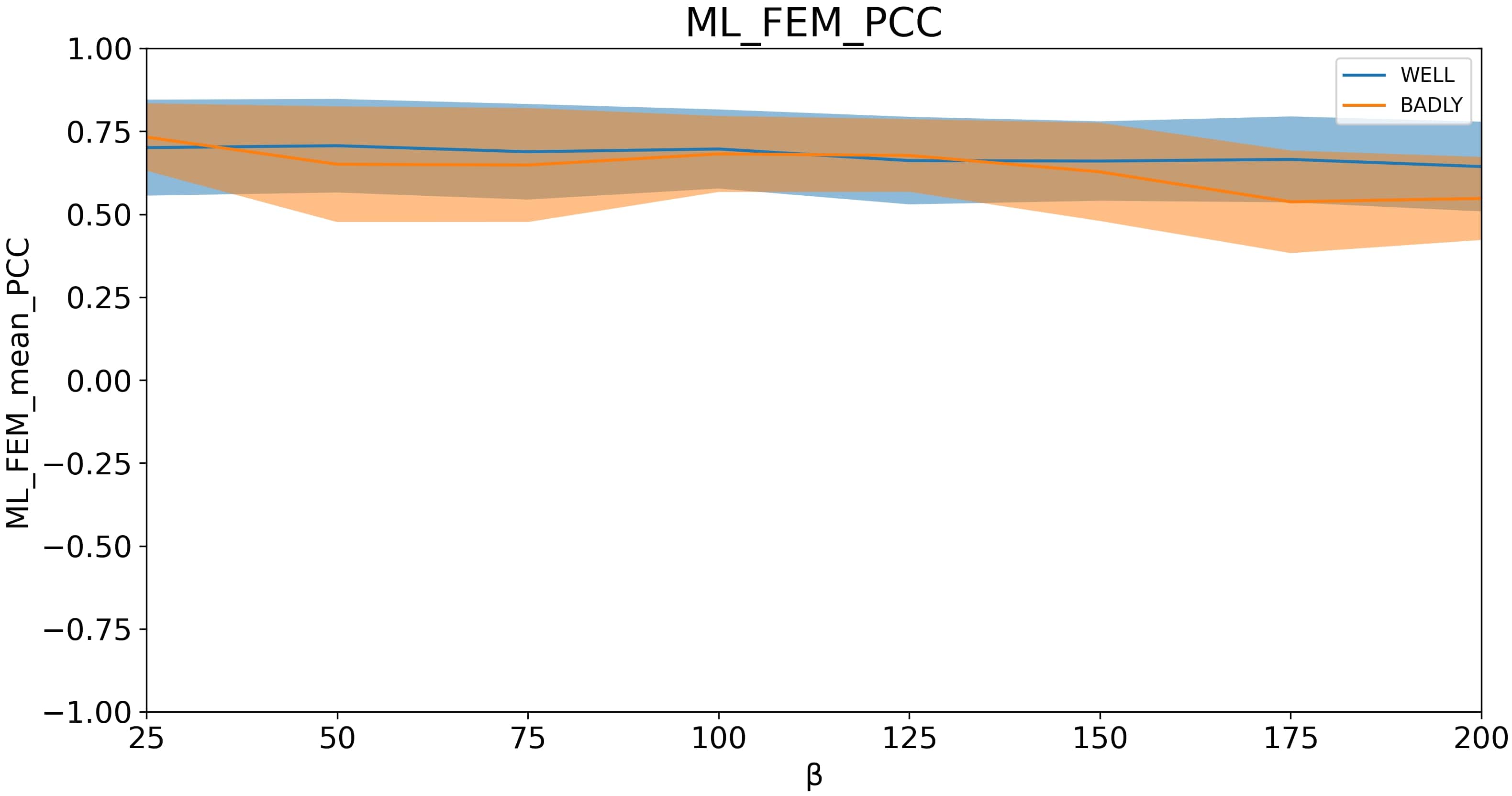}}
        \subfigure []{
            \label{fig:subfig:figs_Uniform_Brightness_Distortion:PCC_gradcam_fig} 
            \includegraphics[scale=0.040]{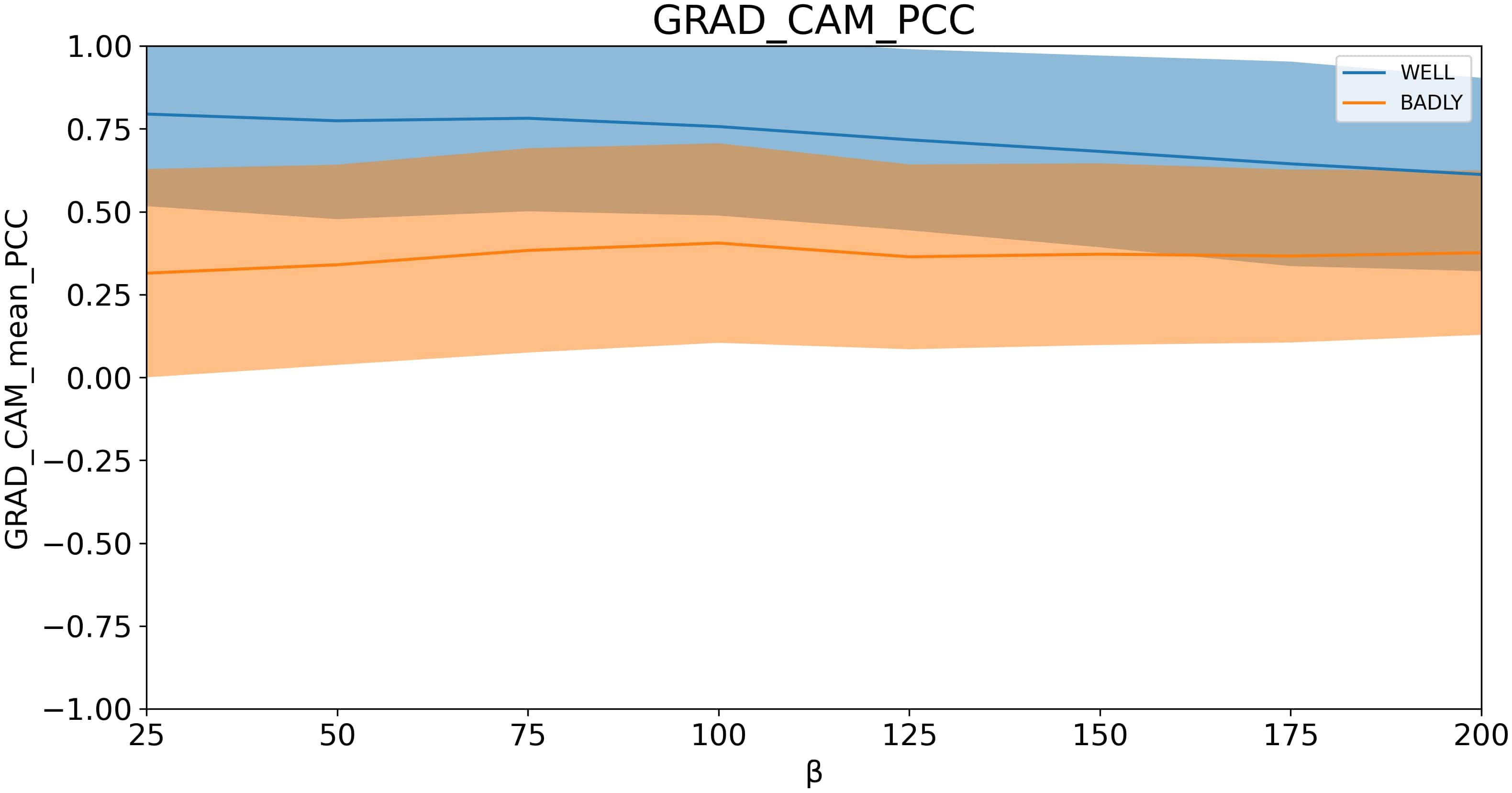}}
        \subfigure []{
            \label{fig:subfig:figs_Uniform_Brightness_Distortion:SIM_fem_fig} 
            \includegraphics[scale=0.040]{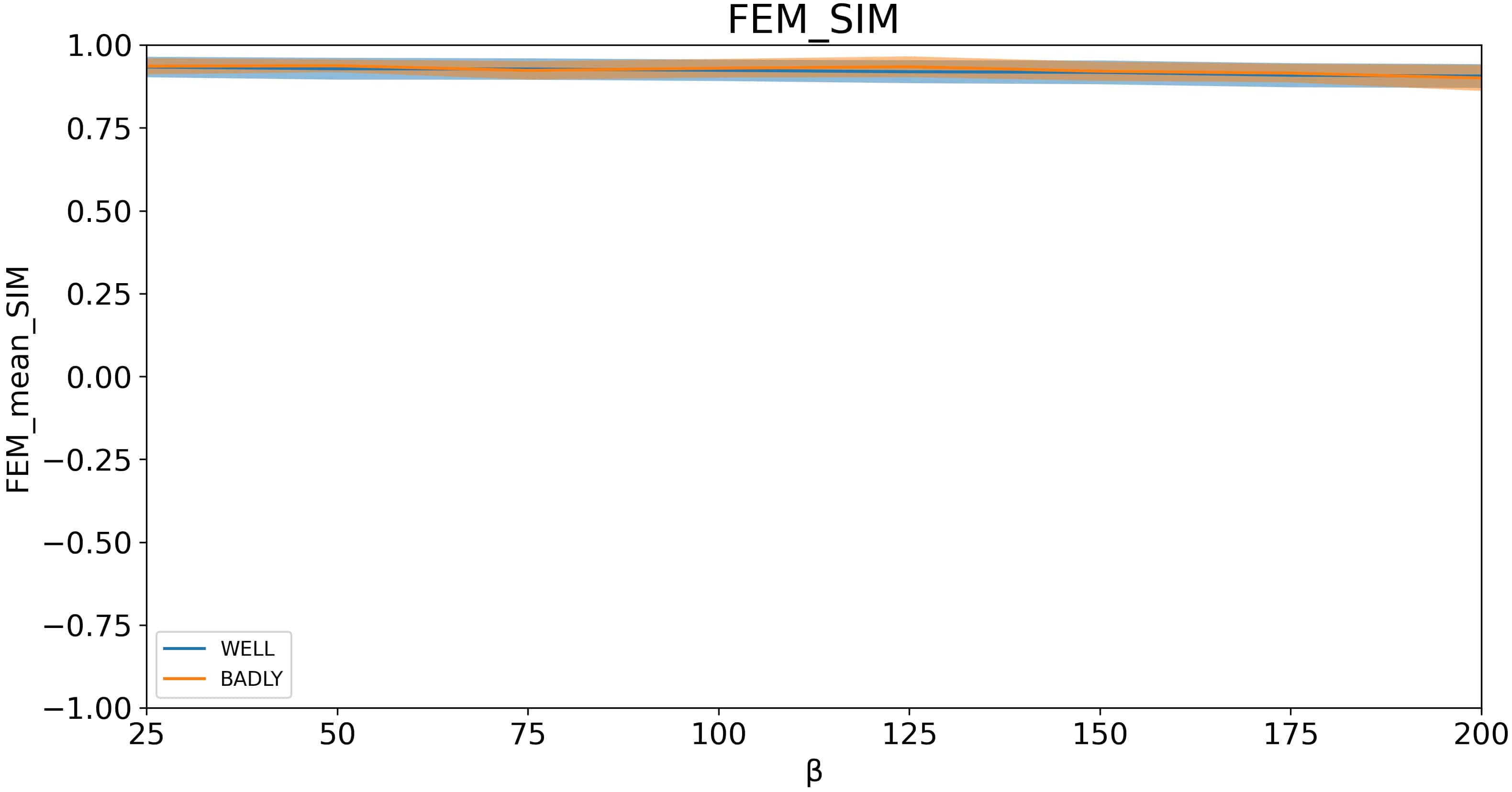}}
        \subfigure []{
            \label{fig:subfig:figs_Uniform_Brightness_Distortion:SIM_mlfem_fig} 
            \includegraphics[scale=0.040]{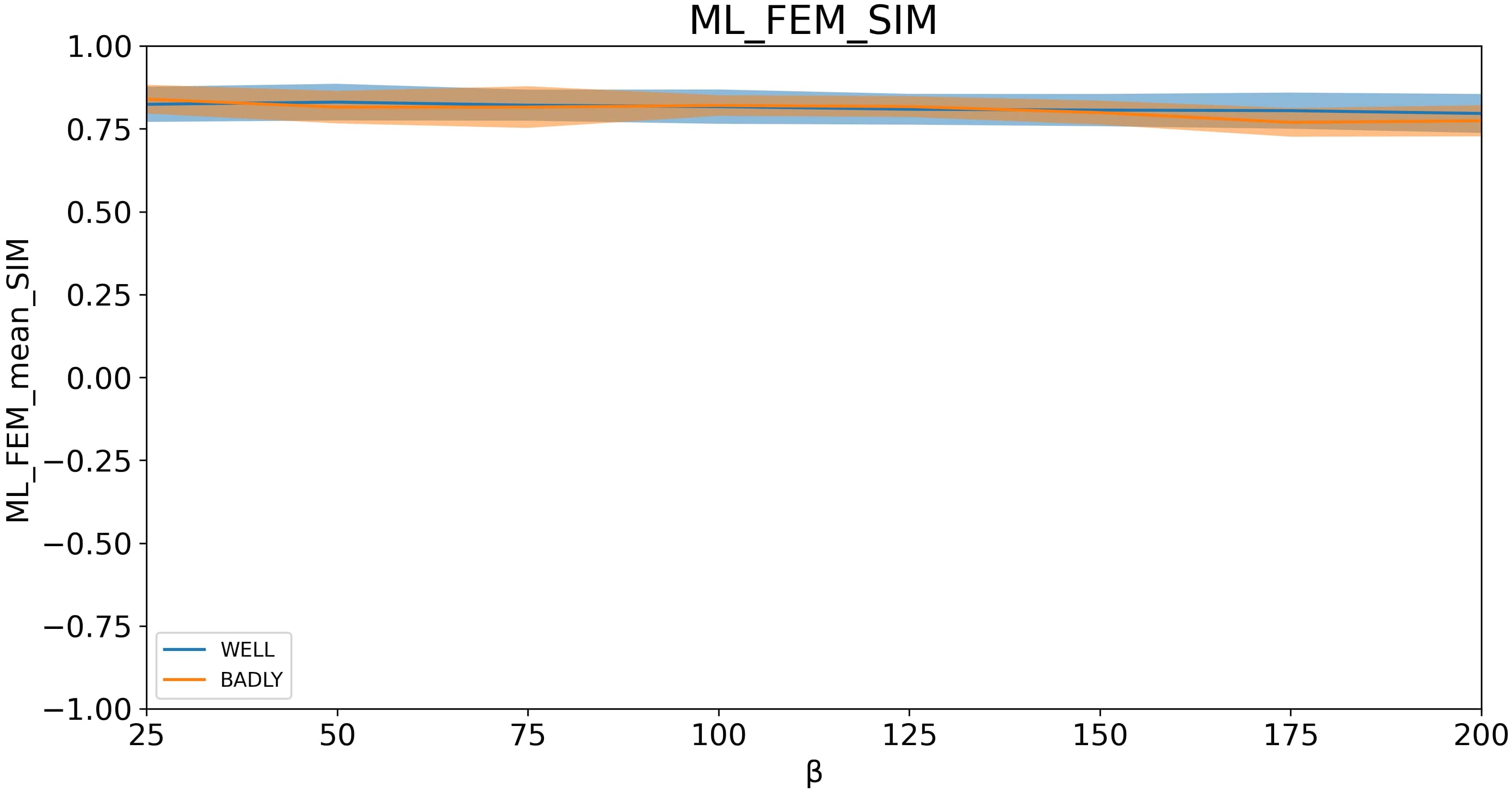}}
        \subfigure []{
            \label{fig:subfig:figs_Uniform_Brightness_Distortion:SIM_gradcam_fig} 
            \includegraphics[scale=0.040]{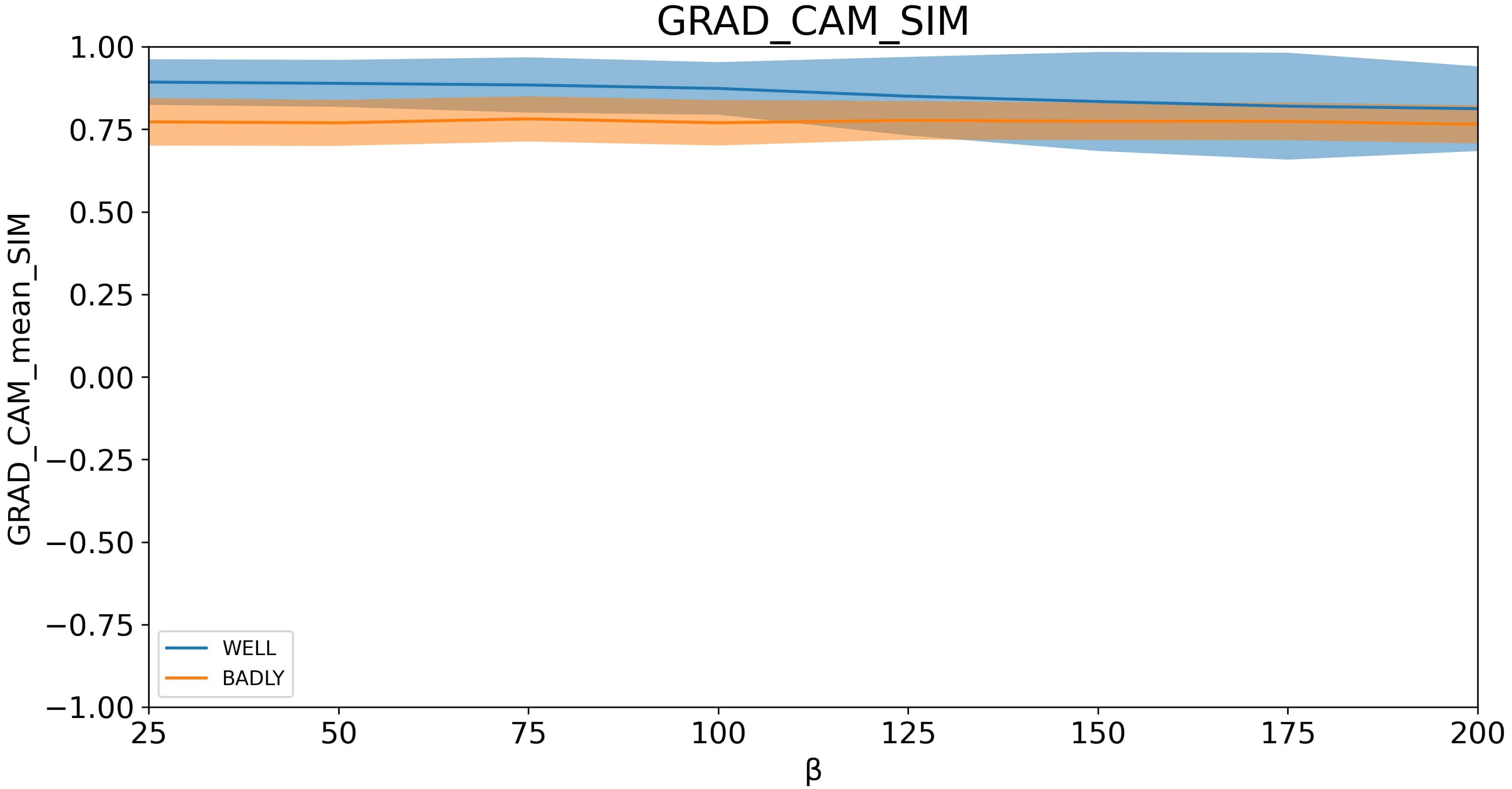}}
         \caption {Uniform Brightness Distortion: Behaviour of Lipschitz constant, of PCC and of SIM measures as a function of noise level: (a)FEM-Lipschitz, (b)MLFEM-Lipschitz, (c)GRAD-CAM-Lipschitz, (d)FEM-PCC, (e)MLFEM-PCC, (f)GRAD-CAM-PCC, (g)FEM-SIM, (h)MLFEM-SIM, (i)GRAD-CAM-SIM}
      \label{fig:figs_Uniform_Brightness_Distortion} 
    \end{figure}

            \label{subsubsection:Experiment_Uniform Brightness Distortion_Results of the experiment}
            This experiment demonstrates interesting results related to well-classified and badly-classified images. The behaviour of the Lipschitz constant is similar compared to previous experiments (Gaussian noise and Gaussian blur), see figures \ref{fig:subfig:figs_Uniform_Brightness_Distortion:L_fem_fig}, \ref{fig:subfig:figs_Uniform_Brightness_Distortion:L_mlfem_fig}, \ref{fig:subfig:figs_Uniform_Brightness_Distortion:L_gradcam_fig}. 
            But it is worth noting that with this distortion, some randomness of the final data is manifested, as the stability over distorted data is lower for all methods, see figures \ref{fig:subfig:figs_Uniform_Brightness_Distortion:PCC_fem_fig}, \ref{fig:subfig:figs_Uniform_Brightness_Distortion:PCC_mlfem_fig}, \ref{fig:subfig:figs_Uniform_Brightness_Distortion:PCC_gradcam_fig} and \ref{fig:subfig:figs_Uniform_Brightness_Distortion:SIM_fem_fig}, \ref{fig:subfig:figs_Uniform_Brightness_Distortion:SIM_mlfem_fig}, \ref{fig:subfig:figs_Uniform_Brightness_Distortion:SIM_gradcam_fig} and tables \ref{tab:L_Mean_Sigma_tab_Uniform_Brightness_Distortion}, \ref{tab:PCC_Mean_Sigma_tab_Uniform_Brightness_Distortion} and \ref{tab:SIM_Mean_Sigma_tab_Uniform_Brightness_Distortion}. Perhaps this is due to the disappearance of objects of attention in the images with even not too strong distortion, nevertheless, the FEM method demonstrates itself as more reliable. \\

            Consensus of metrics is  given in the table \ref{tab:correlation_of_metrics_Uniform_Brightness_Distortion}. From these figures, it can be concluded that in the case of Uniform Brightness Distortion, the consensus values are lower than in previous experiments.
            In presence of such kind of noise, it is better to keep reference-based evaluation.

\begin{figure}[H]
      \centering
        \subfigure []{
            \label{fig:subfig:figs_Perspective_Distortion:L_fem_fig} 
            \includegraphics[scale=0.040]{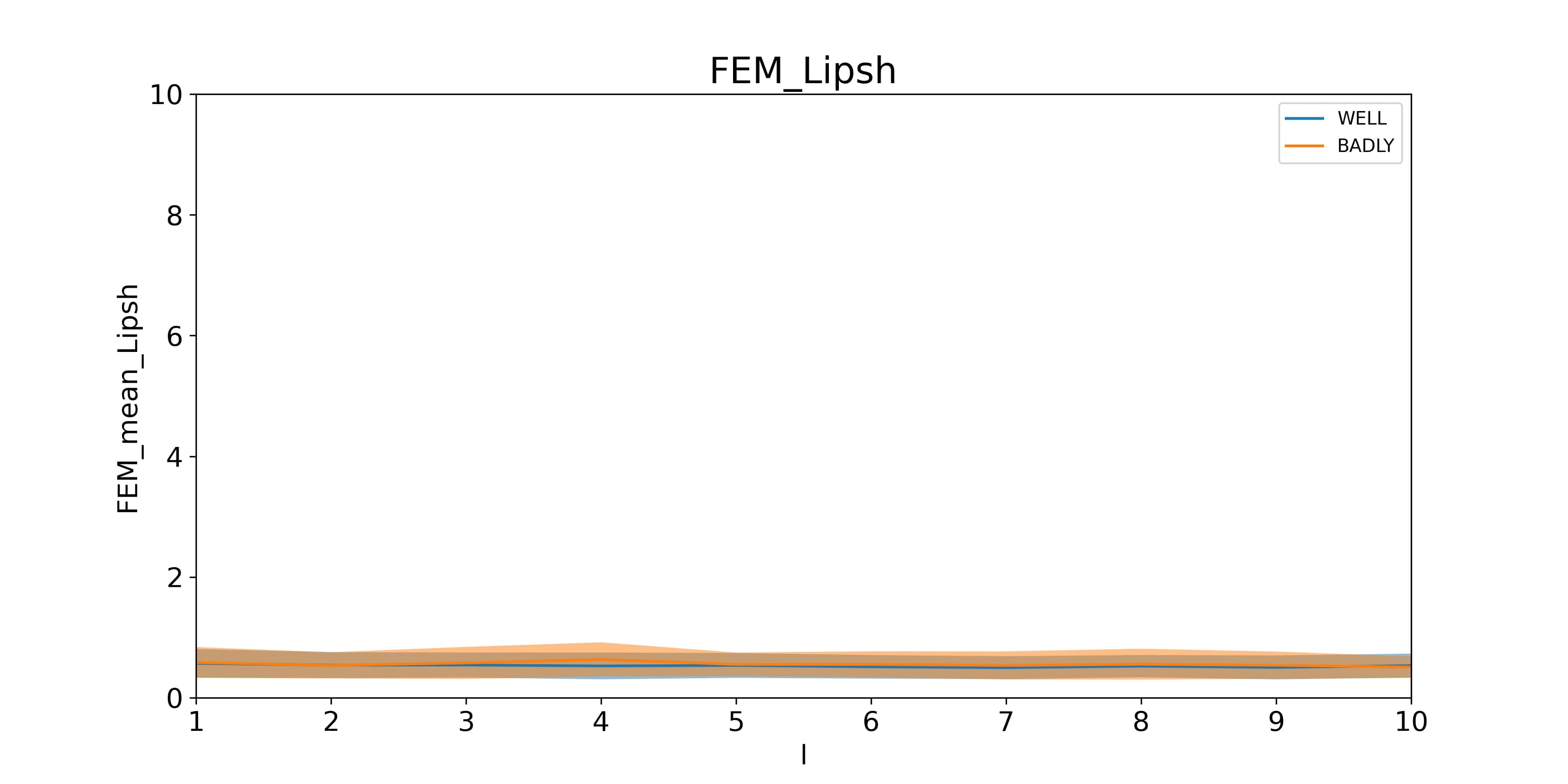}}
        \subfigure []{
            \label{fig:subfig:figs_Perspective_Distortion:L_mlfem_fig} 
            \includegraphics[scale=0.039]{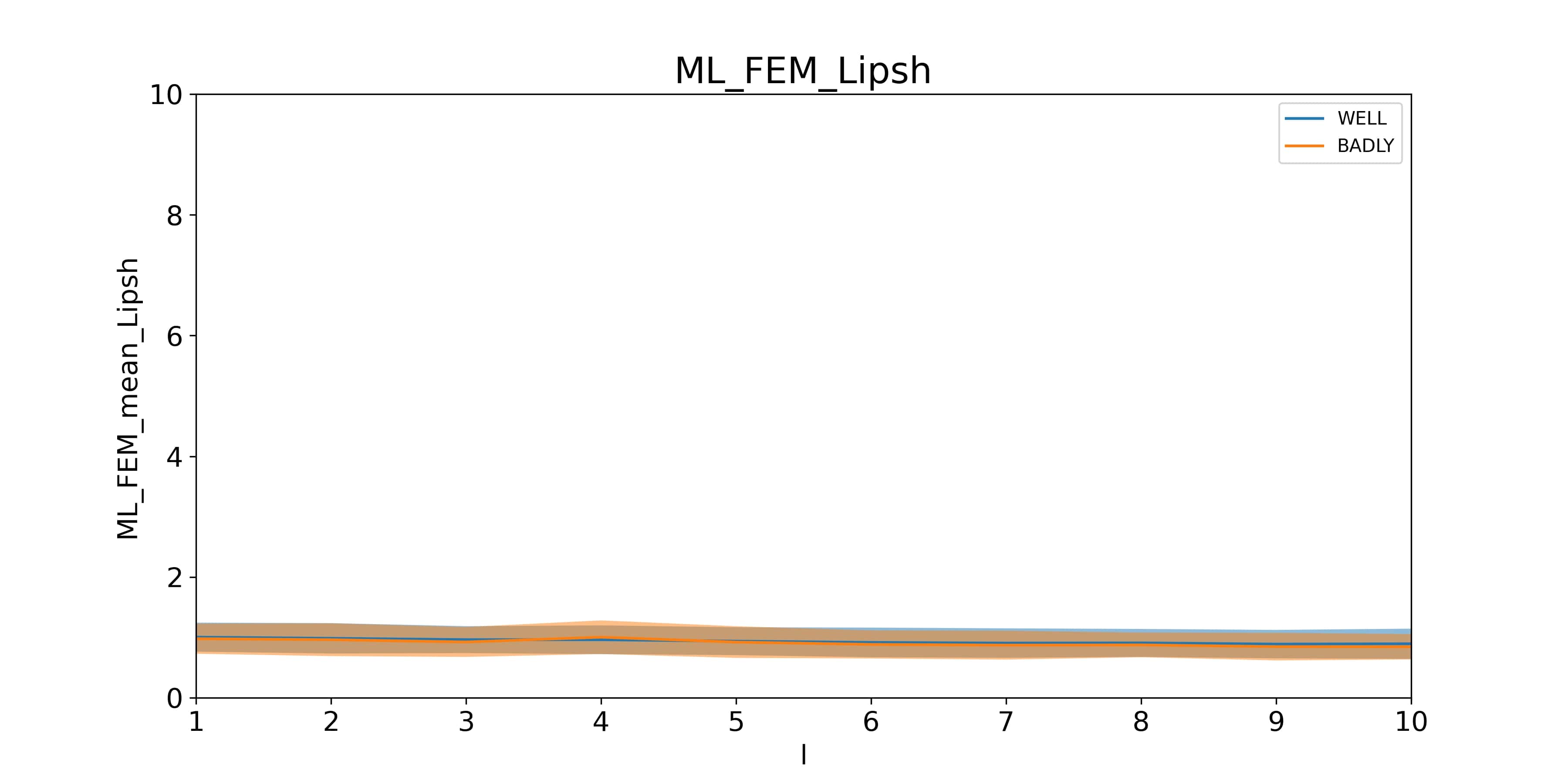}}
        \subfigure []{
            \label{fig:subfig:figs_Perspective_Distortion:L_gradcam_fig} 
            \includegraphics[scale=0.039]{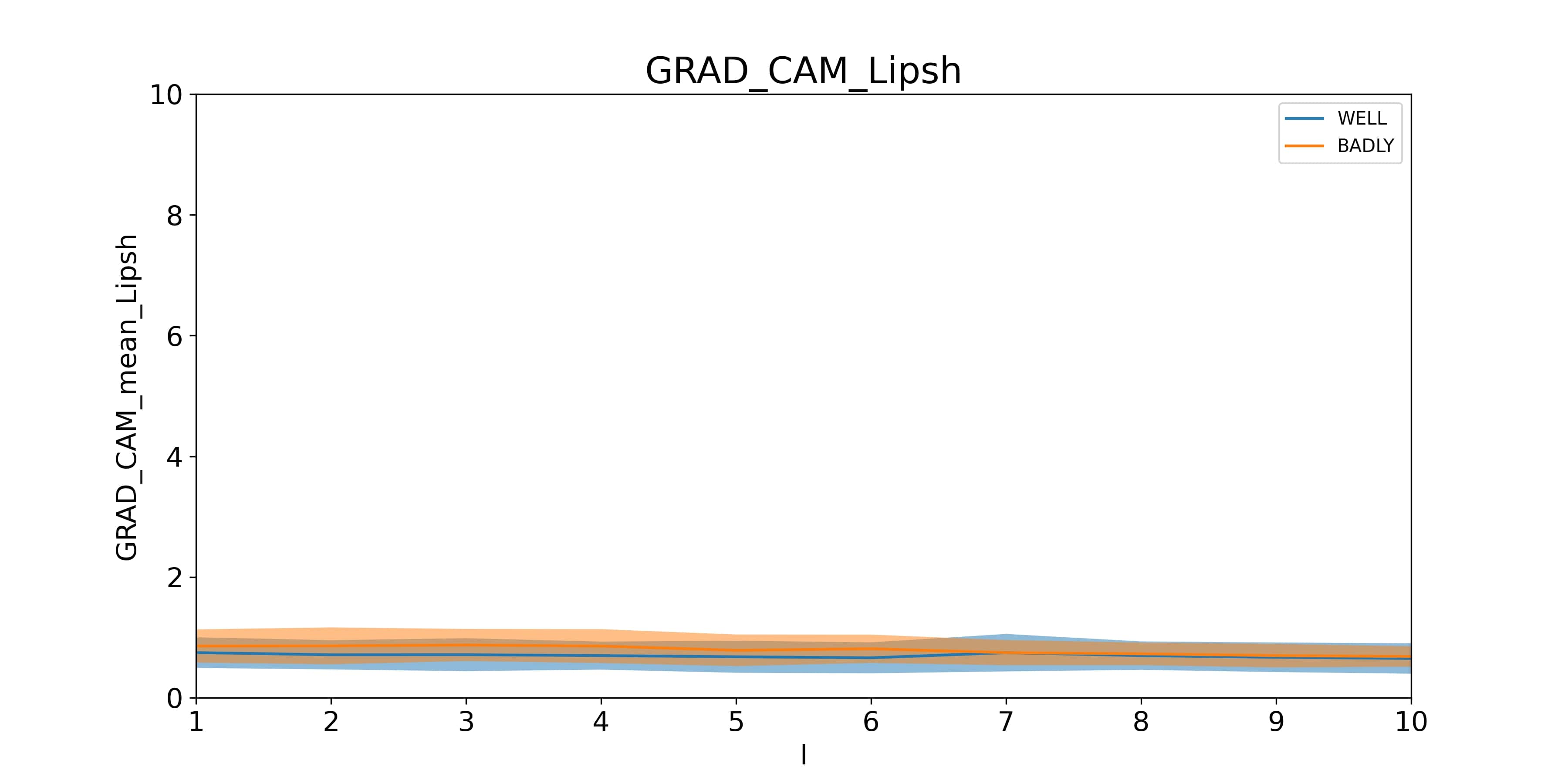}}
        \subfigure []{
            \label{fig:subfig:figs_Perspective_Distortion:PCC_fem_fig} 
            \includegraphics[scale=0.040]{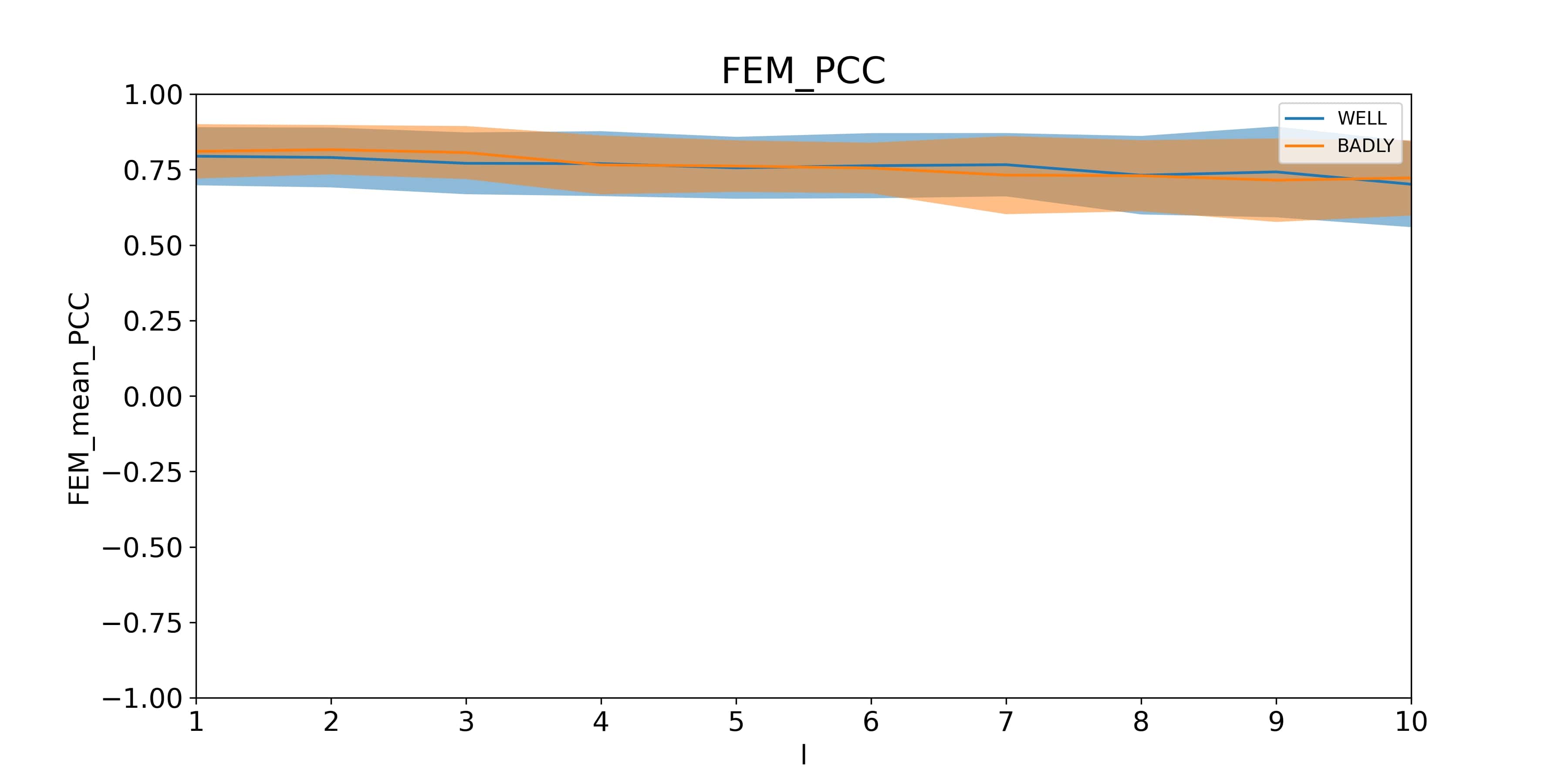}}
        \subfigure []{
            \label{fig:subfig:figs_Perspective_Distortion:PCC_mlfem_fig} 
            \includegraphics[scale=0.039]{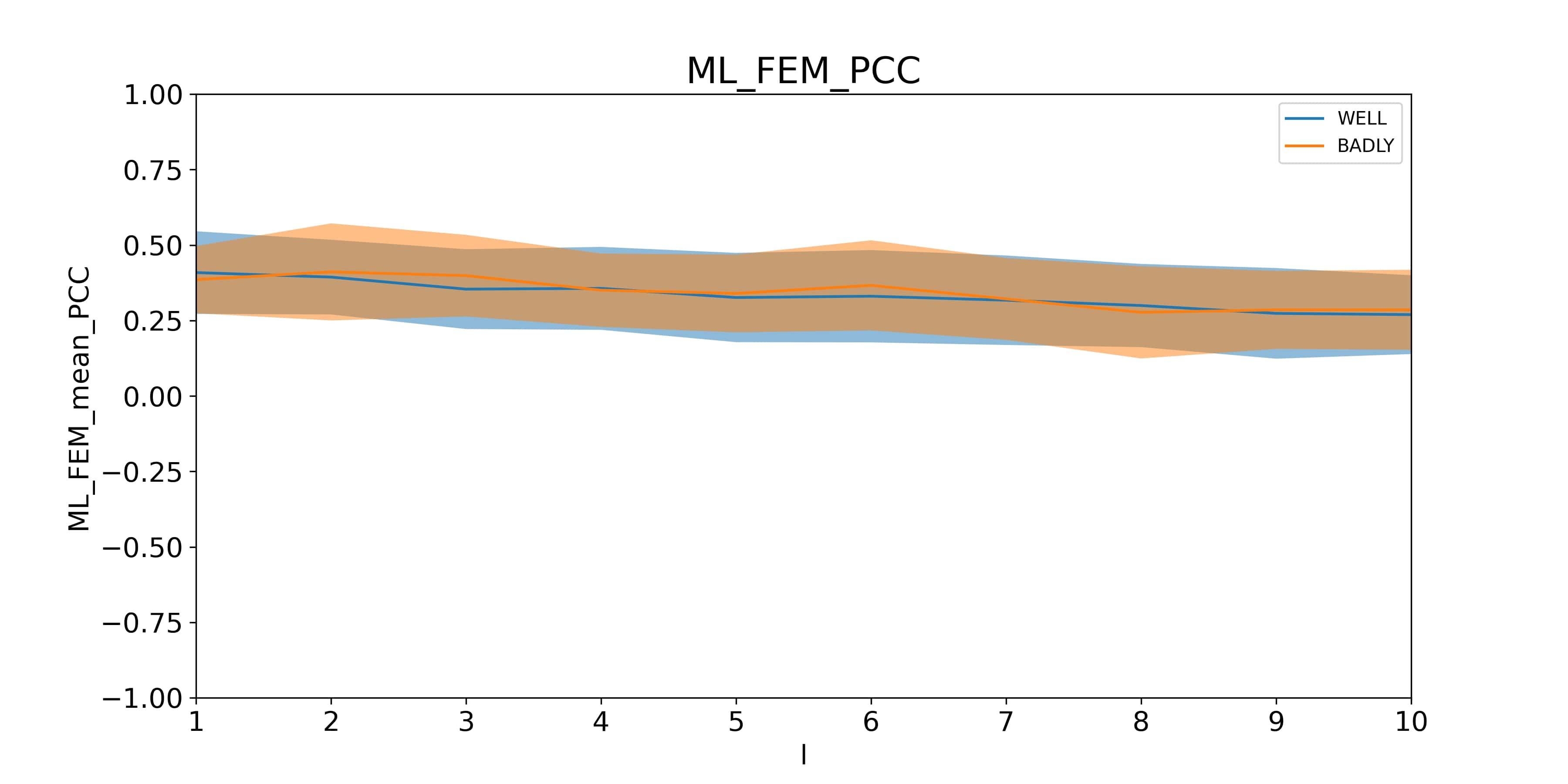}}
        \subfigure []{
            \label{fig:subfig:figs_Perspective_Distortion:PCC_gradcam_fig} 
            \includegraphics[scale=0.039]{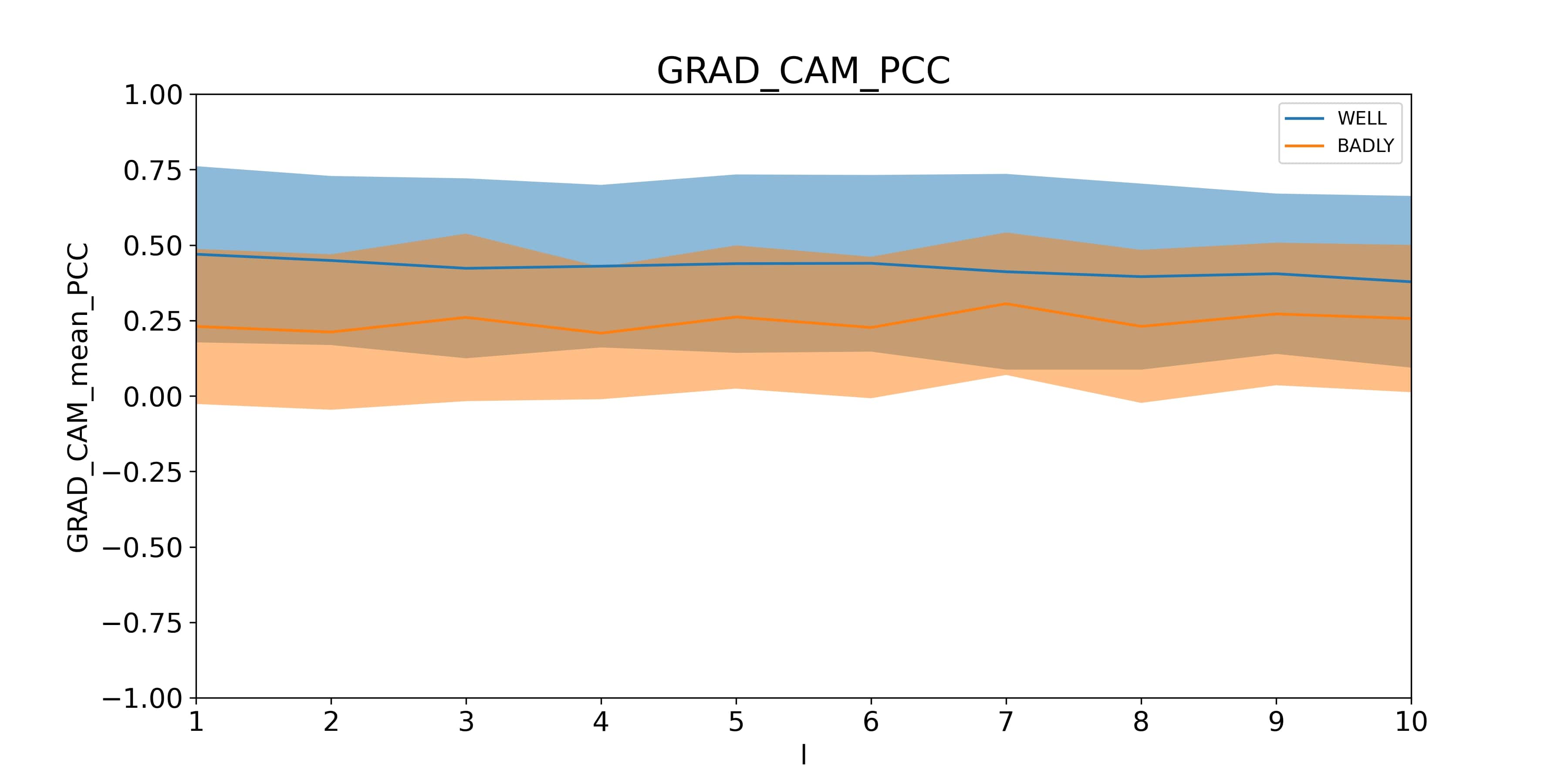}}
        \subfigure []{
            \label{fig:subfig:figs_Perspective_Distortion:SIM_fem_fig} 
            \includegraphics[scale=0.040]{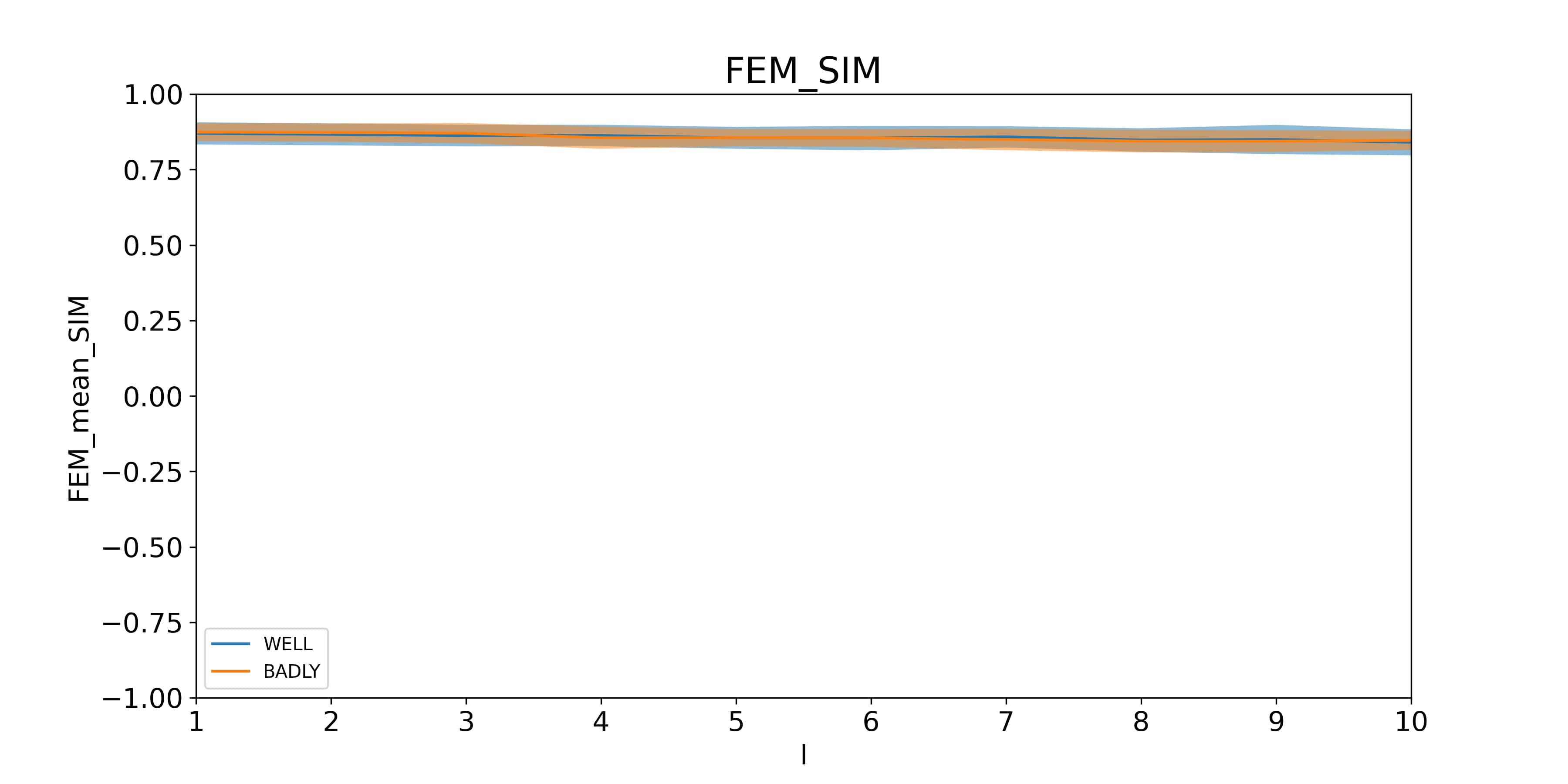}}
        \subfigure []{
            \label{fig:subfig:figs_Perspective_Distortion:SIM_mlfem_fig} 
            \includegraphics[scale=0.039]{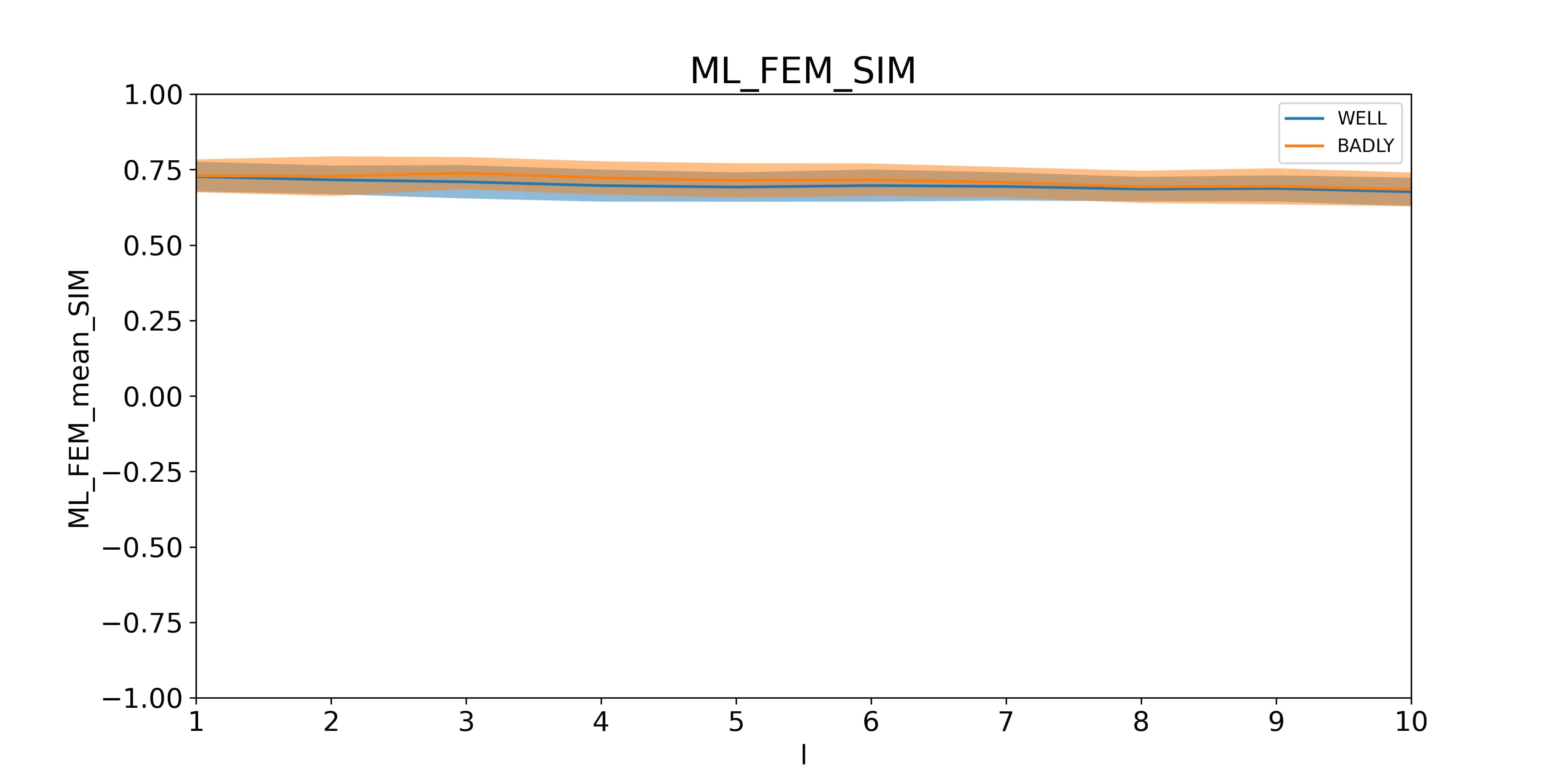}}
        \subfigure []{
            \label{fig:subfig:figs_Perspective_Distortion:SIM_gradcam_fig} 
            \includegraphics[scale=0.039]{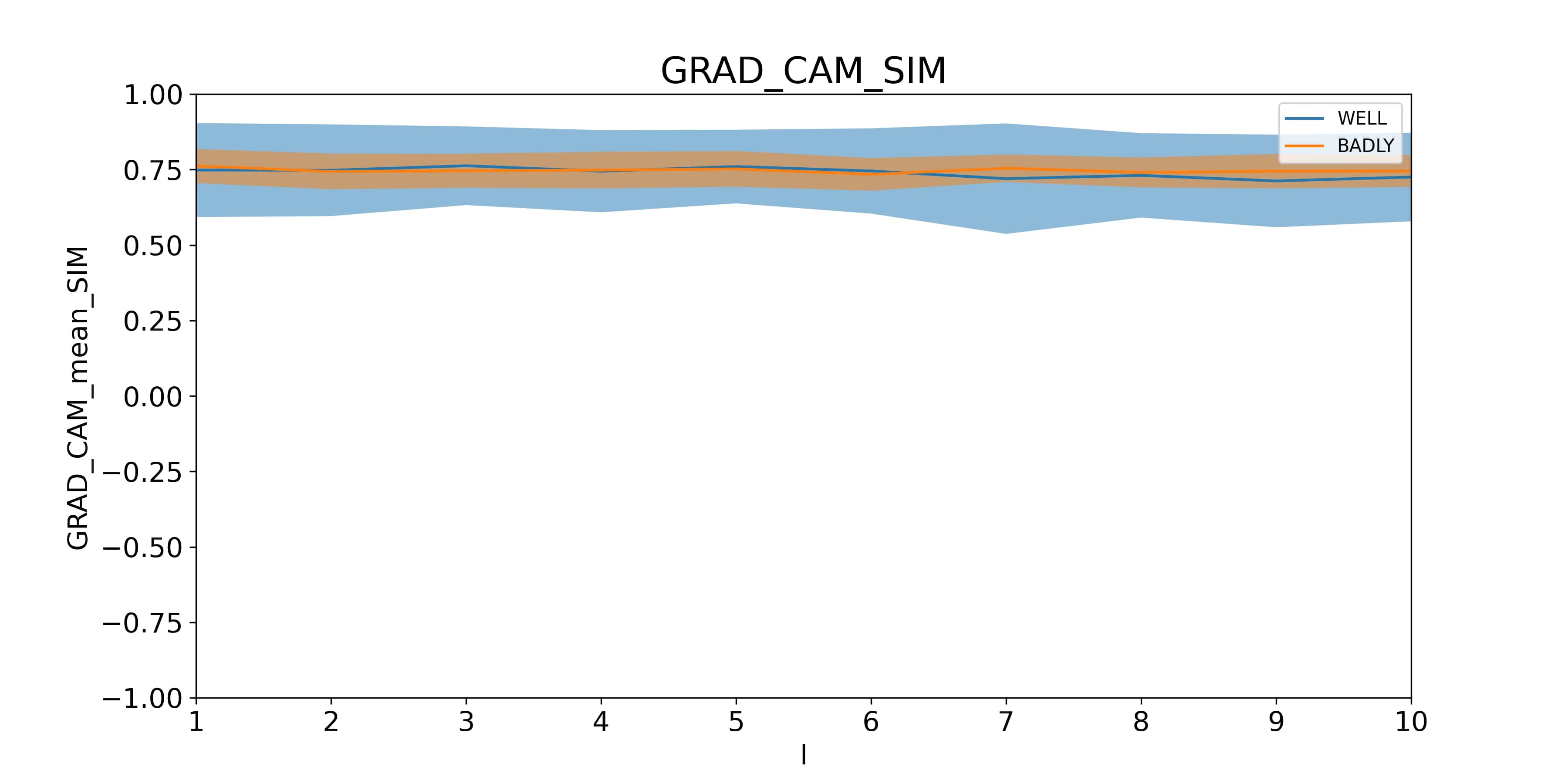}}
         \caption {Perspective Distortion: Behaviour of Lipschitz constant, of PCC and of SIM measures as a function of distortion level: (a)FEM-Lipschitz, (b)MLFEM-Lipschitz, (c)GRAD-CAM-Lipschitz, (d)FEM-PCC, (e)MLFEM-PCC, (f)GRAD-CAM-PCC, (g)FEM-SIM, (h)MLFEM-SIM, (i)GRAD-CAM-SIM}
      \label{fig:figs_Perspective_Distortion} 
    \end{figure}

        \paragraph{\textbf{Perspective Distortion}}
        \label{subsection:Experiment_Perspective Distorsion}
            With this distortion, the Lipschitz constant shows lower values compared to other degradations even with minimal perspective changes in the images (figures \ref{fig:subfig:figs_Perspective_Distortion:L_fem_fig},
            \ref{fig:subfig:figs_Perspective_Distortion:L_mlfem_fig},
            \ref{fig:subfig:figs_Perspective_Distortion:L_gradcam_fig}). A drop in the Lipschitz constant indicates an increase in distortion, which at some point in degradation scale can lead to stability of the constant (maps and images themselves are too different from the original ones). Indeed, in these degradations, important areas in images ``move". Low intensity values move to the high intensity locations and vice versa. Thus, the difference between the original image and the degraded one becomes higher.  Thus, higher is the difference between the explanation maps.  Indeed, explanation maps capture strong features which are situated on deformed and displaced borders of objects. Therefore, gaze fixation density maps differ a lot from good explanation maps on the distorted image. The behaviour of explanation methods is also illustrated in figures \ref{fig:subfig:figs_Perspective_Distortion:PCC_fem_fig},
            \ref{fig:subfig:figs_Perspective_Distortion:PCC_mlfem_fig},
            \ref{fig:subfig:figs_Perspective_Distortion:PCC_gradcam_fig} in terms of PCC and in figures
            \ref{fig:subfig:figs_Perspective_Distortion:SIM_fem_fig},
            \ref{fig:subfig:figs_Perspective_Distortion:SIM_mlfem_fig},
            \ref{fig:subfig:figs_Perspective_Distortion:SIM_gradcam_fig} in terms of SIM.
            The stability of metrics is presented in tables \ref{tab:L_Mean_Sigma_tab_Perspective_Distortion}, \ref{tab:PCC_Mean_Sigma_tab_Perspective_Distortion} and \ref{tab:SIM_Mean_Sigma_tab_Perspective_Distortion}. From table \ref{tab:L_Mean_Sigma_tab_Perspective_Distortion} one can see that the Lipschitz constant does not change practically. \\

            Consensus of metrics is presented in the table \ref{tab:correlation_of_metrics_Perspective_Distortion}. Based on the data obtained, it can be concluded that in the case of Perspective Distortion, the no-reference stability metric also as with Gaussian noise or with Gaussian blur demonstrates consensus with SIM referenced-based metric. Thus, the Lipschitz constant can be used to determine the quality of explainers.
            

\section{Conclusion and Discussion}
\label{section:conclusion}
In this work, we have studied Lipschitz constant as a non-reference metric of the quality of  explanation methods. 
We have applied it for three explanation methods Grad-CAM, FEM and MLFEM. 
We also studied the agreement of this metric with previously used  reference-based PCC and SIM metrics computed by comparing ExMs with Gaze Fixation Density Maps available for classified images. We did these studies on images corrupted with growing noise, separating two cases: i)images which were misclassified after their corruption by distortions, ii)images which kept their class labels correctly. \\

Accordingly to the results obtained, it can be concluded that the Lipschitz constant does not increase with increasing distortions of the image.  Since with a serious change in the original image, the explanation map must also seriously change relatively to the original map, which leads to stabilization and non-growth of the Lipschitz constant. The experimental behaviour thus confirms our hypothesis. In addition, the correlation with other, reference-based,  metrics demonstrates high values. \\

As a conclusion, we have two points. 
(i) Comparing the three methods, we state that FEM method is the best explainer from the point of view of all three metrics, reference - based and non-reference ones;
(ii) Due to the very good agreement between Lipschitz constant and  SIM values and PCC at a lower extent on the best explainers FEM and MLFEM, this non-reference stability metric can be used in case when the Gaze Fixation Density Maps are not available for images to classify. This opens tremendous perspectives of evaluation of explainers in DNN classification of non-visual data.  


\bibliographystyle{unsrt}
\bibliography{main}

\appendix
\section{Evaluated Explainers}

    \label{section:ExplanationMethods}
\begin{figure}[H]
      \centering
         \subfigure[] {
        \includegraphics[scale=0.15]{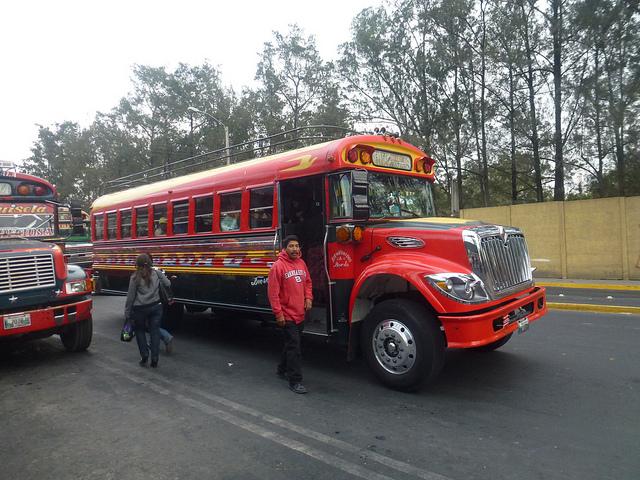}}%
         \subfigure []{
        \label{fig:original_superimposed_FEM}
        \includegraphics[scale=0.15]{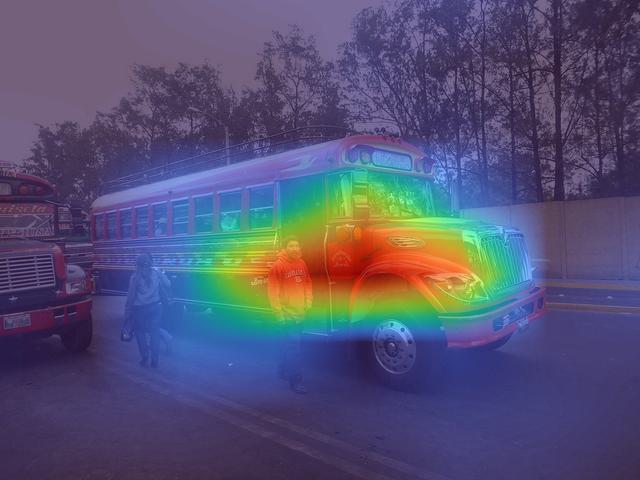}}%
         \subfigure []{
            \label{fig:original_superimposed_MLFEM} 
            \includegraphics[scale=0.15]{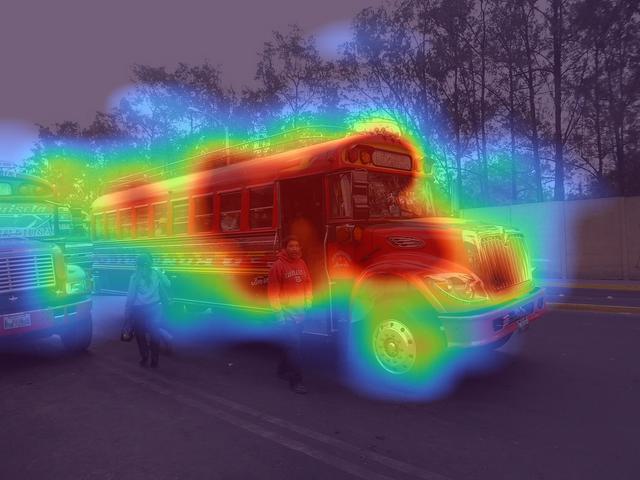}}%
        \subfigure[] {
        \label{fig:original_superimposed_GRAD_CAM}
        \includegraphics[scale=0.15]{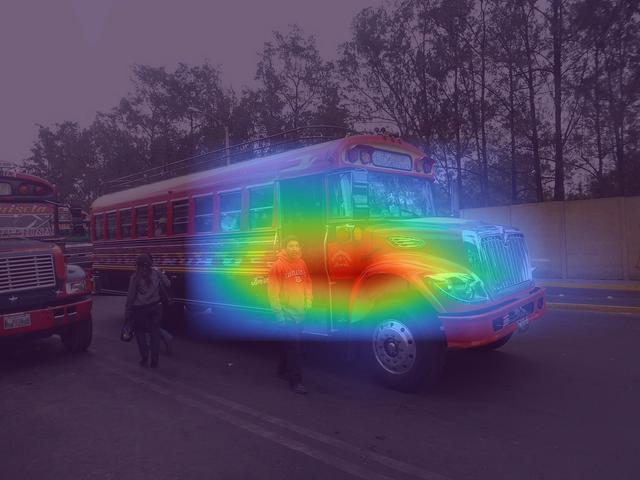}}
         \caption {Illustration of the explainers: (a) Original image, (b) Superimposed with FEM, (c) superimposed with MLFEM, (d) superimposed with Grad-CAM\label{fig_all_sota}}

    \end{figure}

    This section describes the explainers evaluated in this paper: FEM~\cite{ahmedasiffuad:hal-02963298}, ML-FEM~\cite{DBLP:conf/icprai/BourrouxBBG22} and Grad-CAM~\cite{DBLP:journals/corr/SelvarajuDVCPB16}. Examples of ExMs obtained by these methods on one image are presented in (Figure~\ref{fig_all_sota}).

\begin{figure}[H]
        \centering
        \includegraphics[width=1\textwidth]{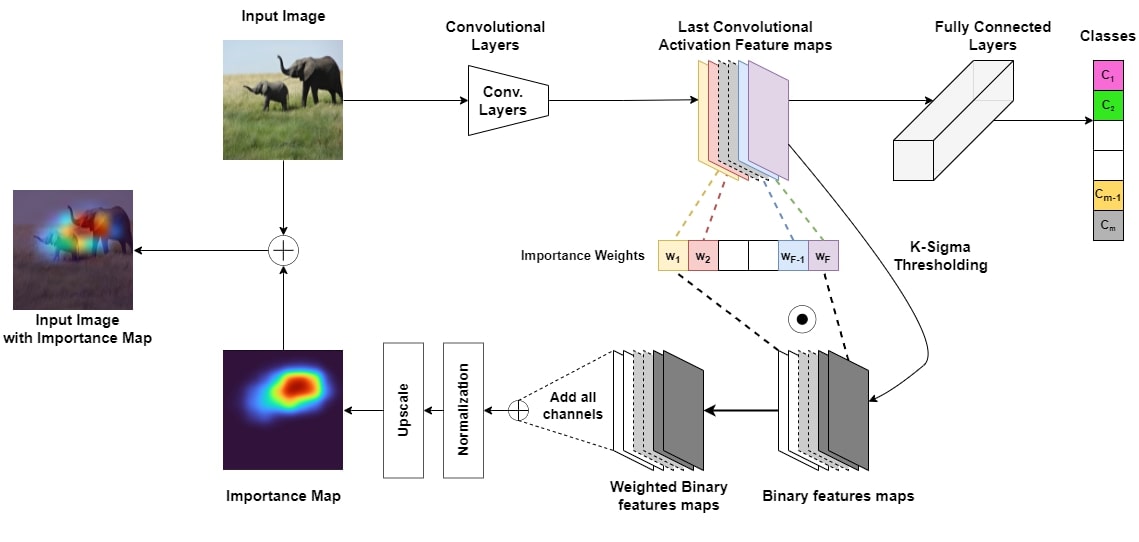}
        \caption{\label{fig:overview_of_FEM}Overview of the explainer FEM. Features are extracted from the last convolutional layer after the activation function (upper part). Binary features maps are generated. Importance weights are calculated. Importance map is computed as a normalized linear combination with channel weights and visualized as a heat-map on the original image \cite{ahmedasiffuad:hal-02963298}}
    \end{figure}

    \paragraph{\textbf{Feature Explanation Method (FEM)}}
        \label{subsection:FEM}
        The essence of FEM  (Figure~\ref{fig:overview_of_FEM}) is the reverse tracking of the most high features from the last layer of features, namely from the last convolution layer of a CNN. It can be used to identify network solutions at the generalization stage.  At this generalization stage, a fragment of the input images for classification is transmitted in the forward direction over the trained network. In CNN, convolution layers act as feature extractors, and the last fully connected layers act as their classifiers. The upper levels of convolution extract low-level features from the input data, while the deeper ones extract higher-level semantic features. Consequently, features from the last layer of convolution are extracted and analysed. The features are picked up after the activation layer and immediately before the fully connected layers.

   A CNN is feed with an image of size $(W \times H)$ processed over various convolutional and pooling layers. The activation map of the last convolutional layer contains several channels (one per convolutional kernel).
   FEM  generates a binary map for each channel of the feature tensor to assign an importance value for each feature in a channel. In order to detect the strongest features, the developers of the method hypothesize that the feature values in feature maps follow Gaussian distributions independently by channel. From this, we can conclude that the means are positive, since they extract features after the commonly used non-linearity ReLU, which converts negative values into 0.
%
 %
    On the last convolutional layers, features are the most "vivid", but only some of them are of value. According to the Gaussian distribution hypothesis, the most valuable for studying is the correct distribution queue, which corresponds to rare but strong features. In this way, limits are created for maps of features of $x_i,c$ in accordance with the $K$-sigma rule. Mean $\mu_c$ and standard deviation $\sigma_c$ are calculated
    for each channel $c$. Then a binary map per channel is built which
    marks the strong features, as in:\\

    \begin{equation}
    b_c(R(x_i,c)) =       
    \begin{cases}
        1  & \quad \text{if } x_i,c \geq \mu_c + K . \sigma_c\\
        0  & \quad \text{otherwise}
    \end{cases}
    \label{eq:FEM}
    \end{equation} \\
   
    After that, the importance map $M$ of the explainer is calculated as a linear combination of all binary channel maps $b_c$ using the weights $\mu_c$ of these channels with its subsequent normalization to \quad [0; 1]. When normalizing, Min-Max scaling of M values is used. Finally, the normalized importance map $M'$ is upscaled to the original image/video frame dimension $W \times H$ by a bi-linear interpolation. An example of explanation maps obtained by FEM method is illustrated in figure \ref{fig:original_superimposed_FEM} below.\\

    \paragraph{\textbf{Multi Layered Feature Explanation Method (MLFEM)}}
        \label{subsection:MLFEM}
        The Multi Layered Feature Explanation Method (MLFEM) relies on FEM: while in FEM  the analysis of activations is performed at the last convolution level only, MLFEM extends it to all convolutional layers of a CNN. Since each layer of CNN embeds information at a different scale, the authors of \cite{DBLP:conf/icprai/BourrouxBBG22} suggest that calculating FEM at multiple layers and combining them will improve the quality of explanation maps. 
      %
        Applying FEM to each layer of CNN consisting of $L$ convolutional layers will give $L$ different maps of the importance of features. Since all importance maps are interpolated in the FEM method, we get $L$ input resolution maps as a result. The information provided by the maps depends on the convolutional layer in the network, and they need to be combined into a single pixel importance map.
        
        \begin{figure}[H]
            \centering
            \includegraphics[width=0.9\textwidth]{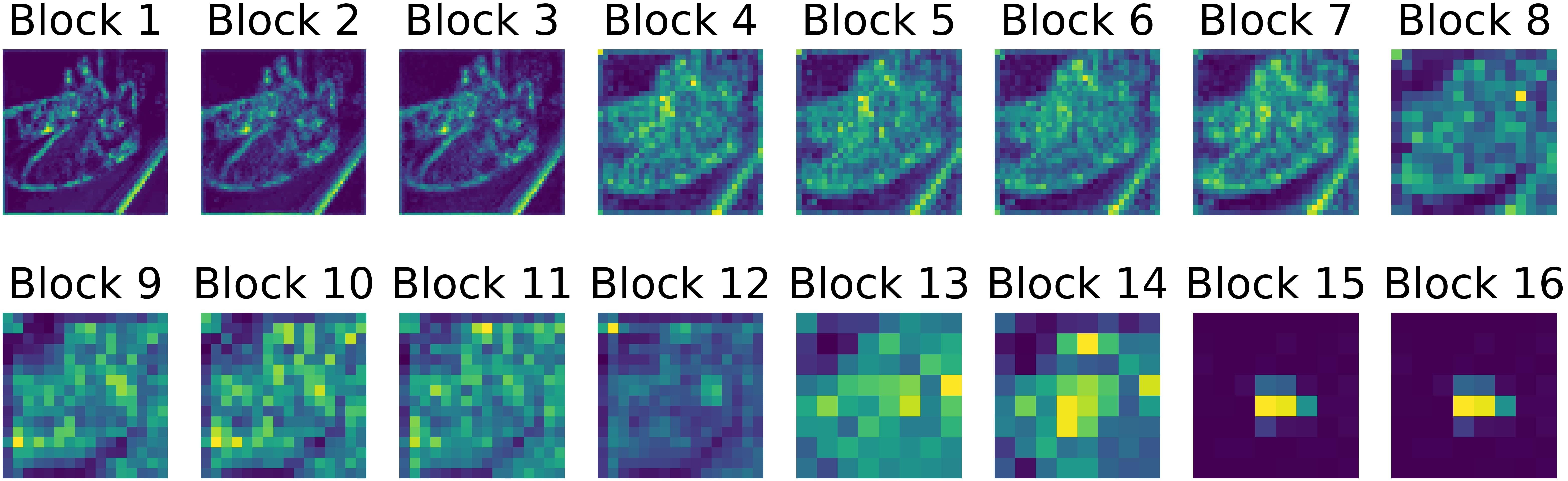}
            \caption{\label{fig:FEM_every_block}FEM applied on every convolutional block of a typical ResNet50 architecture. Resolution is higher for the first layers. \cite{DBLP:conf/icprai/BourrouxBBG22}}
        \end{figure}
        
        The network passes the input image through several convolution layers, giving results independent of the position of these layers. They are designed to capture spatio-local features. With each step deeper into the network, the convolutional layer captures more and more abstract concepts in the image (see figure \ref{fig:FEM_every_block}). The very first layers usually perform edge detection, while later ones extract abstract concepts such as “face”, “car”, etc. This is the rationale for the idea of repeatedly applying the same method of explanation on different layers of the network.\\
        
        For the combination of each of the $l = 1, ..., L$ maps, the authors of MLFEM method propose to train a shallow convolutional network which uses as the ground-truth Gaze Fixation Density Maps and a Euclidean Loss function \cite{DBLP:conf/icprai/BourrouxBBG22}. They show that such a fusion method is model-agnostic. An example of pixel importance map obtained by MLFEM method is given in figure \ref{fig:original_superimposed_MLFEM} for the same image as in  figure \ref{fig:original_superimposed_FEM}. It can be seen that the MLFEM map is much more detailed and captures the important details in the image, such as wheels of the car in this case.  
        

    \paragraph{\textbf{Gradient-weighted Class Activation Mapping (Grad-CAM)}}
    \label{subsection:Grad-CAM}
    Information about space is preserved naturally in convolutional features, but is lost in fully connected layers. It follows from this that we can find the best compromise between high-level semantics and detailed spatial information on the last convolutional layers. This is the assumption of one of the most popular explainers, the Grad-CAM method \cite{DBLP:journals/corr/SelvarajuDVCPB16}.  Neurons in these layers search for semantic information (part of an object) related to a specific class in the image. To understand the influence of each neuron for our solution, Grad-CAM uses gradient information coming into the last convolutional layer of CNN.\\
    
    \begin{figure}[H]
        \centering
        \includegraphics[width=1\textwidth]{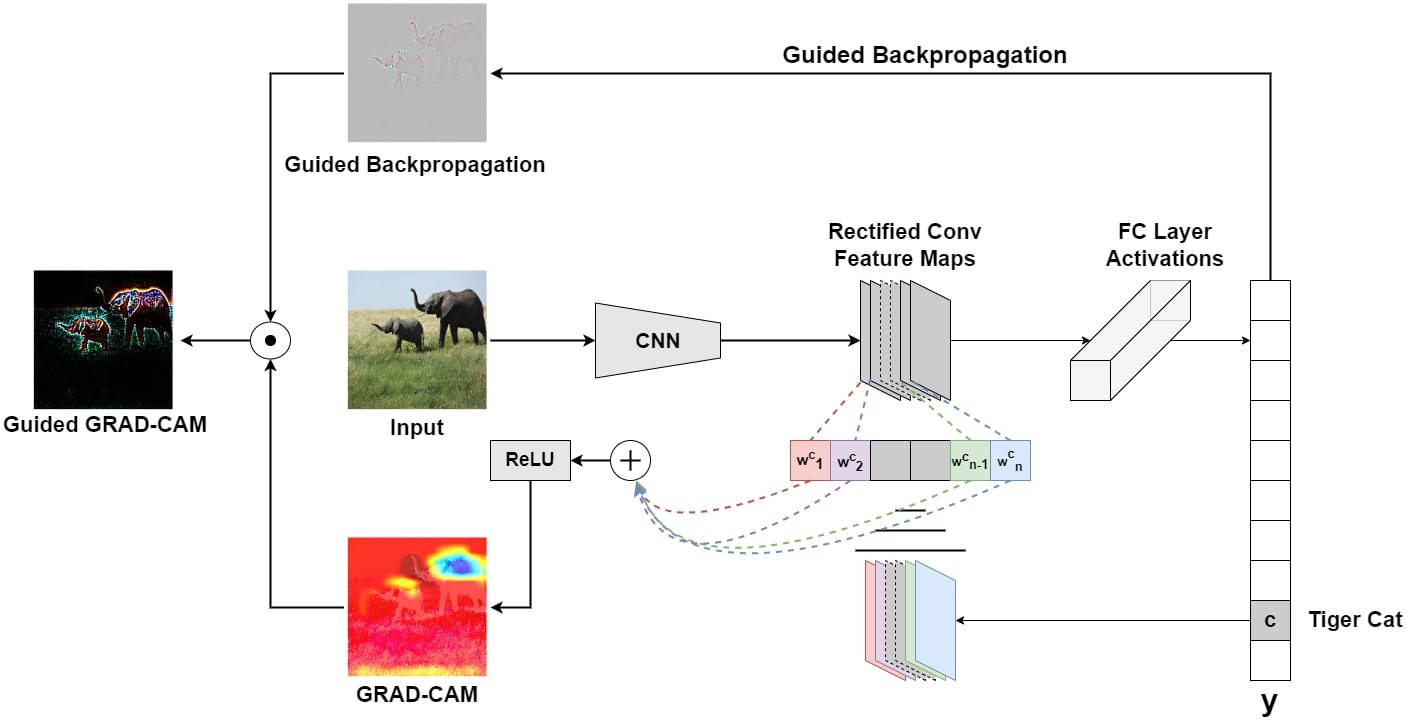}
        \caption{\label{fig:GRAD_CAM}Grad-CAM Overview: Given an image and a class of interest as input, we need to pass the image through a part of the CNN model, and then through calculations depending on the task we choose to get an initial estimate for the category. \cite{DBLP:journals/corr/SelvarajuDVCPB16}}
    \end{figure}
    
\begin{figure}[H]
      \centering
         \subfigure []{
        \label{fig:subfig:original} 
        \includegraphics[scale=0.3]{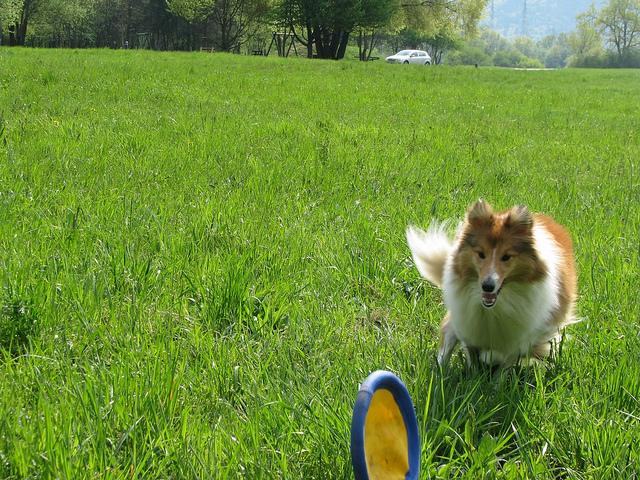}}
         \subfigure []{
        \label{fig:subfig:GFDM} 
        \includegraphics[scale=0.3]{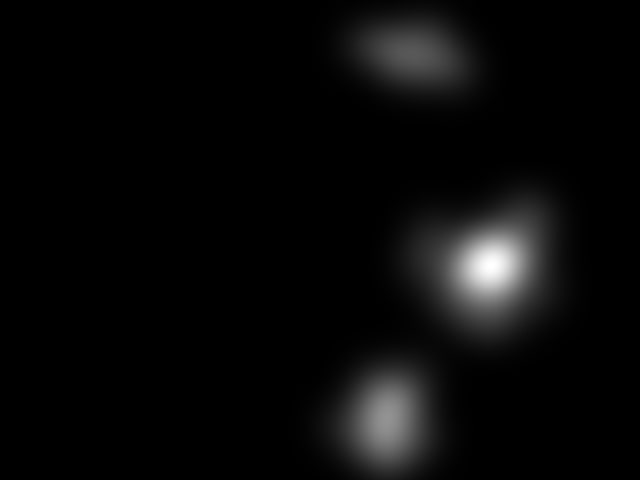}}
         \caption {Illustration of GFDM: (a) Original image, (b) GFDM}
      \label{fig:original_GFDM} 
    \end{figure}

\begin{figure}[H]
      \centering
         \subfigure []{
        \label{fig:subfig:FEM} 
        \includegraphics[scale=0.22]{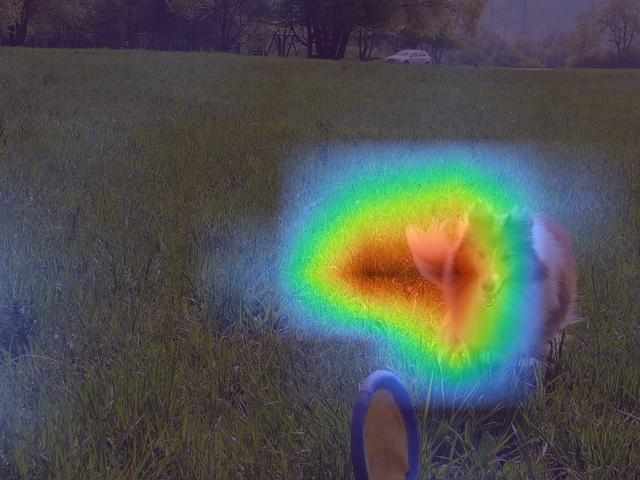}}
         \subfigure []{
        \label{fig:subfig:MLFEM} 
        \includegraphics[scale=0.22]{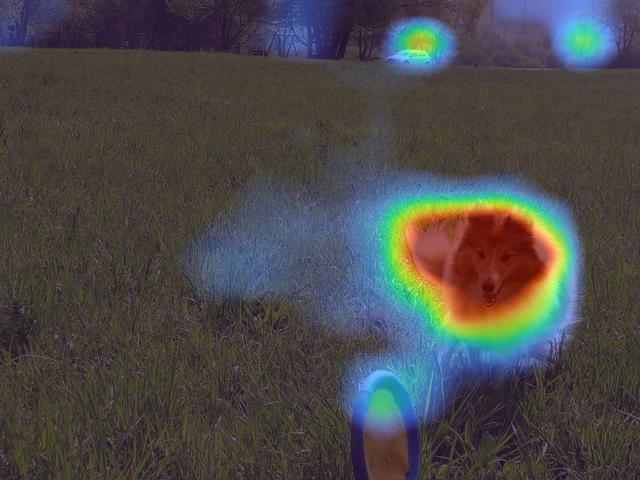}}
         \subfigure []{
        \label{fig:subfig:GRAD_CAM} 
        \includegraphics[scale=0.22]{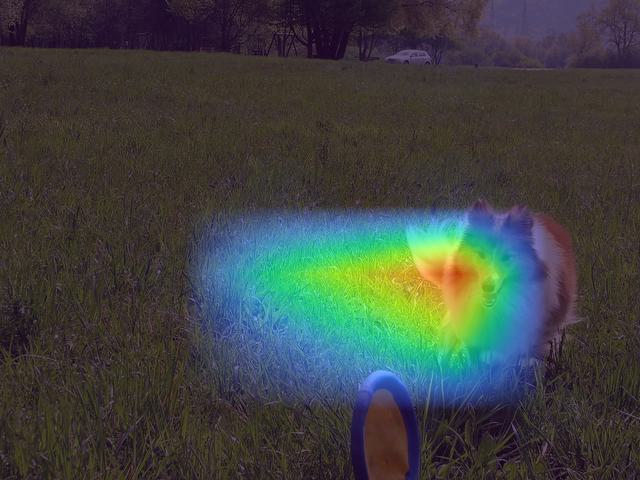}}
         \caption {Explanation maps superimposed on the original frame of Figure \ref{fig:original_GFDM} (a): (a) Superimposed with FEM, (b) Superimposed with MLFEM, (c) Superimposed with Grad-CAM}
      \label{fig:FEM_MLFEM_GRAD_CAM} 
    \end{figure}

    Gradients are set to zero for all classes except the desired class, which will be set to 1. Then this signal is transmitted back to the corrected maps of convolutional features of interest to us, which we combine to calculate the rough localization of Grad-CAM (blue heat map in figure \ref{fig:GRAD_CAM}), which represents where the model we are testing should look, to make a certain decision. Finally, we dot-multiply the heat map using controlled back propagation to obtain a controlled Grad-CAM visualization with high resolution and taking into account a certain concept. The overview of the method is schematized in figure \ref{fig:GRAD_CAM}. To illustrate ExMs obtained with a Grad-CAM method we first show an image with its GFDM In figure \ref{fig:original_GFDM}. Then in figure \ref{fig:FEM_MLFEM_GRAD_CAM} we show maps resulting from the three methods: FEM, MLFEM and Grad-CAM. We can see that the Grad-CAM map is less precise and covers a lot of background.

\section{Generation of Corrupted Images with Controlled Distortions}
        \label{section:Noisy_images}
        To evaluate the influence of distortions on the explainers in image classification tasks, we select a variety of natural images and apply a set of parameterized distortions such as additive Gaussian Noise, Gaussian Blur, Uniform Brightness distortion and Perspective distortion.
        
        \subsection{Additive Gaussian Noise}
        \label{subsection:GaussianNoise}
        
        We consider independent additive Gaussian noise.  Thus, each pixel value will be incremented by a randomly generated shift $\alpha$.  
        
        \begin{equation}
            I'(u,v) = I(u,v) + \alpha(u,v)
            \label{eq:noise}
        \end{equation}
        
        Here, $\alpha(u,v) \sim \frac{1}{\sigma_{agn}\sqrt{2\pi}}\times\exp{\left(-\frac{t^2}{2\times\sigma_{agn}^2}\right)}$ with $t$ is the independent variable, $\sigma_{agn}$ is the scale parameter of the Gaussian distribution.\\

        The same generated shift is applied for each colour channel, but it is different for each pixel. Thus, we consider i.i.d noise processes in each pixel. The effect is the appearance of black-and-white disturbances (noise).
        We parameterize the strength of the noise by the maximal absolute shift value and deduce the $\sigma_{agn}$ parameter of our Gaussian from it. Thus, the maximal magnitude shift $k$ value is the magnitude of the additive Gaussian noise, such that the probability of the noise magnitude to be higher than $k$ is 0.05, accordingly to the two sigma rule of Gaussian distribution.
        To choose the maximal shift value $k$, we propose the following reasoning. \\
        Lipschitz's condition, see equation \ref{eq:Lipschitz} is hold in a certain neighbourhood of the data point $x$. Let us denote the radius of this neighbourhood by $\epsilon$. 
        Thus, the norm of difference of the original image $x$ and corrupted $x'$ satisfies 
    \begin{equation}
        \|{x-x'}\| \leq \varepsilon
        \label{eq:norme}
    \end{equation}
        We will choose $\varepsilon$ in such a way that the channel norm of difference $\|{x_c-x'_c}\|_c = 1,..,n_c$, ($n_c=3$)  satisfies  $\|{x_c-x'_c}\| \leq \gamma$. Thus, $\varepsilon$ will satisfy $\varepsilon = \sqrt{n_c} \gamma$. Channel difference norm is computed as:
        \begin{equation}
            \|{x_c-x'_c}\| = \sqrt{\Sigma_{_{(u,v)\in W\times H}}{\left(x_c(u,v) - x'_c(u,v)\right)}^2}\\
        \label{eq:channel_difference}
        \end{equation}

        Here $x_c(u,v)$ and $x'_c(u,v)$ are the pixel values of original and corrupted image respectively in a colour channel. 
        Thus, we can re-write our channel difference norm as $\|{x_c-x_c'}\| = \sqrt{\sum_{(u,v)\in W\times H}{\alpha_c(u,v)^2}}$, with $\alpha_c(u,v)$ noise value in a pixel of each channel. 
        Let us major $\alpha_c(u,v)^2$ by $k^2$.
        Hence, from equation \ref{eq:norme}, our $k$ should satisfy :
        \begin{equation}
            {k}^2 \times H \times W \leq {\gamma}^2\\
        \label{eq:maxrange}
        \end{equation}\\
        thus being in the range
        \begin{equation}
          -\gamma/\sqrt{H \times W} \leq k \leq \gamma/\sqrt{H \times W}\
          \label{eq:range}
        \end{equation}\\
        consequently, all $\alpha(u,v)$ with $|\alpha(u,v)|\leq |k|$ should be in this range too. In the following, we will omit $(u,v)$. 
        Let us now deduce the scale parameter $\sigma_{agn}$ for our Gaussian distribution for noise generation. 
        Our maximal noise magnitude is $k$, applying "two sigma rule" we can write 
        \begin{equation}
         \sigma_{agn} = k/2 
         \label{eq:sigma}
        \end{equation}\\
        Thus 95\% of generated noise values will be less than $k$ in magnitude. \\
        
        To generate Gaussian noise we will parameterize $k$, and compute $\sigma_{agn}$ and thus will know the neighbourhood radius $\epsilon$ in equation \ref{eq:Lipschitz} too. 
        The algorithm of the generation of each noise value $\alpha(u,v)$ is the following: \\
        
        \textbf{Algorithm}\\
        
        For each pixel $p=(u,v)$ of the original image $I(u,v)$
         \begin{enumerate}[\hspace{1cm}1.]
            \item Generate $Z$ - a random number in the range from 0 to 1
            \item Compute inverse $\alpha (u,v)= F^{-1}(Z)$ of cumulative distribution function $F$ of our Gaussian noise parameterized by $\sigma_{agn}$, see equation \ref{eq:sigma}, for $Z$
            \item If  $|\alpha(u,v)| \geq k/2$, then go to 1
            \item Add $\alpha(u,v)$ accordingly the model of independent additive noise, equation \ref{eq:noise} and crop according to channel range: 
            $I''(u,v) = min(255,max(0,I(u,v)+\alpha(u,v))$
        \end{enumerate}
         Due to the usage of two sigma rule in computation of our $\sigma_{agn}$ parameter from $k$, the internal loops ("go to 1") are rare. \\
         
    An example of images with generated noise for different $k$ parameters is given in figure \ref{fig:GaussianNoise} below. 
    The higher magnitude $k$ of the noise is,  the more the image is corrupted. 
       
       \begin{figure}[H]
      \centering
         \subfigure []{
        \label{fig:subfig:original_GaussianNoise} 
        \includegraphics[scale=0.4]{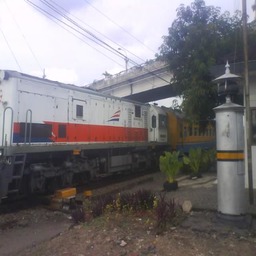}}
         \subfigure []{
        \label{fig:subfig:k50_GaussianNoise} 
        \includegraphics[scale=0.4]{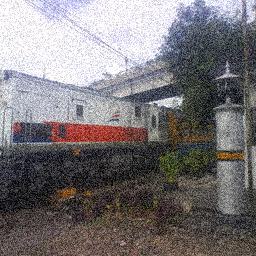}}
        \subfigure []{
        \label{fig:subfig:k125_GaussianNoise} 
        \includegraphics[scale=0.4]{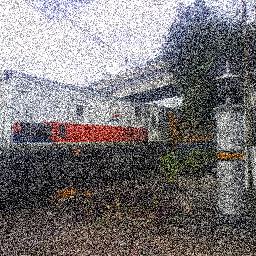}}
         \subfigure []{
        \label{fig:subfig:k200_GaussianNoise} 
        \includegraphics[scale=0.4]{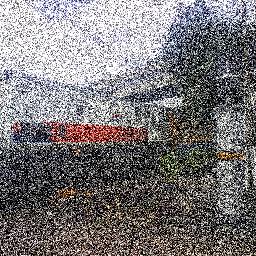}}
         \caption {Original image corrupted with Gaussian noise with different maximal shift parameter:(a) original image, (b) k=50, (c) k=125, (d) k=200}
      \label{fig:GaussianNoise} 
    \end{figure}
    
    \subsection{Gaussian Blur}
    \label{subsection:GaussianBlur}
    The second distortion we apply is the Gaussian blur. 
     \begin{equation}
            I'(u,v) = I(u,v)*g(\mu,\nu) 
            \label{eq:blur}
        \end{equation}
        here, $g(\mu,\nu)$ is the Gaussian filter kernel : $g(\mu,\nu)= \frac{1}{A}\times\exp{\left(-\frac{\mu^2+\nu^2}{2\times\sigma_{gb}^2}\right)}$ with $A$ normalization factor and $*$ is a convolution operation.
        We vary the scale parameter $\sigma_{gb}$ of the Gaussian filter and the size $s$ of the filter mask to generate the same number of corrupted images as for Gaussian noise distortion \ref{subsection:GaussianNoise}. The same filter $g$ is applied to three components R,G,B of colour images. We give examples of blurred images in figures \ref{fig:GaussianBlur_1.5}, 
        and \ref{fig:GaussianBlur_6} below.

    \begin{figure}[H]
        \centering
        \subfigure []{
            \label{fig:subfig:original_GaussianBlur_1.5} 
            \includegraphics[scale=0.31]{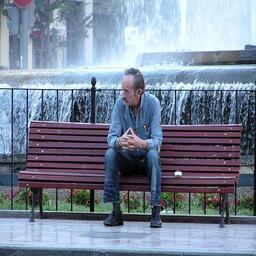}}
        \subfigure []{
            \label{fig:subfig:5x5_GaussianBlur_1.5} 
            \includegraphics[scale=0.31]{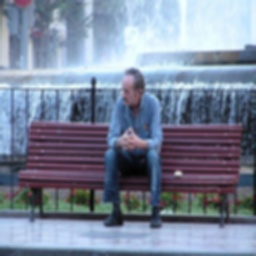}}
        \subfigure []{
            \label{fig:subfig:7x7_GaussianBlur_1.5} 
            \includegraphics[scale=0.31]{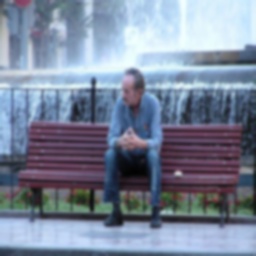}}
        \subfigure []{
            \label{fig:subfig:9x9_GaussianBlur_1.5} 
            \includegraphics[scale=0.31]{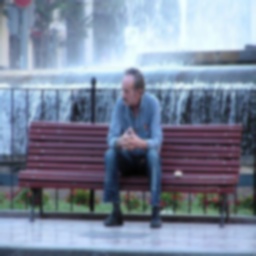}}
        \subfigure []{
            \label{fig:subfig:11x11_GaussianBlur_1.5} 
            \includegraphics[scale=0.31]{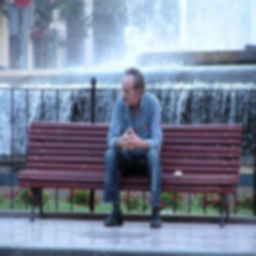}}
        \caption {Original image (a) corrupted with Gaussian blur with the same $\sigma_{gb} =  $ 1.5 and different kernel sizes: (b) 5x5, (b) 7x7, (d) 9x9, (e) 11x11}
        \label{fig:GaussianBlur_1.5} 
    \end{figure}

    \begin{figure}[H]
        \centering
        \subfigure []{
            \label{fig:subfig:original_GaussianBlur_6} 
            \includegraphics[scale=0.31]{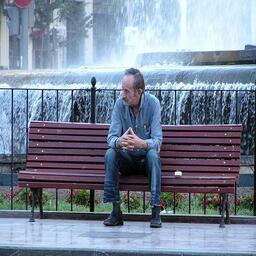}}
        \subfigure []{
            \label{fig:subfig:5x5_GaussianBlur_6} 
            \includegraphics[scale=0.31]{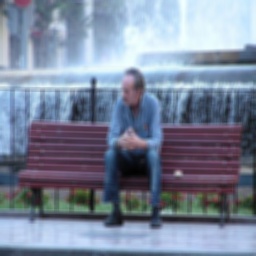}}
        \subfigure []{
            \label{fig:subfig:7x7_GaussianBlur_6} 
            \includegraphics[scale=0.31]{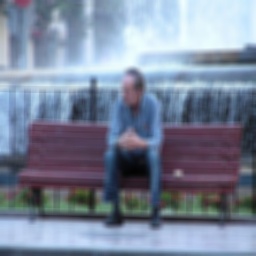}}
        \subfigure []{
            \label{fig:subfig:9x9_GaussianBlur_6} 
            \includegraphics[scale=0.31]{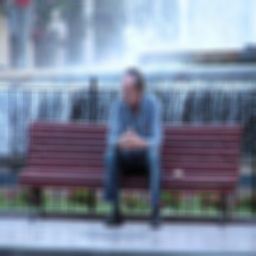}}
        \subfigure []{
            \label{fig:subfig:11x11_GaussianBlur_6} 
            \includegraphics[scale=0.31]{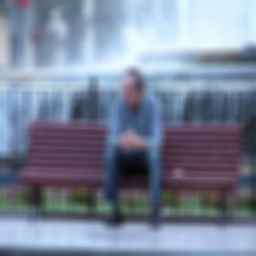}}
        \caption {Original image (a) corrupted with Gaussian blur with one $\sigma_{gb} = $ 6 and different kernel parameters: (b) 5x5, (c) 7x7, (d) 9x9, (e) 11x11}
        \label{fig:GaussianBlur_6} 
    \end{figure}

    \subsection{Uniform Brightness Distortion}
    \label{subsection:uniforBrightness}
\begin{figure}[H]
      \centering
         \subfigure []{
        \label{fig:subfig:original_brightness_white} 
        \includegraphics[scale=0.31]{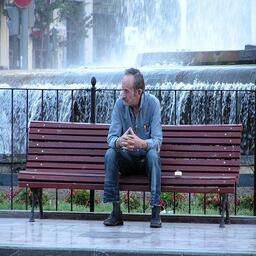}}
         \subfigure []{
        \label{fig:subfig:k50_brightness_white} 
        \includegraphics[scale=0.31]{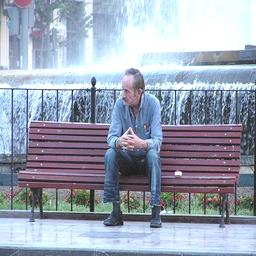}}
        \subfigure []{
        \label{fig:subfig:k125_brightness_white} 
        \includegraphics[scale=0.31]{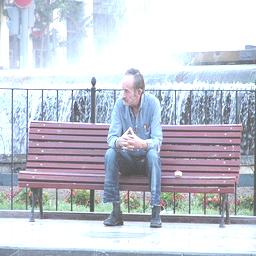}}
         \subfigure []{
        \label{fig:subfig:k200_brightness_white} 
        \includegraphics[scale=0.3]{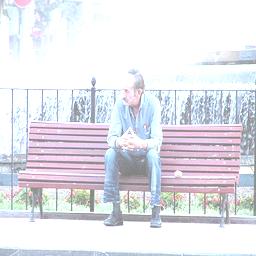}}
         \caption {Original image corrupted with Brightness with different maximal shift parameter: (a) original image, (b) k=50, (c) k=125, (d) k=200}
      \label{fig:brightness_white} 
    \end{figure}

    A uniform brightness distortion consists in adding a random shift $\beta$ to all three-colour components of the image. The choice of $\beta$ value is realized randomly accordingly to the method described in \ref{subsection:GaussianNoise} parameterized by the maximal shift magnitude parameter $k$. The new colour value in each channel is computed as  
    \begin{equation}
       I'(u,v) = min\left(255,max\left(0,I\left(u,v\right)+\beta\right)\right) 
    \end{equation}
    Examples of brightness distortion are given in figure \ref{fig:brightness_white}
    
    \subsection{Perspective Distortion}
    \label{subsection:PerspectiveDistorsion}
    Finally, we apply the geometric distortion of image plane applying a perspective transformation $F$ : $I'(u',v') = I(F(u,v))$, as it is the most general case, compared to e.g. pure zoom transformation. It is expressed as 
    \begin{equation}
    \label{eq:homography}
     \textbf{u}' = \textbf{H}\times\textbf{u}  
    \end{equation}
    in homogeneous coordinates, see \cite{Szel'2022} for more details.
    The eight parameters of the homography matrix \textbf{H} are defined by pairs of corresponding points in the source and target images $\{(u_1,v_1),(u'_1,v'_1)\},\\ $..., $\{(u_4,v_4),(u'_4,v'_4)\}$. 
    Note, we used here OpenCv\footnote{https://docs.opencv.org/4.x/d9/dab/tutorial\textunderscore homography.html} implementation of Homography estimation. 
   To parameterize these transforms, we use a trapeze figure with the bases parallel to the horizontal image border. Its upper base $\{(u_1,v_1)\},...,\{(u_2,v_2)\}$ is shorter than the lower one $\{(u_3,v_3)\},...,\{(u_4,v_4)\}$. Then it is rotated three times by 90°. To get the target four points $\{(u'_1,v'_1)\},...,\{(u'_4,v'_4)\}$, we scale the original coordinates of trapeze with a "zoom factor" $l>1$. 
   The values of missing pixels in the target image are bi-linearly interpolated. In this way we generate perspective distortion without holes in the target image which would yield parasite features when passing through a CNN classifier. 
   Examples of distorted images are illustrated in figure 
   \ref{fig:Change_Perspective_10} below.

    \begin{figure}[H]
        \centering
        \subfigure []{
            \label{fig:subfig:original_PerspectiveDistorsion} 
            \includegraphics[scale=0.31]{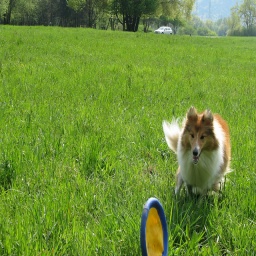}}
        \subfigure []{
            \label{fig:subfig:top1_PerspectiveDistorsion} 
            \includegraphics[scale=0.417]{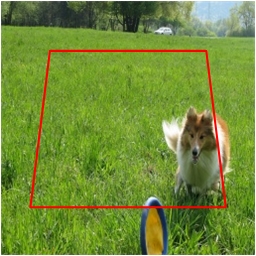}}
        \subfigure []{
            \label{fig:subfig:top2_PerspectiveDistorsion} 
            \includegraphics[scale=0.31]
            {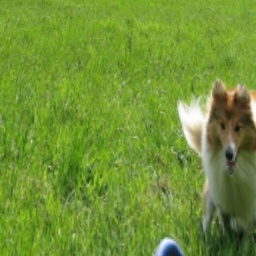}}
        \subfigure []{
            \label{fig:subfig:bottom1_PerspectiveDistorsion} 
            \includegraphics[scale=0.417]{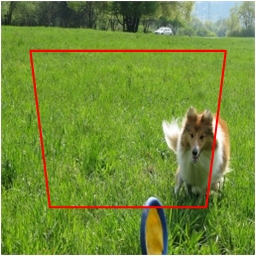}}
        \subfigure []{
            \label{fig:subfig:bottom2_PerspectiveDistorsion} 
            \includegraphics[scale=0.31]{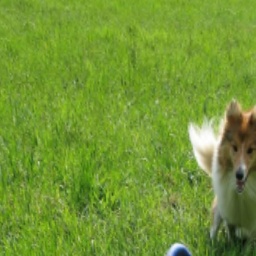}}
        \subfigure []{
            \label{fig:subfig:left1_PerspectiveDistorsion} 
            \includegraphics[scale=0.417]{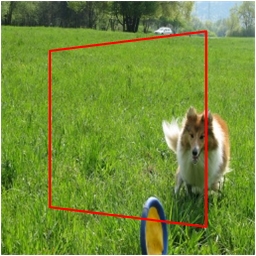}}
        \subfigure []{
            \label{fig:subfig:left2_PerspectiveDistorsion} 
            \includegraphics[scale=0.31]{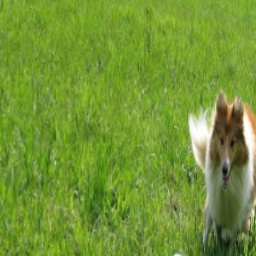}}
        \subfigure []{
            \label{fig:subfig:right1_PerspectiveDistorsion} 
            \includegraphics[scale=0.417]{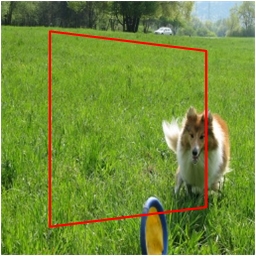}}
        \subfigure []{
            \label{fig:subfig:right2_PerspectiveDistorsion} 
            \includegraphics[scale=0.31]{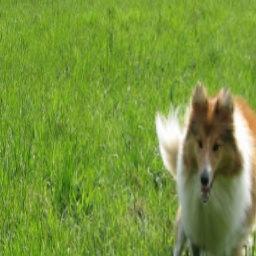}}
        \caption {Original image (a) corrupted with Change of Perspective with one zoom l=10 and different orientations of trapeze: (a) $->$ (b) = (c) top, (a) $->$ (d) = (e) bottom, (a) $->$ (f) = (g) left, (a) $->$ (h) = (i) right}
        \label{fig:Change_Perspective_10} 
    \end{figure}

\section{Complete Result Tables}

\begin{table}[!tb]
        \centering
        \caption{Additive Gaussian noise results}
        \subtable[Gaussian noise: Stability of Lipschitz constant for FEM, MLFEM, GRAD-CAM as a function of noise level         \label{tab:L_Mean_Sigma_tab_Gaussian_noise}]{
            \setlength{\tabcolsep}{1.0mm}
            \begin{tabular}{c|c|c|c|c|c|c}
                \hline
                {} & \multicolumn{2}{c|}{\textbf{FEM}} & \multicolumn{2}{|c|}{\textbf{MLFEM}} & \multicolumn{2}{|c}{\textbf{GRAD-CAM}}\\
                \hline
                {\textbf{$k$}} & {\textbf{Well}} & {\textbf{Badly}} & {\textbf{Well}} & {\textbf{Badly}} & {\textbf{Well}} & {\textbf{Badly}}\\
                \hline 
                \hline
                25  $\rightarrow$ 50  &  40.063\% & 42.828\%            & 42.268\% & 38.370\%            & 36.276\% & 44.419\%\\[5pt]
                50  $\rightarrow$ 75  &  29.169\% & 27.591\%            & 28.073\% & 25.691\%           & 29.982\% & 32.198\%\\[5pt]
                75  $\rightarrow$ 100  &  23.213\% & 19.062\%           & 18.279\% & 19.487\%           & 14.469\% & 20.218\%\\[5pt]
                100 $\rightarrow$ 125  &  10.623\% & 13.929\%           & 14.644\% & 14.946\%           & 17.408\% & 15.661\%\\[5pt]
                125 $\rightarrow$ 150  & 13.698\% & 12.524\%            & 14.526\% & 14.868\%           & 14.540\% & 12.274\%\\[5pt]
                150 $\rightarrow$ 175  &  \textbf{7.242\%} & 12.045\%   & 15.315\% & \textbf{9.046\%}   & 8.183\% & 10.446\% \\[5pt]
                175 $\rightarrow$ 200  &  12.085\% & 10.711\%           & 15.792\% & 9.743\%            & 9.712\% & 10.435\% \\[5pt]
                \hline
            \end{tabular}
        }

    \subtable[Gaussian noise: Stability PCC for FEM, MLFEM, GRAD-CAM as a function of noise
        \label{tab:PCC_Mean_Sigma_tab_Gaussian_noise}]{
  
            \begin{tabular}{c|c|c|c|c|c|c}
                \hline
                {} & \multicolumn{2}{c|}{\textbf{FEM}} & \multicolumn{2}{|c|}{\textbf{MLFEM}} & \multicolumn{2}{|c}{\textbf{GRAD-CAM}}\\
                \hline
                {\textbf{$k$}} & {\textbf{Well}} & {\textbf{Badly}} & {\textbf{Well}} & {\textbf{Badly}} & {\textbf{Well}} & {\textbf{Badly}}\\
                \hline 
                \hline
                25 $\rightarrow$ 50 & 4.576\% & 5.437\%             & 14.935\% & 20.997\%               & 4.459\% & 12.951\% \\[5pt]
                50 $\rightarrow$ 75 & 1.894\% & 3.279\%             & 5.539\% & 18.955\%                & 25.339\% & 0.201\% \\[5pt]
                75 $\rightarrow$ 100 & 1.274\% & 3.250\%             & 3.032\% & 4.273\%                 & 14.707\% & 1.067\%\\[5pt]
                100 $\rightarrow$ 125 & 5.464\% & 2.994\%           & 18.814\% & 7.069\%                & 30.193\% & 16.604\%\\[5pt]
                125 $\rightarrow$ 150 & \textbf{0.533\%} & 2.820\%   & 2.593\% & 6.378\%                 & 30.427\% & 13.620\% \\[5pt]
                150 $\rightarrow$ 175 & 3.560\% & 1.535\%            & 11.594\% & 16.142\%               & 59.305\% & 12.626\%\\[5pt]
                175 $\rightarrow$ 200 & 2.486\% & \textbf{0.083\%}  & 14.685\% & 6.069\%                & 6.525\% & 19.017\% \\[5pt]
                \hline
            \end{tabular}
        }

    \subtable[Gaussian noise: Stability of SIM for FEM, MLFEM, GRAD-CAM as a function of noise.
        \label{tab:SIM_Mean_Sigma_tab_Gaussian_noise}]{

            \begin{tabular}{c|c|c|c|c|c|c}
                \hline
                {} & \multicolumn{2}{c|}{\textbf{FEM}} & \multicolumn{2}{|c|}{\textbf{MLFEM}} & \multicolumn{2}{|c}{\textbf{GRAD-CAM}}\\
                \hline
                {\textbf{$k$}} & {\textbf{Well}} & {\textbf{Badly}} & {\textbf{Well}} & {\textbf{Badly}} & {\textbf{Well}} & {\textbf{Badly}}\\
                \hline 
                \hline
                25 $\rightarrow$ 50 & 2.058\% & 2.508\%     & 3.007\% & 5.767\%             & 5.592\% & 0.456\%\\[5pt]
                50 $\rightarrow$ 75 & 0.769\% & 1.323\%     & 1.402\% & 4.191\%             & 3.805\% & \textbf{0.064\%} \\[5pt]
                75 $\rightarrow$ 100 & 0.586\% & 1.213\%    & 3.559\% & 1.030\%              & 0.138\% & 1.883\%\\[5pt]
                100 $\rightarrow$ 125 & 1.799\% & 1.244\%   & 2.387\% & 1.795\%             & 3.994\% & 1.034\%\\[5pt]
                125 $\rightarrow$ 150 & 0.925\% & 0.590\%    & \textbf{0.073\%}              & 0.222\% & 6.573\% & 1.236\%\\[5pt]
                150 $\rightarrow$ 175 & 1.333\% & 0.573\%   & 1.563\% & 1.543\%             & 11.514\% & 0.474\%\\[5pt]
                175 $\rightarrow$ 200 & 0.696\% & 0.190\%    & 2.220\% & 0.174\%     & 8.181\% & 1.039\%\\[5pt]
                \hline
            \end{tabular}
        }

    \end{table}

    \begin{table}[!tb]
        \centering
        \caption{Gaussian Blur Results}

        \subtable[Gaussian blur: Stability of Lipschitz constant for FEM, MLFEM, GRAD-CAM as a function of distortion level\label{tab:L_Mean_Sigma_tab_Gaussian_blur}]{

            \begin{tabular}{c|c|c|c|c|c|c}
                \hline
                {} & \multicolumn{2}{c|}{\textbf{FEM}} & \multicolumn{2}{|c|}{\textbf{MLFEM}} & \multicolumn{2}{|c}{\textbf{GRAD-CAM}}\\
                \hline
                {\textbf{$\sigma_{gb}$}} & {\textbf{Well}} & {\textbf{Badly}} & {\textbf{Well}} & {\textbf{Badly}} & {\textbf{Well}} & {\textbf{Badly}}\\
                \hline 
                \hline
                        1.25 $\rightarrow$ 1.50  & 1.725\% & 4.426\%  & 4.373\% & 8.293\%          & 8.817\% & 12.304\%\\[1pt]
                        1.50 $\rightarrow$ 1.75  & 1.776\% & 3.408\%  & 1.558\% & 4.476\%          & 2.829\% & 1.117\%\\[1pt]
                        1.75 $\rightarrow$ 2.00  & 0.185\% & 3.713\%  & 1.954\% & 4.842\%           & \textbf{0.046\%} & 3.225\%\\[1pt]
                        2.00 $\rightarrow$ 2.50  & 2.340\% & 10.011\%  & 1.752\% & 5.232\%           & 1.653\% & 7.513\%\\[1pt]
                        2.50 $\rightarrow$ 3.00  & 8.649\% & 1.750\%   & 6.638\% & 3.511\%           & 4.244\% & 5.656\%\\[1pt]
                        3.00 $\rightarrow$ 3.50  & 0.234\% & 3.778\%  & 0.595\% & 1.057\%           & 0.385\% & 2.997\%\\[1pt]
                        3.50 $\rightarrow$ 4.00  & 2.676\% & 5.148\%  & 0.661\% & 2.885\%           & 0.815\% & 1.906\%\\[1pt]
                        4.00 $\rightarrow$ 5.00  & 6.639\% & 3.400\%    & 4.413\% & 0.988\%           & 3.258\% & 0.953\%\\[1pt]
                        5.00 $\rightarrow$ 6.00  & 0.531\% & \textbf{0.646\%}  & 0.656\% & 1.084\%           & 1.436\% & 2.296\%\\[1pt]
                \hline
            \end{tabular}
            }

            \subtable[Gaussian blur: Stability PCC for FEM, MLFEM, GRAD-CAM as function of distortion level\label{tab:PCC_Mean_Sigma_tab_Gaussian_blur}]{
     
            \begin{tabular}{c|c|c|c|c|c|c}
                \hline
                {} & \multicolumn{2}{c|}{\textbf{FEM}} & \multicolumn{2}{|c|}{\textbf{MLFEM}} & \multicolumn{2}{|c}{\textbf{GRAD-CAM}}\\
                \hline
                {\textbf{$\sigma_{gb}$}} & {\textbf{Well}} & {\textbf{Badly}} & {\textbf{Well}} & {\textbf{Badly}} & {\textbf{Well}} & {\textbf{Badly}}\\
                \hline 
                \hline
                        1.25 $\rightarrow$ 1.50  & 2.465\% & 2.119\%    & 0.145\% & 3.332\%            & 4.527\% & 20.823\%\\[1pt]
                        1.50 $\rightarrow$ 1.75  & 1.730\% & 4.205\%     & 9.935\% & 2.968\%              & 3.255\% & 17.881\%\\[1pt]
                        1.75 $\rightarrow$ 2.00  & 1.761\% & 1.410\%     & \textbf{0.044\%} & 2.741\%             & 2.221\% & 6.288\%\\[1pt]
                        2.00 $\rightarrow$ 2.50  & 1.080\% & 0.464\%     & 3.492\% & 4.004\%             & 3.183\% & 12.659\%\\[1pt]
                        2.50 $\rightarrow$ 3.00  & 1.109\% & 1.061\%    & 4.492\% & 6.458\%             & 7.210\% & 0.645\%\\[1pt]
                        3.00 $\rightarrow$ 3.50  & 0.422\% & 0.558\%    & 0.718\% & 11.756\%            & 5.952\% & 2.601\%\\[1pt]
                        3.50 $\rightarrow$ 4.00  & 2.453\% & 1.679\%    & 5.461\% & 4.806\%             & 4.277\% & 2.713\%\\[1pt]
                        4.00 $\rightarrow$ 5.00  & 2.186\% & 1.708\%    & 3.379\% & 1.347\%              & 1.366\% & 5.076\%\\[1pt]
                        5.00 $\rightarrow$ 6.00  & 0.215\% & \textbf{0.261\%}    & 4.969\% & 3.660\%             & 1.596\% & 2.853\%\\[1pt]
                \hline
            \end{tabular}
       }

        \subtable[Gaussian blur: Stability of SIM for FEM, MLFEM, GRAD-CAM as a function of distortion level
        \label{tab:SIM_Mean_Sigma_tab_Gaussian_blur}]{

            \begin{tabular}{c|c|c|c|c|c|c}
                \hline
                {} & \multicolumn{2}{c|}{\textbf{FEM}} & \multicolumn{2}{|c|}{\textbf{MLFEM}} & \multicolumn{2}{|c}{\textbf{GRAD-CAM}}\\
                \hline
                {\textbf{$\sigma_{gb}$}} & {\textbf{Well}} & {\textbf{Badly}} & {\textbf{Well}} & {\textbf{Badly}} & {\textbf{Well}} & {\textbf{Badly}}\\
                \hline 
                \hline
                        1.25 $\rightarrow$ 1.50  & 0.995\% & 0.932\%    & 1.339\% & 0.165\%            & 1.653\% & 1.073\%\\[1pt]
                        1.50 $\rightarrow$ 1.75  & 0.589\% & 0.897\%    & 1.028\% & \textbf{0.063\%}             & 1.208\% & 0.405\%\\[1pt]
                        1.75 $\rightarrow$ 2.00  & 0.540\% & 0.250\%      & 0.375\% & 0.479\%             & 1.346\% & 0.292\% \\[1pt]
                        2.00 $\rightarrow$ 2.50  & 0.089\% & 0.166\%    & 0.104\% & 1.357\%             & 0.678\% & 4.564\%\\[1pt]
                        2.50 $\rightarrow$ 3.00  & 0.593\% & 0.538\%    & 0.639\% & 0.438\%             & 1.425\% & 0.724\%\\[1pt]
                        3.00 $\rightarrow$ 3.50  & 0.120\% & 0.306\%     & 0.076\% & 1.241\%              & 2.626\% & 0.779\%\\[1pt]
                        3.50 $\rightarrow$ 4.00  & 0.551\% & 0.360\%     & \textbf{0.020\%} & 0.083\%            & 1.302\% & 0.377\%\\[1pt]
                        4.00 $\rightarrow$ 5.00  & 0.769\% & 0.221\%    & 0.091\% & 0.221\%               & 0.881\% & 0.639\%\\[1pt]
                        5.00 $\rightarrow$ 6.00  & 0.236\% & 0.120\%     & 0.682\% & 0.604\%             & 0.693\% & 0.674\%\\[1pt]
                \hline
            \end{tabular}
        }
        
    \end{table}

    \begin{table}[!tb]
        \centering
        \caption{Uniform Brightness Results}
        \subtable[Uniform Brightness Distortion: Stability of Lipschitz constant for FEM, MLFEM, GRAD-CAM as a function of distortion level
        \label{tab:L_Mean_Sigma_tab_Uniform_Brightness_Distortion}]{
      
            \begin{tabular}{c|c|c|c|c|c|c}
                \hline
                {} & \multicolumn{2}{c|}{\textbf{FEM}} & \multicolumn{2}{|c|}{\textbf{MLFEM}} & \multicolumn{2}{|c}{\textbf{GRAD-CAM}}\\
                \hline
                {\textbf{$\beta$}} & {\textbf{Well}} & {\textbf{Badly}} & {\textbf{Well}} & {\textbf{Badly}} & {\textbf{Well}} & {\textbf{Badly}}\\
                \hline 
                \hline
                25  $\rightarrow$ 50  & 33.290\% & 45.451\%             & 35.912\% & 33.090\%       & 33.640\% & 37.839\%\\[5pt]
                50  $\rightarrow$ 75 & 27.622\% & 15.289\%               & 27.082\% & 27.643\%         & 27.936\% & 30.533\%\\[5pt]
                75  $\rightarrow$ 100  & 21.495\% & 15.146\%             & 21.912\% & 15.742\%       & 21.992\% & 14.419\%\\[5pt]
                100 $\rightarrow$ 125  & 12.474\% & 36.293\%            & 16.462\% & 25.924\%       & 10.969\% & 31.124\%\\[5pt]
                125 $\rightarrow$ 150  & 16.357\% & 9.060\%              & 14.389\% & 10.996\%         & 14.839\% & 15.203\%\\[5pt]
                150 $\rightarrow$ 175  & 5.678\% & 20.702\%            & 12.191\% & 17.272\%       & 11.275\% & 24.050\%\\[5pt]
                175 $\rightarrow$ 200  & 11.970\% & \textbf{6.384\%}             & 9.656\% & 11.784\%        & \textbf{1.663\%} & 7.522\%\\[5pt]
                \hline
            \end{tabular}
       }
        
    \subtable[Uniform Brightness Distortion: Stability of PCC for FEM, MLFEM, GRAD-CAM as a function of distortion level
        \label{tab:PCC_Mean_Sigma_tab_Uniform_Brightness_Distortion}]{

            \begin{tabular}{c|c|c|c|c|c|c}
                \hline
                {} & \multicolumn{2}{c|}{\textbf{FEM}} & \multicolumn{2}{|c|}{\textbf{MLFEM}} & \multicolumn{2}{|c}{\textbf{GRAD-CAM}}\\
                \hline
                {\textbf{$\beta$}} & {\textbf{Well}} & {\textbf{Badly}} & {\textbf{Well}} & {\textbf{Badly}} & {\textbf{Well}} & {\textbf{Badly}}\\
                \hline 
                \hline
                25 $\rightarrow$ 50 & 1.224\% & 0.453\%        & 0.822\% & 11.190\%        & 2.562\% & 8.081\%\\[5pt]
                50 $\rightarrow$ 75 & \textbf{0.073\%} & 3.694\%        & 2.594\% & \textbf{0.366\%}         & 0.987\% & 12.738\%\\[5pt]
                75 $\rightarrow$ 100 & 0.916\% & 1.885\%       & 1.225\% & 5.152\%        & 3.206\% & 5.804\% \\[5pt]
                100 $\rightarrow$ 125 & 1.064\% & 0.904\%      & 5.013\% & 0.655\%         & 5.271\% & 10.246\%\\[5pt]
                125 $\rightarrow$ 150 & 0.506\% & 2.623\%       & 0.214\% & 7.326\%        & 4.897\% & 2.190\% \\[5pt]
                150 $\rightarrow$ 175 & 2.006\% & 1.061\%      & 0.762\% & 14.372\%        & 5.489\% & 1.482\%\\[5pt]
                175 $\rightarrow$ 200 & 0.505\% & 3.730\%       & 3.244\% & 1.886\%         & 4.996\% & 2.685\%\\[5pt]
                \hline
            \end{tabular}
        }

    \subtable[Uniform Brightness Distortion: Stability of SIM for FEM, MLFEM, GRAD-CAM as a function of distortion level \label{tab:SIM_Mean_Sigma_tab_Uniform_Brightness_Distortion}]{
            \begin{tabular}{c|c|c|c|c|c|c}
                \hline
                {} & \multicolumn{2}{c|}{\textbf{FEM}} & \multicolumn{2}{|c|}{\textbf{MLFEM}} & \multicolumn{2}{|c}{\textbf{GRAD-CAM}}\\
                \hline
                {\textbf{$\beta$}} & {\textbf{Well}} & {\textbf{Badly}} & {\textbf{Well}} & {\textbf{Badly}} & {\textbf{Well}} & {\textbf{Badly}}\\
                \hline 
                \hline
                25 $\rightarrow$ 50 & 0.552\% & 0.165\%         & 0.805\% & 2.882\%             & 0.440\% & 0.410\%\\[5pt]
                50 $\rightarrow$ 75 & \textbf{0.098\%} & 1.507\%        & 1.118\% & \textbf{0.017\%}              & 0.548\% & 1.567\%\\[5pt]
                75 $\rightarrow$ 100 & 0.417\% & 0.708\%        & 0.536\% & 0.606\%            & 1.185\% & 1.511\% \\[5pt]
                100 $\rightarrow$ 125 & 0.456\% & 0.502\%         & 0.962\% & 0.408\%      & 2.684\% & 0.938\%\\[5pt]
                125 $\rightarrow$ 150 & 0.215\% & 1.469\%        & 0.316\% & 2.237\%     & 1.895\% & 0.332\%\\[5pt]
                150 $\rightarrow$ 175 & 0.941\% & 0.622\%       & 0.206\% & 3.682\%             & 1.697\% & 0.054\%\\[5pt]
                175 $\rightarrow$ 200 & 0.266\% & 1.577\%       & 1.058\% & 0.618\%             & 0.936\% & 1.190\%\\[5pt]
                \hline
            \end{tabular}
    }

    \end{table}

    \begin{table}[!tb]
        \centering
        \caption{Perspective Distortion results.}
        \subtable[Perspective Distortion: Stability of Lipschitz constant for FEM, MLFEM, GRAD-CAM as a function of distortion level
        \label{tab:L_Mean_Sigma_tab_Perspective_Distortion}]{
            \begin{tabular}{c|c|c|c|c|c|c}
                \hline
                {} & \multicolumn{2}{c|}{\textbf{FEM}} & \multicolumn{2}{|c|}{\textbf{MLFEM}} & \multicolumn{2}{|c}{\textbf{GRAD-CAM}}\\
                \hline
                {\textbf{$l$}} & {\textbf{Well}} & {\textbf{Badly}} & {\textbf{Well}} & {\textbf{Badly}} & {\textbf{Well}} & {\textbf{Badly}}\\
                \hline 
                \hline
                1 $\rightarrow$ 2  & 5.097\% & 8.216\%              & 1.895\% & 1.757\%          & 4.861\% & 0.292\%\\[5pt]
                2 $\rightarrow$ 3  & 0.243\% & 7.539\%              & 2.240\% & 3.882\%         & 0.203\% & 1.593\%\\[5pt]
                3 $\rightarrow$ 4  & 2.741\% & 9.974\%               & \textbf{0.111\%} & 8.512\%          & 2.051\% & 2.010\% \\[5pt]
                4 $\rightarrow$ 5  & 1.553\% & 12.908\%              & 2.542\% & 8.023\%         & 2.794\% & 8.202\%\\[5pt]
                5 $\rightarrow$ 6  & 4.260\% & \textbf{0.198\%}              & 2.211\% & 4.200\%           & 2.539\% & 3.313\%\\[5pt]
                6 $\rightarrow$ 7  & 2.759\% & 3.260\%             & 1.177\% & 1.329\%         & 12.827\% & 7.982\%\\[5pt]
                7 $\rightarrow$ 8  & 5.036\% & 3.745\%             & 0.348\% & 0.402\%         & 6.508\% & 2.647\%\\[5pt]
                8 $\rightarrow$ 9  & 3.978\% & 3.001\%              & 2.232\% & 3.194\%           & 3.971\% & 4.048\%\\[5pt]
                9 $\rightarrow$ 10 & 5.786\% & 5.913\%               & 0.652\% & 0.330\%        & 2.900\% & 2.034\%\\[5pt]
                \hline
            \end{tabular}
        }
        
    \subtable[Perspective Distortion: Stability of PCC for FEM, MLFEM, GRAD-CAM as a function of distortion level
        \label{tab:PCC_Mean_Sigma_tab_Perspective_Distortion}]{
            \begin{tabular}{c|c|c|c|c|c|c}
                \hline
                {} & \multicolumn{2}{c|}{\textbf{FEM}} & \multicolumn{2}{|c|}{\textbf{MLFEM}} & \multicolumn{2}{|c}{\textbf{GRAD-CAM}}\\
                \hline
                {\textbf{$l$}} & {\textbf{Well}} & {\textbf{Badly}} & {\textbf{Well}} & {\textbf{Badly}} & {\textbf{Well}} & {\textbf{Badly}}\\
                \hline 
                \hline
                1 $\rightarrow$ 2  & 0.515\% & 0.685\%             & 3.667\% & 6.640\%         & 4.414\% & 8.068\%\\[5pt]
                2 $\rightarrow$ 3 & 2.449\% & 1.172\%              & 10.124\% & 3.000\%         & 5.757\% & 22.944\%\\[5pt]
                3 $\rightarrow$ 4  & \textbf{0.138\%} & 5.063\%              & 0.723\% & 12.104\%        & 1.672\% & 20.077\%\\[5pt]
                4 $\rightarrow$ 5 & 1.797\% & 0.525\%              & 8.523\% & 3.104\%        & 1.945\% & 25.778\%\\[5pt]
                5 $\rightarrow$ 6  & 0.923\% & 0.822\%            & 1.339\% & 7.923\%        & 0.265\% & 13.342\%\\[5pt]
                6 $\rightarrow$ 7  & 0.432\% & 3.130\%             & 4.040\% & 12.268\%         & 6.379\% & 34.812\%\\[5pt]
                7 $\rightarrow$ 8  & 4.545\% & 0.266\%             & 5.541\% & 13.854\%         & 3.934\% & 24.610\% \\[5pt]
                8 $\rightarrow$ 9  & 1.499\% & 2.039\%              & 8.666\% & 2.983\%         & 2.432\% & 17.874\% \\[5pt]
                9 $\rightarrow$ 10  & 5.508\% & 1.057\%             & 1.538\% & \textbf{0.047\%}         & 6.594\% & 5.572\%\\[5pt]
                \hline
            \end{tabular}
       }
       
    \subtable[Perspective Distortion: Stability of SIM for FEM, MLFEM, GRAD-CAM as a function of distortion level
        \label{tab:SIM_Mean_Sigma_tab_Perspective_Distortion}]{
            \begin{tabular}{c|c|c|c|c|c|c}
                \hline
                {} & \multicolumn{2}{c|}{\textbf{FEM}} & \multicolumn{2}{|c|}{\textbf{MLFEM}} & \multicolumn{2}{|c}{\textbf{GRAD-CAM}}\\
                \hline
                {\textbf{$l$}} & {\textbf{Well}} & {\textbf{Badly}} & {\textbf{Well}} & {\textbf{Badly}} & {\textbf{Well}} & {\textbf{Badly}}\\
                \hline 
                \hline
                1 $\rightarrow$ 2  & 0.376\% & 0.148\%      & 1.479\% & 0.169\%              & 0.090\% & 2.388\%\\[5pt]
                2 $\rightarrow$ 3  & 0.461\% & 0.280\%      & 0.923\% & 1.396\%             & 1.992\% & 0.313\%\\[5pt]
                3 $\rightarrow$ 4  & \textbf{0.032\%} & 1.711\%      & 1.758\% & 2.167\%              & 2.362\% & 0.307\%\\[5pt]
                4 $\rightarrow$ 5  & 0.905\% & 0.016\%      & 0.705\% & 1.136\%              & 2.068\% & 0.515\%\\[5pt]
                5 $\rightarrow$ 6  & 0.082\% & 0.097\%      & 0.720\% & 0.295\%             & 1.929\% & 2.406\%\\[5pt]
                6 $\rightarrow$ 7  & 0.469\% & 0.595\%      & 0.422\% & 1.193\%             & 3.417\% & 2.785\%\\[5pt]
                7 $\rightarrow$ 8  & 1.226\% & 0.649\%      & 1.255\% & 2.095\%             & 1.514\% & 1.856\%\\[5pt]
                8 $\rightarrow$ 9  & 0.204\% & \textbf{0.007\%}      & 0.319\% & 0.247\%             & 2.533\% & 0.560\%\\[5pt]
                9 $\rightarrow$ 10  & 1.086\% & 0.276\%     & 1.621\% & 1.484\%             & 1.822\% & 0.070\%\\[5pt]
                \hline
            \end{tabular}
        }
        
    \end{table}


\end{document}